\documentclass{article}

\usepackage[main, final, nonatbib]{neurips_2026}
\usepackage{xcolor}

\newcommand{\ap}[1]{\textcolor{black}{#1}}
\newcommand{\rd}[1]{\textcolor{black}{#1}}

\usepackage[utf8]{inputenc} % allow utf-8 input
\usepackage[T1]{fontenc}    % use 8-bit T1 fonts
\usepackage{hyperref}       % hyperlinks
\usepackage{url}            % simple URL typesetting
\usepackage{amsthm}
\usepackage{booktabs}       % professional-quality tables
\usepackage{amsfonts}       % blackboard math symbols
\usepackage{nicefrac}       % compact symbols for 1/2, etc.
\usepackage{microtype}      % microtypography
\usepackage{xcolor}         % colors
\usepackage{graphicx}
\usepackage{bm}
\usepackage{amssymb}
\usepackage{xcolor}
\usepackage{lipsum}
\usepackage{amsmath,amssymb,amsfonts,amsthm}

\usepackage{wrapfig}

\usepackage{hyperref}

\hypersetup{
    colorlinks=true,
    linkcolor=blue,   % internal links (sections, refs)
    citecolor=blue,   % citations
    urlcolor=black    % URLs
}

\usepackage[
    backend=biber,
    style=numeric-comp,
    sorting=none,
    giveninits=true,
    maxnames=1,
    doi=false,
    isbn=false,
    url=false
]{biblatex}

\DeclareCiteCommand{\cite}[\mkbibsuperscript]
  {\usebibmacro{cite:init}\usebibmacro{prenote}}
  {\usebibmacro{citeindex}\usebibmacro{cite:comp}}
  {}
  {\usebibmacro{cite:dump}\usebibmacro{postnote}}

\usepackage{enumitem}
\setlist[itemize]{leftmargin=0pt}
\setlist[enumerate]{leftmargin=0pt}

\newtheorem{theorem}{Theorem}[section]
\newtheorem{proposition}[theorem]{Proposition}
\newtheorem{lemma}[theorem]{Lemma}
\newtheorem{corollary}[theorem]{Corollary}
\theoremstyle{definition}
\newtheorem{definition}[theorem]{Definition}

\theoremstyle{remark}

\begin{document}

\title{The balance between feature learning and collapse\\in \rd{dynamical systems-based} generative models}

\author{
  Julian Brandon \\
  Département d'Etudes Cognitives\\
  Ecole Normale Supérieure, PSL\\
  \texttt{julian.brandon@ens.psl.eu}
    \AND
    N Alex Cayco-Gajic\footnotemark[1] \\
  Département d'Etudes Cognitives\\
  Ecole Normale Supérieure, PSL\\
  \texttt{natasha.cayco.gajic@ens.psl.eu}
  \And  
  Arthur Pellegrino\thanks{These authors jointly supervised this work.} \\
  Département d'Informatique\\
  Ecole Normale Supérieure, PSL\\
  \texttt{arthur.pellegrino@ens.fr}
}

\author{%
  Julian Brandon \quad Bruno Loureiro \quad
  N Alex Cayco-Gajic\thanks{These authors jointly supervised this work.} \quad
  Arthur Pellegrino\footnotemark[1] \\
  \vspace{-3pt}
  \\
  Ecole Normale Supérieure, PSL \\
    \vspace{-3pt}
  \\
  Correspondence: \texttt{julian.brandon@ens.psl.eu}
}

\maketitle

\begin{abstract}
Over the past decade, a growing body of work has established that implicit biases in learning dynamics fundamentally shape the solutions found by neural networks, governing their learned features and generalisation performance. However, despite these strides in deep learning theory, far less is known about representation learning in generative models based on dynamical systems, from normalising flows to diffusion. Importantly, \ap{empirical evidence suggests that} continuous-time generative models can be prone to mode collapse, yet \ap{a learning theory explaining} such biases remains elusive. 
\rd{To address this, we use tools from dynamical systems and operator theory to express the gradient flow of a general class of stochastic differential equations as a composition of linear operators, allowing us to relate collapse to low-rank biases in generative models. In variational inference and flow-matching tasks, we show that this bias can manifest in two different mechanisms: \emph{i)} where collapsed models are stable fixed points of the gradient flow, or \emph{ii)} in sequential feature learning, where smaller modes take progressively longer to be learned, producing an effective collapse for a fixed training budget.  
Together these findings point to a trade-off between feature learning and collapse, suggesting an operator view as a step towards a unified framework for understanding the learning dynamics of dynamical systems-based generative models.}
\end{abstract}

\section{Introduction}

\vspace{-10pt}

Neural networks exhibit learning phenomena that are not predicted by classical optimisation theory developed for convex problems\cite{curth2023u, power_grokking_2022}. This observation has led to the development of new theoretical approaches aimed at studying the \textit{learning dynamics} of neural networks\cite{saxe_exact_2014}. This perspective makes it possible to apply tools from dynamical systems theory to analyse the rich behaviours observed during training\cite{saad1995line, arnaboldi_high-dimensional_2023}. For example, gradient flow analyses on idealised settings provide information about stability of fixed points\cite{soletskyi_theoretical_2024, ninou2026curl}, and different forms of implicit regularisation\cite{soudry2018implicit,ali_implicit_2020}. Despite being idealised, several phenomena observed in large-scale training can be explained in this framework: staircase-like patterns in the loss curve can correspond to trajectories passing near saddle points, reflecting the sequential learning of data features\cite{saad1995line, goldt_dynamics_2020, abbe2021staircase, jacot_saddle--saddle_2021}; and slow changes in parameters can occur even when the loss appears constant, corresponding to a ``silent'' alignment of the network weights with relevant directions in the data\cite{atanasov_neural_2021}. Together, these illustrate that dynamical systems provide a powerful framework to investigate the rich behaviours observed in neural network training.

Despite this success, learning theory is currently under-developed for generative models based on differential equations. Yet, such dynamical systems are at the centre of flow and diffusion models, which are the state-of-the-art for generative models in vision \cite{song2020score, rombach2022high}, biology \cite{geffner2025proteina, shang2025predicting} and physics\cite{huang2024diffusionpde, holzschuh2023solving}. 

\begin{figure}[h]
\centering

\begin{minipage}[t]{0.44\linewidth}\vspace{0pt}
  \setlength{\abovecaptionskip}{0pt}
  \setlength{\belowcaptionskip}{0pt}
  \caption{\textbf{Low-rank biases in learning can drive mode collapse in generative models.} 
  Recent empirical evidence has suggested that generative models can exhibit learning biases such as memorisation of data samples (left) or mode collapse (right). Here, we argue that the latter can be caused by low-rank gradients biasing the learning dynamics.}
\end{minipage}
\hfill
\begin{minipage}[t]{0.54\linewidth}\vspace{0pt}
  \centering
  \includegraphics[width=\linewidth]{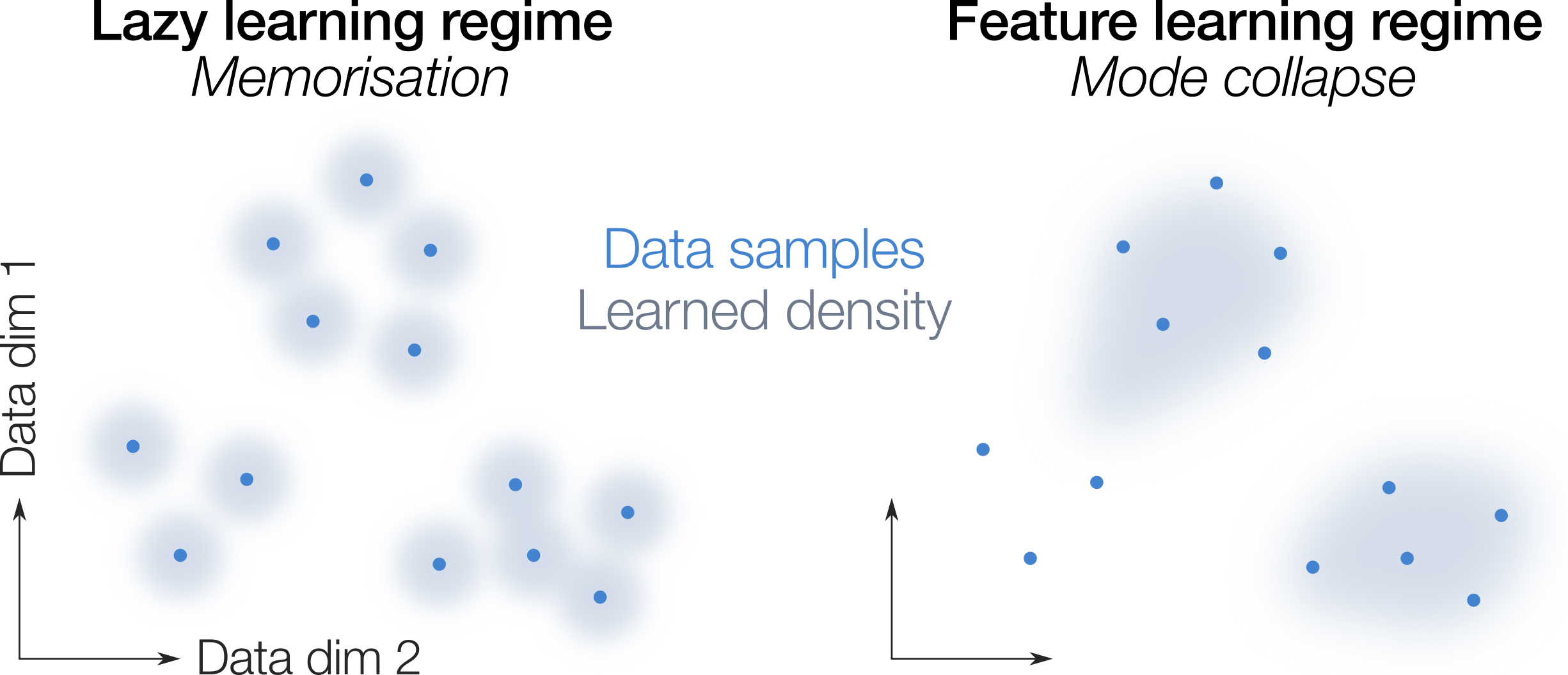}
\end{minipage}

\vspace{-15pt}

\end{figure}

The need for these new mathematical tools is reinforced by the presence of \emph{implicit biases} in the distributions learned by generative models. Implicit biases emerge through training, limiting which solutions are learned through optimisation. For example, even when a generative model is expressive enough to learn the full data distribution, training may favour solutions that capture only a subset of the distribution's modes\cite{minka2005divergence, midgley2023flow, blessing2024beyond, qin_class-balancing_2023,xiao2025algorithmic,zhang2025verbalized}, a phenomenon known as ``mode collapse''.
An immediate adverse consequence of mode collapse is a reduction in the diversity of generated samples, where regions of the data distribution may be ignored. More broadly, the loss of sample diversity could exacerbate more pernicious forms of degeneration if models are recursively trained on synthetic data\cite{shumailov_ai_2024}.

Implicit biases have been extensively characterised in the learning dynamics of deep networks. For example, an emerging line of work in learning theory has shown that deep neural networks tend to exhibit simplicity biases, where learning favours low-rank\cite{huh_low-rank_2023} or low-frequency components of the data\cite{canatar_spectral_2021}. More recently, key advances in dynamical systems learning theory stemming from functional analysis and statistical physics have enabled characterising similar biases in the learning dynamics of dynamical systems\cite{pellegrino2023low, hazelden2025kpflow, wang2025analytical}. 

Despite these results, the relationship between implicit biases induced by training and mode collapse in generative models remains unclear. We hypothesise that while implicit biases drive deep networks to learn low-dimensional data features, which is a beneficial form of representation learning, they are the same mechanism underlying mode collapse in dynamical systems-based generative models.

To support this hypothesis, we employ mathematical tools stemming from dynamical systems and operator theory to \rd{gain insight into collapse in two different model classes.} 
We show that low-rank biases \rd{can arise in generative models as a direct consequence of low-rank operators constraining the rank of the gradient, which can induce collapse.}

To validate this generally applicable theoretical framework in a controlled setting, we focus on an analytical study of simplified linear generative dynamical systems, allowing us to isolate the mechanisms by which the loss shapes the learning dynamics. We show that distinct objectives induce low-rank solutions through different architectural and data-driven effects. For variational inference, we find that mode collapse can arise from a rank bottleneck in the operators defining the gradient, which yields a stable collapsed fixed point in the gradient flow. In contrast, flow-matching objectives avoid stable sub-optimal solutions, but instead exhibit an effective mode collapse: learning proceeds through saddle-to-saddle transitions, leading to sequential acquisition of modes with progressively longer learning times. Finally, we show similar phenomena in \rd{non-linear} generative models, demonstrating that the same qualitative mechanisms persist in practical settings.

\rd{Overall, our results reveal low-rank bias as a driver of different types of collapse, highlighting an operator framework as a key tool in understanding the learning dynamics of dynamical systems-based generative models.}

\vspace{-8pt}

\subsection*{Contributions}

\vspace{-8pt}

\begin{itemize}[leftmargin=1em, itemsep=0pt, parsep=0pt, topsep=0pt]
    
    \item \textit{We apply an operator \rd{decomposition to a general class of models (SDEs), allowing us to derive a mathematical bound on the rank of the gradient for dynamical systems based generative models.}}
    \item \textit{We analytically characterise mode collapse in \rd{linear} continuous normalising flows trained with the reverse Kullback-Leibler (KL) objective, showing that collapse arises as a stable fixed point of the gradient. Collapsed weights have lower effective rank than non-collapsed ones.}
    \item \textit{We show that while flow-matching objectives avoid stable collapsed fixed points, \rd{linear} models \rd{can} exhibit an ``effective collapse'' due to saddle-to-saddle dynamics induced by data imbalance constraining the rank of the operators.}
    \rd{\item \textit{We validate our results in non-linear flow-matching models with imbalanced modes, showing that sequential learning of modes \rd{driving effective collapse} persists in more practical settings\rd{, and that operator ranks provide a diagnostic for the learning time.}}}
\end{itemize}

\vspace{-8pt}

\subsection*{Related works}

\vspace{-6pt}

\textbf{\textit{Learning theory and implicit biases.}} Prior work has shown that gradient flow tends to implicitly favour low-complexity solutions in overparameterised networks, including minimum-norm solutions\cite{gunasekar2017implicit, chizat_implicit_2020, pesme2021implicit}, maximum-margin classifiers\cite{soudry2018implicit}, algorithmically simple functions \cite{shah2020pitfalls,valle2018deep}, low-frequency modes\cite{rahaman_spectral_2019, canatar_spectral_2021}, and low-rank embeddings\cite{gunasekar2017implicit,arora2019implicit,huh_low-rank_2023}. Low-rank biases in particular have been studied extensively in deep linear networks, generally with small weight initialisations, where it has been shown that the data singular values are learned one by one\cite{saxe_exact_2014}. This sequential feature learning produces plateaus in the loss referred to as ``saddle-to-saddle'' dynamics, where each saddle corresponds to an increase in the rank of the weights\cite{saad1995line, goldt_dynamics_2020,jacot_saddle--saddle_2021,pesme_saddle--saddle_2023,bantzis2025saddle}. Here we argue that similar saddle-to-saddle learning dynamics unfold during flow matching, and that increased learning time between each saddle drives an effective collapse. While the learning dynamics of generative models are less studied overall, linear diffusion has been shown to produce distributional simplicity biases\cite{wang2025analytical, merger2025generalization, bardone2026theory}. In particular, linear denoisers have been shown to sequentially learn eigenmodes of the input covariance\cite{wang2025analytical}.  
\rd{Unlike these works however, here we focus on understanding these biases in models trained on multi-modal distributions, especially under data imbalance.}

\textbf{\textit{Operator view of learning dynamics.}} Recent work has suggested that low-rank biases also emerge in recurrent architectures by using adjoint dynamics to represent gradient flow as a composition of linear operators\cite{pellegrino2023low,hazelden2025kpflow}. In particular, one study derived bounds on the rank of weight updates in linear recurrent networks using this operator view\cite{pellegrino2023low}. 
Koopman operator theory has \rd{also} been applied to diffusion to interpret the sampling process\cite{berman2025one, bai2025hierarchical}, but not training dynamics.
Our work applies a similar operator perspective to the learning dynamics of continuous-time generative models\rd{, and in particular applies it to mode collapse}.

\textbf{\textit{Collapse in generative models.}} Generative models are known to exhibit mode collapse, in which the full diversity of modes in the data distribution are not captured\cite{nicoli2023detecting,abbahaddougeometry,cui_asymptotics_2024, barcelo_avoiding_2024, he2024training}. Collapse can come from many sources: for example, distributional biases can cause mode collapse when there is class imbalance due to long-tailed distributions\cite{qin_class-balancing_2023, deasy_heavy-tailed_2022,debiasing2026}. In more extreme cases, e.g., when generative models are used to train other generative models, this can lead to model collapse\cite{shumailov_ai_2024}. Architectural biases due to the network parametrisation of diffusion models can in some cases promote strong generalisation\cite{kadkhodaie2023generalization}, or strong collapse\cite{cui_solvable_2025}. Mode collapse can also come from the objective, for example recently studied in variational inference\cite{soletskyi_theoretical_2024, fogliani2026annealing, gx2025kl}. Previous work has noted a balance between generalisation and memorisation in flow matching and diffusion\cite{yoon2023diffusion,gao2024flow,li2026kinetic, gao_simplicity_2015-1,garnier2026biased}. Our results tie this trade-off to the rank of weight parameters and a balance between generalisation and collapse. 

\section{Theory}

\textbf{Problem setup.} 
\rd{The results in this section relate to} a general class of dynamical systems-based generative models:
\begin{equation}\label{eq:SDE_def}
    \textstyle \rd{\mathbf{d}}\mathbf{x}\rd{(t)} = \mathbf{f}_{\boldsymbol{\theta}}(\mathbf{x}\rd{(t)}, t)dt + G_{\boldsymbol{\theta}}(\mathbf{x}\rd{(t)}, t) \circ \mathbf{dB}, \quad \mathbf{x}(0) \sim p_0
\end{equation}
where $\mathbf{x}(t)$ is the state of the system, $\mathbf{dB}$ increments of a Brownian motion, $\mathbf{f}_{\boldsymbol{\theta}}:\mathbb{R}^d\times \mathbb{R}\rightarrow\mathbb{R}^d$ and  $G_{\boldsymbol{\theta}}:\mathbb{R}^d\times\mathbb{R}\rightarrow \mathbb{R}^{d\times d}$ the drift and diffusion components of the stochastic differential equation with parameters $\boldsymbol{\theta}\in \Theta$\rd{, $(\circ)$ denotes the Stratonovich integral, and $p_0$ is a chosen base distribution}. 
We assume a general loss function:
\begin{equation}\label{eq:loss_def}
    \mathcal{L}(\boldsymbol{\theta}) = \mathbb{E}_{\mathbf{x}_0, \mathbf{B}}\left[\int_0^T \ell(\mathbf{x}(t), t) dt \rd{\,+\,\Phi(\mathbf{x}(T))}\right],
\end{equation}
where $\ell:\mathbb{R}^d \times [0, T] \rightarrow \mathbb{R}$ is the instantaneous loss \rd{and $\Phi:\mathbb{R}^d\to\mathbb{R}$ is the terminal loss which only depends on $\boldsymbol{\theta}$ through $\mathbf{x}(T)$}. Depending on the form of \rd{$\ell(\cdot)$ and $\Phi(\cdot)$}, this setting includes variational, flow matching and score-matching objectives as special cases. 

Here, we are interested in the case where $\boldsymbol{\theta}$ contains weights $W\in\mathbb{R}^{\rd{k \times q}}$ \rd{which act linearly on activations}, parametrising the drift and diffusion terms $\mathbf{f}_{\boldsymbol{\theta}}$ and $G_{\boldsymbol{\theta}}$\rd{, or projecting to the instantaneous loss}. In particular, the rank of $W$ can constrain the trajectories of the dynamical system and, therefore, the probability distribution learned by the generative model. As we argue below, implicit low-rank biases can arise in the gradient flow learning dynamics, which can in turn bias the model towards collapsing certain modes of the learned distribution. To analyse this phenomenon, we study gradient flow by decomposing it into two linear operators.

\textbf{An operator view of gradient flow.} We are interested in the learning dynamics of generative models under gradient flow. Using adjoint dynamics, we show that, similarly to the neural tangent kernel of \textit{deep networks}, the gradient of weight parameters of dynamical systems can be written as a composition of linear operators. In appendix~\ref{apd:adjoint_derivation} we provide a thorough derivation of these results along with illustrative examples. In summary:

\begin{theorem}\label{thm:operator}
    \rd{Consider a model as defined in Eq.~\ref{eq:SDE_def} with loss as defined in Eq.~\ref{eq:loss_def}. For any parameter $W\in\mathbb{R}^{k\times q}\subseteq \boldsymbol{\theta}$ which acts linearly on activations $\mathbf{h}_{(\cdot)}(t)\in\mathbb{R}^q$ via $\mathbf{z}_{(\cdot)}(t)=W\mathbf{h}_{(\cdot)}(t)\in\mathbb{R}^{k}$, where $(\cdot)$ denotes a projection to either the instantaneous loss, drift or diffusion terms, the gradient of $\mathcal{L}$ with respect to $W$ can be written as the composition of two linear operators:}
    \[
        \nabla_W \mathcal{L} = A_{(\cdot)} \circ H_{(\cdot)}, \qquad A_{(\cdot)}:L^2(\mathbb{R}) \rightarrow \mathbb{R}^k, \qquad H_{(\cdot)}: \mathbb{R}^q \rightarrow L^2(\mathbb{R})
    \]
    where the linear operators act via integration:
    \[
        A_{(\cdot)}(f) = \mathbb{E}_{\mathbf{x}_0, \mathbf{B}}\left[\int_0^T \mathbf{a}_{(\cdot)}(t)f(t) dt\right], 
        \rd{\qquad (H_{(\cdot)}(\mathbf{v}))(t) = \mathbf{h}_{(\cdot)}(t) \cdot \mathbf{v}} 
    \]
    where $f\in L^2(\mathbb{R})$, $\mathbf{v}\in \mathbb{R}^q$ \rd{, and $\mathbf{a}_{(\cdot)}(t)\in \mathbb{R}^k$. The specific form of the adjoint $\mathbf{a}_{(\cdot)}(t)$ depends on the projection $(\cdot)$ (App.~\ref{apd:full_gradient_AoH}).}
    ~\hfill \hyperref[proof:operator]{\sc\small [Proof]}
\end{theorem}
\rd{Thus the gradient with respect to $W$ can be expressed in terms of two operators, an \emph{adjoint} operator $A$ and a \emph{state} operator $H$, which} compose as an integral over rank-1 terms.

\textbf{Low-rank biases.} Key results from functional analysis\cite{conway2019course} say that $A$ and $H$ are compact, and admit singular value decompositions:
\[
    A = U^{a} \circ S^{a} \circ V^{a}, \qquad H = U^{h} \circ S^{h} \circ V^{h}
\]
Furthermore, because they are maps between or from a finite-dimensional vector space, they have finite rank (i.e. finitely many non-zero diagonal entries of $S^a$ and $S^h$). It immediately follows that the singular values of their composition (i.e. of the gradient) are directly determined by i) the overlap between the right singular vectors $V^{a}$ and the left singular vectors $U^{h}$ and ii) the singular values $S^a, S^h$. 

For example, usual bounds on the rank of matrix products also hold for linear operators, and the rank of the gradient of the weights can be bounded as: $\operatorname{Rank}(\nabla_W \mathcal{L}) \leq \min \{\operatorname{Rank}(A), \operatorname{Rank}(H)\}$.
In practical settings, however, these operators will be full rank, so we may be more interested in a numerically stable definition of the \emph{effective} rank.
\begin{definition}
    The stable rank of a compact bounded linear operator $A$ is defined as:
    \[
        \operatorname{SRank}(A) := \frac{\lVert A \rVert_*}{\lVert A \rVert_2} = \sum_{i=1}^m\frac{ s_i^a}{s_1^a}
    \]
    where $\lVert \,\cdot\, \rVert_*$ denotes the nuclear norm and $\lVert \,\cdot\, \rVert_2$ the spectral norm and $s_i^a$ the singular values of $A$.
\end{definition}
In particular, if all nonzero singular values are equal, the stable rank coincides with the rank. Furthermore, because it is only normalised by the top singular value, the stable rank is less sensitive to the tails of the singular value distribution than other smooth ranks, such as the participation ratio.
\begin{theorem}\label{thm:srank_bound} The stable rank of the gradient can be bounded as:
    \[
        \operatorname{SRank}(\nabla_W \mathcal{L}) \leq \frac{\lVert A \rVert_2 \lVert H \rVert_2}{\lVert A\circ H\rVert_2}\min \left\{\operatorname{SRank}(A), \operatorname{SRank}(H)\right\}.
    \]
     where $A$ and $H$ are the operators defined above. ~\hfill \hyperref[proof:srank_bound]{\sc\small [Proof]}
\end{theorem}

\rd{The prefactor depends on the alignment between the right and left singular vectors of $A$ and $H$ respectively, equalling 1 when they are aligned and bounded above by the top singular vectors according to $\lvert \langle \mathbf{v}_1^A, \mathbf{u}_1^H\rangle \rvert^{-1}$ (App.~\ref{proof:srank_bound}).}

\textit{Thus, in general, the weights will be biased towards lower-rank solutions if either the $A$ or $H$ operator is low rank. Such low-rank structure can arise through various mechanisms, including properties of the loss function or constraints imposed by the model architecture. In subsequent sections, we show that this bias can cause dynamical systems-based generative models to collapse certain data modes, either by converging to stable but suboptimal gradient fixed points during variational inference, or by exhibiting slow learning due to saddle-to-saddle dynamics in flow matching.}

\section{Results}

\subsection{Mode collapse as an attractor in learning dynamics under a variational objective}\label{sec:CNF_results}

To understand how learning biases drive collapse, we first consider a variational inference task, for which generative models are known to have a mode-seeking behaviour. Let $p_\theta(\mathbf{x})$ denote a parametric family of distributions, whose parameters $\theta$ are optimized to fit a target distribution $p^*(\mathbf{x})$ by minimizing the `reverse' Kullback Leibler divergence: 
\[
D_{\mathrm{KL}}(p_\theta(\mathbf{x})\,\|\,p^*(\mathbf{x})) = \mathbb{E}_{p_\theta}\left[ \log p_\theta(\mathbf{x}) \right] - \mathbb{E}_{p_\theta}\left[ \log p^*(\mathbf{x}) \right].
\]
The reverse KL is commonly used to train neural samplers when only an unnormalised density or an energy function is available \cite{nicoli2023detecting,he2024training}. Due to its asymmetric form, the reverse KL is intrinsically `mode-seeking', as regions in the data distribution are ignored where $p_\theta(\mathbf{x})=0$. This provides an ideal case study to analytically study how the interplay between the state and adjoint operators leads to a low-rank bottleneck in the weight updates. As we will demonstrate, this results in collapse becoming an attractor of the gradient flow. 

\textbf{Model.} For analytical tractability, we consider a linear \emph{continuous normalising flow} (CNF):
\[
    \textstyle\frac{d}{dt}\mathbf{x}(t) = W\mathbf{x}(t),\qquad \mathbf{x}(0) \sim p_0
\]
\rd{where $p_0$ is the base distribution which can be sampled from.} The model is trained on a variational objective under gradient flow:\vspace{-5pt}
\[
    \mathcal{L}(W) = D_{\mathrm{KL}}(p_1(\mathbf{x})\,\|\, p^*(\mathbf{x})), \qquad \textstyle\frac{d}{d\tau} W(\tau) = - \nabla_W \mathcal{L}(W(\tau))
\]
where $\tau$ is the training iteration and $p_1(\mathbf{x})$ is the distribution obtained by pushing forward samples from $p_0$ to time $t=1$. \rd{In the notation of Eq.~\ref{eq:SDE_def} and Eq.~\ref{eq:loss_def}, this model has $G=0$, and $\ell$ independent of $\mathbf{x}$, so the adjoint is driven only by the drift term $\mathbf{f}_{\boldsymbol{\theta}}$.}
The operators can be found using the adjoint method \cite{pontryagin2018mathematical}. %\\

\begin{definition}
    For a dynamical system $\dot{\mathbf{x}}=\mathbf{f}_{\theta}(\mathbf{x}(t), t)$ defined on $t\in[0,1]$, with loss functional $\mathcal{L}: \mathbb{R}^d\rightarrow\mathbb{R}$, the state adjoint is defined as $\mathbf{a}(t)=\frac{d\mathcal{L}(\mathbf{x}(1))}{d\mathbf{x}(t)}$. Its solution is given by the following dynamical system and terminal condition \rd{depending on} $J_{\mathbf{f}}$ the Jacobian of $\mathbf{f}_\theta$:
    \[
        \textstyle\frac{d}{dt}\mathbf{a} = -J^\top_{\mathbf{f}} \mathbf{a}, \qquad \mathbf{a}(1) = \frac{d\mathcal{L}}{d\mathbf{x}(1)}
        \vspace{-5pt}
    \]
\end{definition}
For the linear model and variational loss above, the adjoint dynamical system is therefore given by:
\[
    \textstyle\frac{d}{dt}\mathbf{a} = -W^\top \mathbf{a}, \qquad \mathbf{a}(1)  =-\nabla_{\mathbf{x}} \log p^{\star}(\mathbf{x}(1))
\]
\rd{Since there is no hidden layer, the state operator $H$ is simply the transported data $\mathbf{x}(t)$.}
We can then analytically derive the gradient as a Fréchet derivative of the matrix exponential of $W^\top$. 
\begin{theorem}\label{thm:CNF_gradient}
    The \rd{weight gradient is} $\nabla_W \mathcal{L} = A \circ H -I_d$, 
    \rd{where the first term }simplifies to:\vspace{-4pt}
    \[
        A \circ H = \mathbb{E}_{\mathbf{x}_0\sim p_0}\left[\int_0^1 \exp{\left(W^\top (1-t)\right)}(\mathbf{a}(1)\otimes \mathbf{x}(0))\exp{\left(W^\top t\right)}dt\right]
        \vspace{-17pt}
    \] 
    ~\hfill \hyperref[proof:CNF_gradient]{\sc\small [Proof]}
\end{theorem}

\begin{figure}[h]
    \centering
    \includegraphics[width=\linewidth]{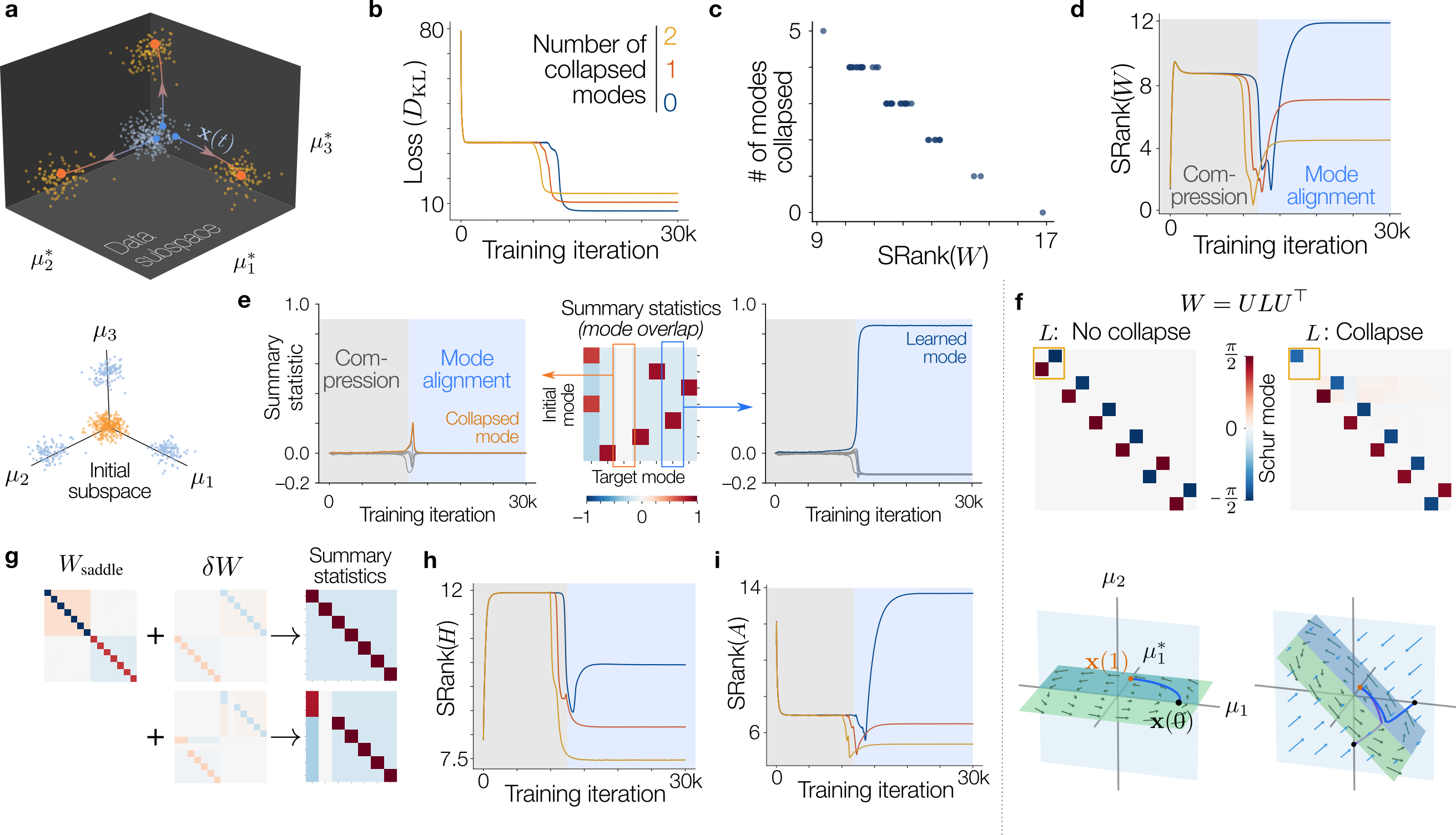}
    \vspace{-10pt}
    \caption{\textbf{Stable low-rank solutions in continuous normalising flows drive mode collapse.} \textbf{a.} The model is trained to transport a Gaussian mixture from orthogonal initial and data subspaces. \textbf{b.} Loss over iterations in models with different numbers of collapsed modes. 
    \textbf{c.} The number of collapsed modes scales linearly with the stable rank of the weights $W$ at convergence. \textbf{d.} The stable rank of the weights over training in models with different numbers of modes. \textbf{e.} The overlap between  a pushed forward initial mode and a target mode is a summary statistic and gives us a low-dimensional description of training.
    This highlights two phases of learning: compression of the initial modes and learning the data modes. \textbf{f.} Top: Schur decomposition of the post-training weights in collapsed and non-collapsed models. Bottom: Non-collapsed models \rd{\textit{(left)}} learn a rotational vector field. Collapsed models \rd{\textit{(right)}} reveal attraction towards a stable slow manifold with rotational dynamics. \textbf{g.} The saddle weights and perturbations in the direction of collapsed or non-collapsed weights. Training from this point induces the respective solutions, represented by the summary statistic matrix at convergence.
    \textbf{h-i.} The stable rank of the state and adjoint operators over training.
    }
    \label{fig:f2}
\end{figure}

\textbf{Task.} In this linear model, it is not possible for the flow to split an initial mode (i.e., from Gaussian $p_0$) to multiple target modes. Instead, we constrain both the initial and data distributions to be equally-weighted Gaussian mixtures, with $p_0$ and $p^{\star}$ containing an identical number of modes. \rd{Because the linear flow has no bias term, it can only transport mass about the origin. We therefore also centre both mixtures, which leaves all inter-mode distances unchanged and is equivalent to an affine flow whose translation is learned trivially. The initial and target distributions are then defined as:}
\[
    p_0(\mathbf{x}) = \sum_{i=1}^m \frac{1}{m}\mathcal{N}(\mathbf{x} ; \boldsymbol{\mu}_i, \sigma^2 I_d), \quad p^*(\mathbf{x}) = \sum_{i=1}^m \frac{1}{m}\mathcal{N}(\mathbf{x} ; \boldsymbol{\mu}_i^*, \sigma^2 I_d), \quad \boldsymbol{\mu}_i \cdot \boldsymbol{\mu}^*_j = 0, \quad \rd{\boldsymbol{\bar{\mu}}=\boldsymbol{\bar{\mu}}^*=\mathbf{0}}
\]
\rd{where $\boldsymbol{\bar{\mu}}=\sum_i\boldsymbol{\mu}_i$, $\boldsymbol{\bar{\mu}}^*=\sum_i\boldsymbol{\mu}_i^*$. By constructing the initial and target modes symmetrically, we also ensure the bound prefactor in Theorem~\ref{thm:srank_bound} is 1 at initialisation (App.~\ref{apd:cnf_prefactor_proof}).} 
Under these assumptions, there exists a linear transport between $p_0$ and $p^{\star}$ by rotating the initial subspace to the data subspace (Fig.~\ref{fig:f2}a), meaning that any collapse is due to implicit bias rather than expressivity. We found that training the linear CNF with different weight initialisation seeds made it converge to different values of $D_\mathrm{KL}$, corresponding to different numbers of learned data modes (Fig.~\ref{fig:f2}b). 

\textbf{The rank of $W$ directly determines collapse.}

\rd{Empirically, we observe mode collapse to occur when the weights fail to learn a rotation mapping an initial mode to a target, instead contracting the difference between two modes and sending them to a single target. Each correctly transported mode requires a rank-2 rotational block in $W$,} so we expect that the rank deficiency of $W$ should lead to more collapsed modes. To quantify this effect, we computed the stable rank of the weights and observed that lower-rank solutions exhibited a greater number of collapsed modes (Fig.~\ref{fig:f2}c-d). \rd{Moreover, the learning dynamics leading to collapse were lower-dimensional than in non-collapsed solutions (Fig.~\ref{sup_fig:cnf}b-c).} To further characterise these solutions, we performed a \rd{real} Schur decomposition $W = U L U^\top$, where $L$ is quasi-lower triangular. Non-collapsed models implemented the correct rotations between initial and target modes, reflected in anti-symmetric blocks of $L$, each $2 \times 2$ block representing a rotation in the subspace from one initial mode to one target mode (Fig.~\ref{fig:f2}f). Instead, collapsed models did not learn all rotations, exhibiting compressive dynamics corresponding to negative real eigenvalues along the diagonal. 

\textbf{Summary statistics explain the low-rank collapse.} To understand the learning dynamics underlying such collapse\rd{, it can be informative to examine summary statistics of the gradient at different points in training (App.~\ref{apd:CNF_analyses}).}
In short, \rd{we defined } summary statistics \rd{$s_{ij}$} \rd{as} the \rd{scalar} overlaps between the transported initial means and target means \rd{at $t=1$}:\vspace{-3pt}
\[
    s\in \mathbb{R}^{m\times m}, \quad s_{ij} = \frac{(e^{W}\boldsymbol{\mu}_i) \cdot \boldsymbol{\mu}_j^*}{R^2}
    \vspace{-2pt}
\]
Over training, \rd{these }summary statistics \rd{underwent} two phases. In an initial ``compression phase'', \rd{the model reduces the loss by compressing the initial modes and reducing the effective covariance of the distribution under the flow. Here, the weights have no rotational component, so the overlaps $s_{ij}$ remain close to zero.
\rd{In the subsequent ``mode alignment phase'', the model rapidly} learns to map initial modes to target modes \rd{by a rotation}, settling at either collapsed or non-collapsed solutions. }

\rd{In non-collapsed runs, the summary statistics converge to a permutation matrix, whereas collapsed runs have columns with no large entries, corresponding to the target modes with no pushed forward initial mode aligned with them (Fig.~\ref{fig:f2}e). The compression, collapsed, and non-collapsed weights \rd{can be analytically shown to be} fixed points of the gradient flow (App.~\ref{apd:CNF_fixed_points}). From the compression fixed point, we can induce either solution by a perturbation in the direction of the corresponding fixed point, indicating that the compression weights are a saddle with unstable directions towards either collapsed or non-collapsed solutions (Fig.~\ref{fig:f2}g).}

\textbf{Adjoint \rd{and state} operator \rd{ranks track collapse.}}
\rd{To understand the interplay between collapse, summary statistics and the gradient operators}, we looked at the stable rank of the \rd{state and} adjoint operators over training (Fig.~\ref{fig:f2}h-i). \rd{This} revealed that in the mode alignment phase, \rd{both operator ranks} corresponding to collapsed solutions are lower than \rd{in} non-collapsed solutions, mirroring the decrease in rank we previously observed in the weights. \rd{The prefactor starts at 1 and then increases to 8.9 in the compression phase, before splitting across collapsed and non-collapsed models with collapsed models having a higher value than non-collapsing ones (Fig.~\ref{sup_fig:cnf}d).
The operator ranks exhibit non-linear responses to perturbations around the saddle, with perturbations in the direction of different solutions leading to distinct rank values (Fig.~\ref{fig:apd_perturbations}). This suggests that the operator rank could be useful as an early diagnostic of collapse in less tractable models, where it is not clear what are the relevant summary statistics.}

\textit{In sum, variational objectives can make CNFs collapse to suboptimal solutions, even when the model is expressive enough to learn the optimal transport between the initial and target distributions. 
}

%\clearpage

\subsection{Saddle-to-saddle dynamics drive effective collapse during \rd{linear} flow matching}

In the previous section, we showed that collapse is a stable fixed point in the gradient flow. This tendency towards collapse comes from the form of the reverse KL which is known to produce mode seeking behaviour. However, \rd{under different objectives,} it is possible to effectively get mode collapse from saddle-to-saddle dynamics, which lead to sequential feature learning, similar to what has been observed in the learning dynamics of deep linear networks\cite{saxe_exact_2014}. 
\rd{Although this type of collapse is structurally distinct from the attractor-based collapse in CNFs, we argue that low-rank implicit biases combined with imbalanced data lead to \emph{effective} collapse under a finite training budget.}

\rd{\textbf{Model.}} Flow-matching models are trained by regressing the optimal velocity, $\mathbf{u}^*(\mathbf{x},t)\equiv\mathbf{u}^*_t(\mathbf{x})$, directly with a neural network model $\mathbf{u}^\theta(\mathbf{x},t)\equiv\mathbf{u}_t^{\theta}(\mathbf{x})$:
\[
    \mathcal{L}_{\text{FM}}=\int_0^1 \mathbb{E}_{\bar{\mathbf{x}}\sim p_t}\left[\left\|\mathbf{u}_t^*(\bar{\mathbf{x}})-\mathbf{u}_t^\theta(\bar{\mathbf{x}})\right\|^2\right]dt
\]
where $\bar{\mathbf{x}}\sim p_t$ is sampled from the chosen interpolant trajectory which interpolates between base distribution $p_0$ and target distribution $p^*$ over time $t\in[0,1]$, and $\mathbf{u}^*_t(\mathbf{x})$ is the vector field that generates it. 
\rd{This formulation contains no terminal loss, so $\Phi(\mathbf{x}(T))=0$ in Eq.~\ref{eq:loss_def}. Moreover, since the loss is integrated along the ground-truth trajectory $\mathbf{\bar x}$, the drift term $\mathbf{f}_{\boldsymbol{\theta}}$ from Eq.~\ref{eq:SDE_def} contains no learnable parameters. Thus, $\boldsymbol{\theta}$ enters only through the instantaneous loss $\ell$.}
In practice, the optimal field is intractable, and so an objective conditioned on individual data samples is used, which yields the same gradient as the unconditional objective above (App.~\ref{apd:FM}).

\rd{To study the training dynamics in a tractable setting analogous to the CNF case, we trained linear flows in two different parameterisations using a teacher-student setup with optimal velocity: 
\[
    \mathbf{u}_{t,\text{shallow}}^\theta(\mathbf{x})=W\mathbf{x}, \qquad \mathbf{u}_{t,\text{deep}}^\theta(\mathbf{x})=W_2W_1\mathbf{x} \qquad \mathbf{u}_t^*(\mathbf{x})=W^* \mathbf{x}
\]
where $W^*$ is constructed to transport initial modes to their respective targets. This ensures that the model is well-specified, i.e. the student is able to achieve the optimal solution, isolating the effects due to the training dynamics from the expressivity of the model.
We use the shallow model as a direct comparison to the CNF model, and the deep model to study sequential learning dynamics.
As before, our initial and target distributions are Gaussian mixtures, but now we take our modes to be orthogonal: $\boldsymbol{\mu}_i\cdot\boldsymbol{\mu}_j=\boldsymbol{\mu}_i^*\cdot\boldsymbol{\mu}_j^*=R^2 \delta_{ij}$.} 
\rd{This choice means the model can act independently on each mixture mode, allowing us to study the learning time of each mode separately in the context of sequential learning. By varying the weighting of the $i^\text{th}$ target mode, denoted by $\pi_i$, we will study how imbalanced data affects learning dynamics driven by low-rank implicit biases.}

\begin{figure}[h]
\centering

\begin{minipage}[t]{0.52\linewidth}\vspace{0pt}
  \centering
  \includegraphics[width=0.93\linewidth]{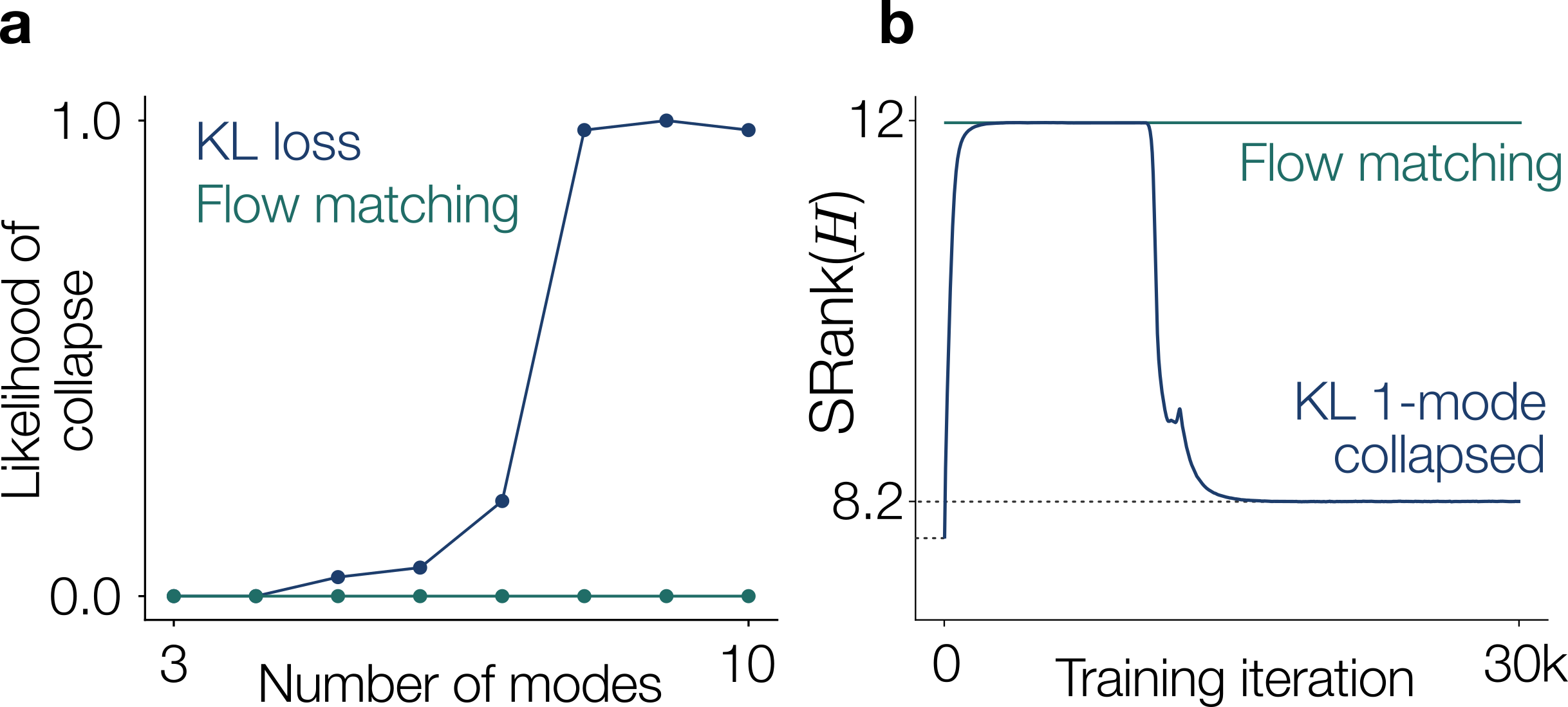}
\end{minipage}
\hfill
\begin{minipage}[t]{0.43\linewidth}\vspace{0pt}
  \setlength{\abovecaptionskip}{0pt}
  \setlength{\belowcaptionskip}{0pt}
  \caption{\textbf{In flow matching, collapse is not an attractor in the gradient flow.} 
  \textbf{a.} Likelihood of collapse as a function of the number of modes in the data for variational inference and flow-matching objectives. \textbf{b.} Rank of the state operators in the variational objective and flow-matching models over training.}
  \label{fig:fm_linear}
\end{minipage}

\vspace{-5pt}

\end{figure}

\textbf{Linear flow matching is robust to stable collapse.} 
As expected, \rd{when trained with equal mode weightings,} the model was able to recover the optimal vector field, and did not exhibit mode collapse, even as the number of modes in the target distribution increased (Fig. \ref{fig:fm_linear}a). This \rd{suggests} that the gradient flow of the flow-matching objective does not induce collapse as a stable attractor, in contrast to the variational objective.

To better understand the mechanisms that could lead to collapse under the flow-matching objective, we next adopt the operator view. The gradient of the linear flow-matching model can be written as:
\vspace{-3pt}
\[
    \nabla_W \mathcal{L} = A \circ H  = \mathbb{E}\left[\int_0^1 \mathbf{r}(t) \otimes \mathbf{\bar x}(t) dt\right]
    \vspace{-3pt}
\]
where the state operator $H$ is given by the interpolant trajectory $\mathbf{\bar x}(t)$, and the adjoint operator $A$ by the residual, $\mathbf{r}(t) =  W\mathbf{\bar x}(t) - \mathbf{u}_t^*(\bar{\mathbf{x}}(t))$. 
Importantly, here the state operator is independent of the weights, depending only on the optimal trajectory; thus its rank is constant throughout training. In contrast, under the variational objective, the stable rank of the state operator dropped at the transition between the compression and mode-learning phases, driven by changes in the weights $W$ (Fig. \ref{fig:fm_linear}b). \rd{This suggests that the training dynamics of flow-matching models may become biased if the interpolant trajectory spectrum is imbalanced, affecting the stable rank of the state operator.}

\begin{figure}[b]
\centering
\begin{minipage}[t]{0.48\linewidth}\vspace{0pt}
  \setlength{\abovecaptionskip}{0pt}
  \setlength{\belowcaptionskip}{0pt}
  \caption{\textbf{Flow matching can have saddle-to-saddle dynamics under imbalanced modes.} \textbf{a.} Flow-matching loss and summary statistics of the model over training (with weights $\pi=[0.01, 0.09, 0.90]$). Both display staircase-like pattern corresponding to the sequential matching of initial to target modes over learning. Plateaus sit at roughly $s_{ii}\approx0.79$ as predicted. \textbf{b.} Rank of the state operator as a function of the imbalance of the weights of the mixture for 7 modes with small mode $\pi_\text{min}=\alpha$ and other modes $\pi_i=\frac{1-\alpha}{6}$. The mode imbalance is $1-\frac{\pi_\text{min}}{\pi_i}$.}
  \label{fig:saddle_linear_fm}
\end{minipage}
\hfill
\begin{minipage}[t]{0.5\linewidth}\vspace{0pt}
  \centering
  \includegraphics[width=\linewidth]{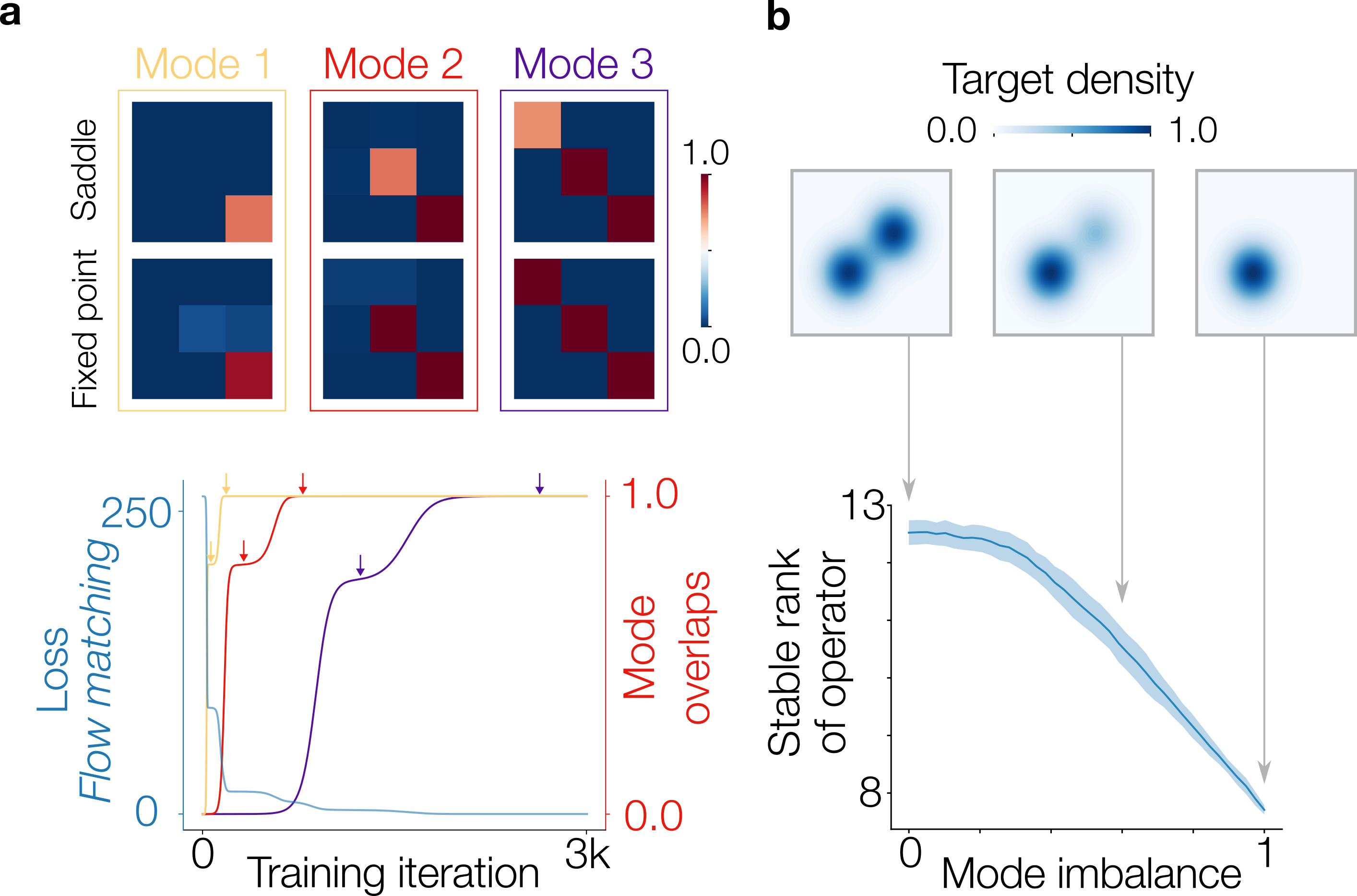}
\end{minipage}

\vspace{-5pt}

\end{figure}

\textbf{\rd{Separated learning times of imbalanced modes} drive effective collapse.} Prior work has shown that deep linear networks exhibit sequential feature learning \cite{saxe_exact_2014}. We asked whether this could serve as a mechanism for effective collapse, causing individual modes to be learned sequentially at different time scales. 
\rd{Since the state operator directly depends on the data spectrum, we hypothesised that introducing non-uniform class weightings might drive the model towards biased solutions. We thus trained linear models 
on Gaussian mixtures with imbalanced mode weightings.}
In this setting, the singular value spectrum of the \rd{state operator }can be explicitly derived as a function of these weights, and used to analytically calculate \rd{its rank, showing that larger mode imbalance decreases the operator rank} (App.~\ref{apd:rank_data_op}, Fig~\ref{fig:saddle_linear_fm}b). 
For imbalanced target distributions,  deep linear models exhibited saddle-to-saddle patterns in the loss (Fig.~\ref{fig:saddle_linear_fm}a).

\rd{\textbf{Summary statistics characterise mode learning times.} Computing the overlap summary statistics, $s_{ij}$, for each initial-target mode pair over training revealed that progressively smaller modes are learned sequentially, with each mode undergoing its own saddle-like phase. Following prior work\cite{saxe_exact_2014}, the learning dynamics can be solved exactly, and we find that the learning time of mode $i$ is inversely related to its corresponding mixture weight $\pi_i$, scaling as $\mathcal{O}(\pi_i^{-1})$ (App.~\ref{apd:FM_sequential_learning}). 
Interestingly, learning a data mode corresponds to learning two singular mode directions with different learning times, so each mode's learning trajectory contains a plateau. The first rank-1 direction to be learned is a shear, mixing the initial and target directions to achieve partial overlap of modes. The second direction adds another rank-1 component which cancels the symmetric part of the shear and induces the rotation mapping from initial to target modes.
The bound prefactor began at 1 at initialisation and increased sequentially with each mode learnt to 6 at convergence (Fig.~\ref{fig:apd_linear_ops}).}

\textit{Thus, although the flow-matching objective prevents stable collapsed fixed points of the gradient flow in linear generative models, it can induce effective collapse by having slow collapsed saddles. 
}

%\clearpage

\vspace{5pt}

\subsection{Non-linear flow-matching models sequentially learn data modes}\label{sec:non-linear_FM}

\vspace{3pt}

So far we have focused on linear models for analytical tractability. However, our theoretical framework is general and applies to non-linear models. Non-linear models, unlike linear flows, can create new modes. We therefore asked whether similar sequential learning dynamics emerge in non-linear models, and whether this could contribute to a form of effective collapse.

First, we assume a Gaussian mixture target distribution with equal, isotropic covariances $\sigma_*^2 I_d$ and unequal mixture weights. However here, unlike in the linear settings, the initial distribution is a single-mode Gaussian.

\textbf{\rd{Operators in a tractable non-linear model.}} We trained a one hidden-layer ReLU network parameterised to match the form of the optimal velocity (App.~\ref{apd:optimal_FM_field}):
\[
    \mathbf{u}_\theta(\mathbf{x},t)=c(t)\mathbf{x}+(1-tc(t)) W_2\sigma(W_1\mathbf{x}+\mathbf{b}), \qquad c(t)=\frac{t\sigma_*^2 - (1-t)\sigma^2}{t^2\sigma_*^2 +(1-t)^2\sigma^2}
\]
where $c(t)$ is a scalar given analytically and $W_2\in\mathbb{R}^{n\times m}$, $W_1\in\mathbb{R}^{m\times n}$, $\mathbf{b}\in\mathbb{R}^m$ are trained parameters. By using such a parametrisation, the time-dependency is handled purely by the scalar prefactors, and the model only needs to learn the spatial component of the field. For such a model trained under the MSE loss with conditional velocity field target (App.~\ref{apd:nonlinear_fm}), we can write the gradients with respect to the two weight matrices as a composition of two operators (App.~\ref{apd:fm_non_linear_operators}):\vspace{5pt}
\[
    \nabla_{W_1}\mathcal{L}=\mathop{\mathbb{E}}\limits_{\mathbf{x}_0,\mathbf{x}_1}\left[\int_0^1 \underbrace{\rd{J} R(\bar{\mathbf{x}}(t))}_{A_1} \otimes \underbrace{\bar{\mathbf{x}}(t)}_{H_1} dt\right],
    \quad
    \nabla_{W_2}\mathcal{L}=\mathop{\mathbb{E}}\limits_{\mathbf{x}_0,\mathbf{x}_1}\left[\int_0^1 \underbrace{R(\bar{\mathbf{x}}(t))}_{A_2} \otimes \underbrace{\mathbf{h}_1(\bar{\mathbf{x}}(t))}_{H_2} dt\right]\vspace{7pt}
\]
where $\bar{\mathbf{x}}(t)$ is a point along the interpolant trajectory, $R(\bar{\mathbf{x}}(t))$ is the residual error per sample, $\mathbf{z}=W_1\mathbf{x}+\mathbf{b}$ is the network pre-activations, $\mathbf{h}$ is the network post-activations and $J = \text{diag}\left(\sigma'(\mathbf{z}_1)\right)W_2^\top$ is the network Jacobian. Similarly to the linear flow matching case, these are both of the form $A\circ H$, where the adjoint operator $A$ propagates errors backwards and the operator $H$ represents the state of the network. \rd{As before}, the $W_1$ gradient has an operator which \rd{only depends} on the data distribution.

\begin{figure}[!h]
    \centering
    \includegraphics[width=\linewidth]{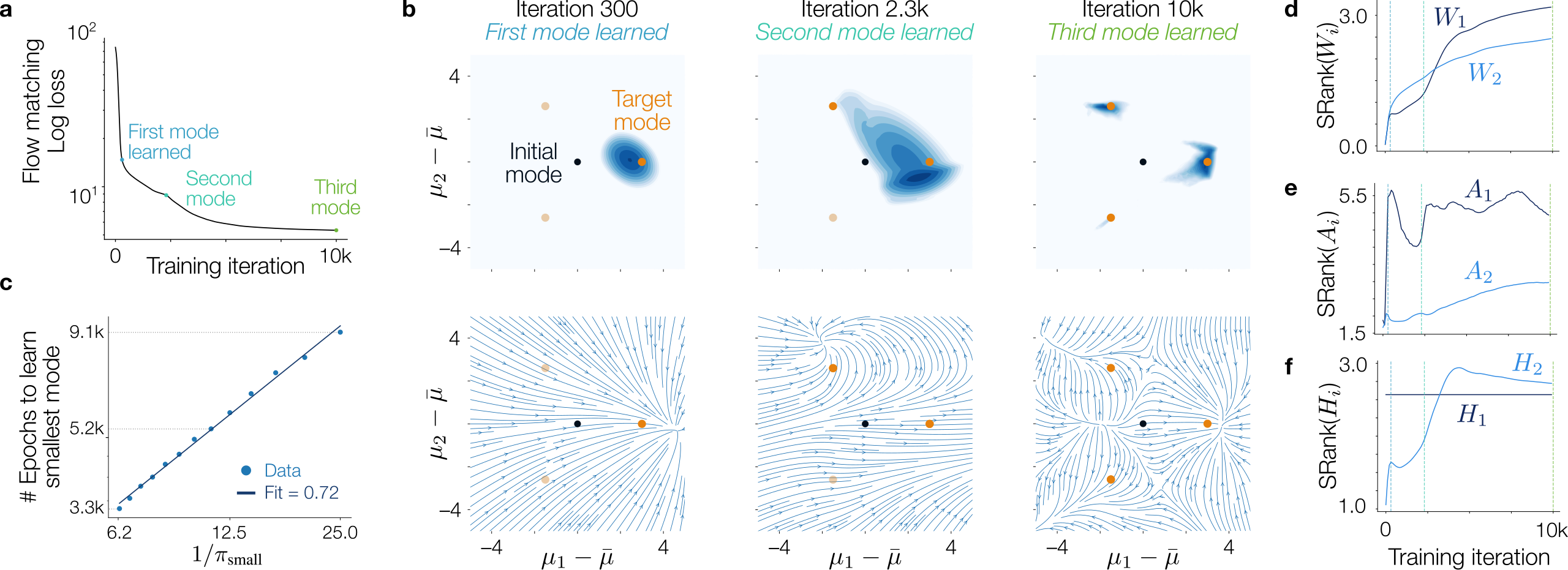}
    \caption{\textbf{Non-linear flow-matching models exhibit effective collapse due to low-rank biases.} \textbf{a.} Flow-matching loss over training displaying progressively longer learning time for each mode. \textbf{b.} (\textit{Top}) Generative log-density at three time points of training evaluated via the continuity equation (App. \ref{apd:nonlinear_fm}). The modes are progressively learned (\textit{Bottom}) Corresponding vector field, where additional fixed points are created with each learned mode. \textbf{c.} Learning time as a function of the inverse of the smallest mode, with $\pi_\text{min}\in[0.04, 0.16]$. Scaling exponent computed as $0.72\pm0.01$. \textbf{d.} Stable rank of the weights over training. The hidden layer weights $W_1$ show progressive increase in their rank with each learned mode. \textbf{e.} Adjoint (residual) operator rank for the two weight matrices. \textbf{f.} State operator rank for the two weight matrices.}
    \label{fig:nonlinear-fm}
\end{figure}

Similar to the linear model, we observed sequential learning dynamics corresponding to each mode being learned one-by-one (Fig.~\ref{fig:nonlinear-fm}a). The first inflection point in the loss corresponded to the probability density being transported to the dominant mode. After a few thousand training iterations a second inflection point occurred when the density expanded to cover the second strongest mode. By the point the loss saturated, the transported density covered all modes, with attracting dynamics surrounding each mode in the learned vector field (Fig.~\ref{fig:nonlinear-fm}b).

As in the linear model, we aimed to explain these phenomena by studying the rank of the weights and operators. First, the ranks of the weights correlated with the learning of each modes, with an increase in the rank of $W_1$ at each time a mode was learned (Fig.~\ref{fig:nonlinear-fm}d). Moreover, the first layer adjoint rank and second layer state rank also reflect mode learning, with sharp changes at each inflection point (Fig.~\ref{fig:nonlinear-fm}e-f). By systematically varying the imbalance between the strongest and weakest mode weightings, we observed the time to learn the smallest mode increasing \rd{according to a scaling law of $\tau_\text{learn}\propto\pi_\text{min}^{-0.72\pm0.01}$   (Fig.~\ref{fig:nonlinear-fm}c). The linear theory predicts an exponent of $1$, so the non-linear model learned faster than the linear prediction.} Finally, In appendix \ref{apd:nonlinear_fm} we also show that less tractable models trained on MNIST data display similar learning time scaling.

\vspace{3pt}

\textit{Together, these results show that the effective collapse learning dynamics observed in linear flow-matching models persist in non-linear settings, \rd{with} progressively delayed learning of modes.}

\section{Discussion}\label{sec:discussion}

\textbf{Summary.} Implicit biases are generally considered as creating beneficial biases that help representation learning in deep networks. Here, we examined how low-rank implicit biases can cause mode collapse in dynamical systems-based generative models. We derived the gradient of linear parameters as a composition of a state operator and an adjoint operator, and demonstrated how this view can explain collapse in various models. First, in linear continuous normalising flows under a variational objective, we used the adjoint view to show analytically that collapse corresponds to a stable attractor in the gradient flow. Interestingly, in this setting collapse can occur even when the target distribution is balanced, in contrast to prior studies where imbalance was necessary for collapse\cite{soletskyi_theoretical_2024,fogliani2026annealing}. Next, we showed in non-linear flow-matching models that an effective mode collapse could occur due to sequential mode learning.

\rd{\textbf{Future work.} This work focused on describing the learning mechanisms underlying mode collapse in generative models, but did not consider how to mitigate it. Since operator stable ranks are lower in collapsing models, it might be possible to use stable rank as a diagnostic to abort runs exhibiting low-rank operators early in training. To further mitigate collapse, it may be possible to use optimiser methods which re-weight the gradient singular value spectrum, such as Muon\cite{jordan2024muon}. Indeed, recent work has shown that Muon can remove saddle-to-saddle dynamics classically observed in gradient flow\cite{dragutinovic2026use}, and can be beneficial in settings where there is a class imbalance\cite{vasudeva2026muon}. While in our linear models the collapsed and optimal solutions could be understood in terms of the Schur decomposition, future works could use recent dynamical systems tools to study the geometry of collapsed representations in non-linear models\cite{pellegrino2026rnns,cayco2026geometry,karczewski2026spacetime}. Finally, while we mainly used summary statistics as a diagnostic tool to identify collapse, future work could give a full low-dimensional description of the gradient flow.} 

\textbf{Limitations.} This work relies on simplified analytical settings, including linear and controlled non-linear generative models with structured data distributions, which may not fully capture the complexity of large-scale modern architectures and real-world datasets. While the operator-based framework is generally applicable and reveals clear mechanisms for low-rank bias and collapse, the study of stable sub-optimal fixed points or the explicit derivation of saddles in the dynamics of learning of non-linear models remains a challenge of learning theory. Moreover, most of our results concern the population loss, and future works could study how finite sampling affects collapse.

\clearpage

\textbf{Funding statement.} This work was supported by the Agence Nationale de la Recherche (ANR-23-IACL-0008, ANR-26-CE23-2356-01, ANR-17-910 EURE-0017) and the European Research Council (Grant No. 101170541). The authors declare no competing interests.

\printbibliography

\clearpage

\appendix
\counterwithin{figure}{section}
\renewcommand{\thefigure}{\thesection\arabic{figure}}
\begin{refsection}

\section{Theory}\label{apd:adjoint_derivation}

%\begin{proof}[Proof of proposition \ref{thm:operator}]\label{proof:operator}

\subsection{Proof of theorem \ref{thm:operator}}\label{proof:operator}

In this section, we derive the exact gradient of a generalised loss function in a non-linear flow model, with respect to a linear weight parameter $W$. We consider a system governed by a Stratonovich stochastic differential equation (SDE), where $W$ affects the instantaneous loss, the drift, and the diffusion terms through potentially distinct feature inputs. This derivation encompasses the three types of losses commonly encountered in generative modelling with dynamical systems: variational objectives, flow and score matching.
%\rd{\begin{definition}

\ap{\textbf{Mathematical setting.}}
    Let $\mathbf{x}(t) \in \mathbb{R}^d$ denote the state of the system at time $t \in [0, T]$, evolving according to the \textit{Stratonovich SDE}:
\begin{equation}\label{eq:app_SDE_def}
    \mathbf{dx}(t) = \mathbf{f}_{\rd{\boldsymbol{\theta}}}(\mathbf{x}(t), t)dt + G_{\rd{\boldsymbol{\theta}}}(\mathbf{x}(t), t) \circ \mathbf{dB}, \qquad \mathbf{x}\rd{(0)} \sim p_0
\end{equation}
\rd{where $\mathbf{x}(t)$ is the state of the system, $\mathbf{dB}$ increments of a Brownian motion, $\mathbf{f}_{\boldsymbol{\theta}}:\mathbb{R}^d\times \mathbb{R}\rightarrow\mathbb{R}^d$ and  $G_{\boldsymbol{\theta}}:\mathbb{R}^d\times\mathbb{R}\rightarrow \mathbb{R}^{d\times d}$ the drift and diffusion components of the stochastic differential equation with parameters $\boldsymbol{\theta}\in \Theta$.} 
We use the Stratonovich integral ($\circ$) rather than the Itô integral, but we note that they can be interchanged. 

We evaluate the system via a continuous-time expected loss functional:
\begin{equation}\label{eq:app_loss_fn}
    \mathcal{L}(\rd{\boldsymbol{\theta}}) = \mathbb{E}_{\mathbf{x}_0,\mathbb{B}}\left[\int_0^T \ell(\mathbf{x}(t), t)dt \rd{\,+\, \Phi(\mathbf{x}(T))}\right]
\end{equation}
\rd{where $\ell:\mathbb{R}^d\times[0,T]\to\mathbb{R}$ is an instantaneous loss and $\Phi:\mathbb{R}^d\to\mathbb{R}$ a terminal loss which only depends on $\rd{\boldsymbol{\theta}}$ through $\mathbf{x}(T)$.} The core architectural assumption is that \rd{$\boldsymbol{\theta}$ contains }the parameter matrix $W \in \mathbb{R}^{\rd{k\times q}}$ which acts linearly. For the sake of generality, we assume that the weight parameter could potentially be part of any of the \rd{instantaneous} loss, drift and diffusion functions. In several cases the drift and diffusion models are parameterised by different networks, in which case any particular weight would only be part of one of them. In specific scenarios, such as physics-informed neural networks and some generative architectures, a network and its derivative can be parameterising the drift and diffusion terms at once, in which case $W$ appears in both terms.

To account for this, we assume $W$ may \ap{act on} distinct post-activation feature vectors $\mathbf{h}_\ell(t), \mathbf{h}_\mathbf{f}(t), \mathbf{h}_G(t) \in \mathbb{R}^q$ for the instantaneous loss $\ell$, drift, $\mathbf{f}$ and diffusion $G$ terms respectively. The pre-activations feature vectors are thus given by:
\[
    \mathbf{z}_\ell(t) = W\mathbf{h}_\ell(t), \quad \mathbf{z}_\mathbf{f}(t) = W\mathbf{h}_\mathbf{f}(t), \quad \mathbf{z}_G(t) = W\mathbf{h}_G(t)
\]
\rd{so that $\mathbf{z}_{(\cdot)}(t)\in\mathbb{R}^k$.}
\ap{While here we assume a shared parameter for the sake of generality, in many instances different neural network models will separately parameterise the loss, drift and diffusion terms, and therefore some of these partial derivatives will be zero.} Our goal is to derive $\nabla_W \mathcal{L}$ in terms of two operators, one which will be directly related to the post-activation vector, and one adjoint operator.

\textbf{Adjoints.} Because the state $\mathbf{x}(t)$ depends on $W$ through the integration of the SDE, we cannot differentiate $\mathcal{L}(W)$ by simply moving the gradient operator inside the expectation. Instead we must define the adjoint system.
\rd{\begin{definition}
    The adjoint $\mathbf{a}(t) \in \mathbb{R}^d$, represents how small changes in the trajectory of the dynamical system relate to small changes in the loss function:
\[
    \mathbf{a}(t) = \frac{\partial \mathcal{L}}{\partial \mathbf{x}(t)}
\]
where $\mathbf{x}(t)$ is the state of the system at time $t$.
\end{definition}}
\rd{\begin{lemma}\label{lemma:adjoint_solution}
    The adjoint is the solution to the following dynamical system:
    \[
    \mathbf{da}(t) = -\left(\nabla_{\mathbf{x}} \ell^\top + (\nabla_{\mathbf{x}} \mathbf{f})^\top \mathbf{a}(t)\right) dt - (\nabla_{\mathbf{x}} G)^\top \mathbf{a}(t) \circ \mathbf{dB}(t), \quad \mathbf{a}(T) = \mathbf{a}_T
    \]
    where $\mathbf{a}_T=\nabla_\mathbf{x}\Phi(\mathbf{x}(T))$, or $\mathbf{a}_T=0$ if there is no terminal loss.
\end{lemma}}
\ap{\begin{proof}
    For the case with a terminal loss and no instantaneous loss (i.e. $\ell=0$), the equation is the stochastic adjoint equation of Kidger et al. (2021, Eq. 6)\cite{kidger2021efficient}. To include a running loss, we can augment the state with $\tilde{\mathbf{x}}=(\mathbf{x}, y)$ where $dy=\ell(\mathbf{x}(t),t)dt$. Then the loss is terminal:
    \[
        \int_0^T \ell dt + \Phi(\mathbf{x}(T)) = y(T) + \Phi(\mathbf{x}(T))
    \]
    $\ell$, $\mathbf{f}$, and $G$ do not depend on $y$, so the adjoint of $y$ satisfies $da_y = 0$ with $a_y(T)=1$, which gives $a_y=1$. The $\mathbf{x}$-component of the augmented adjoint then contains the extra term $-(\nabla_\mathbf{x}\ell)^\top a_ydt=-\nabla_\mathbf{x}\ell^\top dt$ which gives the stated equation.
\end{proof}}

\ap{Importantly, the gradient of the model parameters can be written as an expectation of this adjoint, over initial states and Brownian paths.}

\begin{lemma}\label{lemma:parameter_gradient}
\ap{Dropping the dependence on $\mathbf{x}(t)$ for notational clarity, the gradient of model parameters can be written as:}
\[
    \nabla_\theta \mathcal{L} = \mathbb{E}_{\mathbf{x}_0,\mathbf{B}} \left[ \int_0^T \left( \nabla_\theta \ell + (\nabla_\theta \mathbf{f})^\top \mathbf{a}(t) \right) dt + \int_0^T (\nabla_\theta G)^\top \mathbf{a}(t) \circ d\mathbf{B}(t) \right]
\]
\end{lemma}
\ap{\begin{proof}
    Let $\boldsymbol{\eta}=\frac{\partial \mathbf{x}}{\partial \theta}$. Differentiating the SDE in Eq.~\ref{eq:app_SDE_def} gives:
    \[
        \mathbf{d}\boldsymbol{\eta}=(\nabla_\mathbf{x} \mathbf{f}\boldsymbol{\eta}+\nabla_\theta \mathbf{f})dt+(\nabla_\mathbf{x}G\boldsymbol{\eta} +\nabla_\theta G)\circ \mathbf{dB}
    \]
    By the Stratonovich product rule and Lemma~\ref{lemma:adjoint_solution}:
    \begin{align*}
        \mathbf{d}(\mathbf{a}^\top\boldsymbol{\eta}) &= \mathbf{da}^\top\boldsymbol{\eta} + \mathbf{a}^\top \mathbf{d}\boldsymbol{\eta} \\
        &= -\nabla_\mathbf{x}\ell \boldsymbol{\eta}dt + \mathbf{a}^\top \nabla_\theta \mathbf{f}dt + \mathbf{a}^\top \nabla_\theta G \circ \mathbf{dB}
    \end{align*}
    Integrate from $0$ to $T$. The left hand side is $\mathbf{a}(T)^\top \boldsymbol{\eta}(T) - \mathbf{a}(0)^\top \boldsymbol{\eta}(0)=\nabla_\mathbf{x}\Phi\boldsymbol{\eta}(T)$ since $\boldsymbol{\eta}(0)=0$. So:
    \[
        \int_0^T \nabla_\mathbf{x}\ell \boldsymbol{\eta} dt + \nabla_\mathbf{x}\Phi\boldsymbol{\eta}(T)= \int_0^T\mathbf{a}^\top\nabla_\theta \mathbf{f}dt + \int_0^T \mathbf{a}^\top \nabla_\theta G\circ \mathbf{dB}
    \]
    Now we recognise the pathwise gradient of the loss in Eq.~\ref{eq:app_loss_fn} is $\int_0^T (\nabla_\mathbf{x}\ell \boldsymbol{\eta} +\nabla_\theta \ell)dt+\nabla_\mathbf{x}\Phi \boldsymbol{\eta}(T)$. Substituting into the above, transposing, and taking expectations gives the result.
\end{proof}}

In our setting, the parameter of interest is the matrix $W$, which acts linearly, thus endowing the SDE of the gradient with a particular structure.

We evaluate the parameter Jacobians $\nabla_W \ell$, $\nabla_W \mathbf{f}$, and $\nabla_W G$ by applying the chain rule through the linear projections $\mathbf{z}_\ell, \mathbf{z}_\mathbf{f}$, and $\mathbf{z}_G$. First, the terms of the loss which do not depend on $\mathbf{x}$ have gradients independent of the adjoint. Since $\mathbf{z}_\ell(t) = W\mathbf{h}_\ell(t)$, differentiating with respect to the matrix $W$ gives an outer product:
\[
    \nabla_W \ell = \mathbf{a}_\ell(t) \otimes \mathbf{h}_\ell(t)
\]
where $\mathbf{a}_\ell(t) = \partial_{\mathbf{z}_\ell} \ell(t)$. The sensitivity of the drift and diffusion terms are mediated by the adjoint state $\mathbf{a}(t)$. We define the drift error vector $\mathbf{a}_\mathbf{f}(t) \in \mathbb{R}^k$ by pulling the adjoint back through the drift function:
\[
    \mathbf{a}_\mathbf{f}(t) = \left(\nabla_{\mathbf{z}_\mathbf{f}} \mathbf{f} \right)^\top \mathbf{a}(t)
\]
Applying the chain rule through $\mathbf{z}_\mathbf{f}(t) = W\mathbf{h}_\mathbf{f}(t)$ gives:
\[
    (\nabla_W \mathbf{f})^\top \mathbf{a}(t) = \mathbf{a}_\mathbf{f}(t) \otimes \mathbf{h}_\mathbf{f}(t)
\]
Similarly:
\[
    \mathbf{a}_G(t) = \left( \nabla_{\mathbf{z}_G} G(\mathbf{x}(t), \mathbf{z}_G(t), t) \right)^\top \mathbf{a}(t) \qquad (\nabla_W G)^\top \mathbf{a}(t) = \mathbf{a}_G(t) \otimes \mathbf{h}_G(t)
\]
\rd{\begin{corollary}\label{crly:gradient_outer_prod}
    For a weight $W$ that enters only through $\mathbf{z}_{(\cdot)}=W\mathbf{h}_{(\cdot)}$,
    \[
        \nabla_W \mathcal{L} = \mathbb{E} \left[ \int_0^T \Big( \mathbf{a}_\ell(t) \otimes \mathbf{h}_\ell(t) + \mathbf{a}_\mathbf{f}(t)  \otimes \mathbf{h}_\mathbf{f}(t) \Big) dt + \int_0^T \left(\mathbf{a}_G(t)  \otimes  \mathbf{h}_G(t)\right) \circ \mathbf{dB}(t) \right]
    \]
    where $\mathbf{a}_\ell=\nabla_{\mathbf{z}_\ell}\ell$, $\mathbf{a}_\mathbf{f}=(\nabla_{\mathbf{z}_\mathbf{f}}\mathbf{f})^\top\mathbf{a}$ and $\mathbf{a}_G=(\nabla_{\mathbf{z}_G} G)^\top \mathbf{a}$. 
\end{corollary}}
\ap{\begin{proof}
    Since $\mathbf{z}_{(\cdot)}=W\mathbf{h}_{(\cdot)}$, the chain rule gives:
    \[
        \frac{\partial }{\partial W_{ij}}= \left( \frac{\partial }{\partial \mathbf{z}_{(\cdot)}} \right)_i \mathbf{h}_{(\cdot),j}
    \]
    Applying this to each term,
    \[
        \nabla_W \ell=\mathbf{a}_\ell \otimes \mathbf{h}_\ell, \quad \left( \nabla_W \mathbf{f} \right)^\top \mathbf{a} = \mathbf{a}_\mathbf{f} \otimes \mathbf{h}_f, \quad \left( \nabla_W G \right)^\top \mathbf{a} = \mathbf{a}_G \otimes \mathbf{h}_G
    \]
    Then substituting these into Lemma~\ref{lemma:parameter_gradient} gives the result.
\end{proof}}
In particular, \ap{because the contribution of the loss, drift and diffusion terms to the gradient are additive,} the \rd{per sample} instantaneous gradient is \rd{at most} rank $3$. \ap{As mentioned earlier, in some instances the weight parameter will in fact appear in only one of these objects (e.g., the drift and loss terms in the CNF and FM models in the main text), and therefore the instantaneous gradient may be of even lower rank.}

\phantomsection
\textbf{Gradient as an operator.} \label{apd:full_gradient_AoH}
\rd{From corollary~\ref{crly:gradient_outer_prod},} the gradient can be rewritten as in Theorem~\ref{thm:operator}:
    \[
        \nabla_W \mathcal{L} = A_\ell \circ H_\ell + A_\mathbf{f} \circ H_\mathbf{f} + A_G \circ H_G
    \]
    where $A_{(\cdot)}$ and $H_{(\cdot)}$ are linear operators:
    \[
        A_{(\cdot)} : L^2(\mathbb{R}) \rightarrow \mathbb{R}^k , \qquad A_{(\cdot)}(f) = \mathbb{E}_{\mathbf{x}_0,\rd{\mathbf{B}}}\left[ \int_0^T \mathbf{a}_{(\cdot)}(t) f(t) dt \right]
    \]
    \rd{\[
        H_{(\cdot)}:\mathbb{R}^q\rightarrow L^2(\mathbb{R}), \qquad (H_{(\cdot)}(\mathbf{v}))(t)=\mathbf{h}_{(\cdot)}(t) \cdot \mathbf{v}
    \]}
    
First, we briefly describe how more specific losses commonly used in generative modelling fall within this framework, before describing general spectral properties of the gradient in terms of these operators.

\textbf{Examples.}
Now we briefly provide a more concrete example how two different types of losses affect the same generative model. We consider an SDE whose drift is parameterised by a one-hidden-layer neural network:
\[
    \mathbf{dx} = W_2\phi(W_1\mathbf{x})dt + \sigma\, \mathbf{dB}
\]
For the sake of illustration we consider two simplified losses:
\[
    \text{Drift fitting:}\quad \mathcal{L}^\mathrm{d.f.} = \mathbb{E}_{\mathbf{x}_0, \mathbb{B}}\int_0^T \lVert W_2\phi(W_1\mathbf{\bar x}(t)) - \mathbf{u}^*(\mathbf{\bar x}(t))\rVert^2dt
\]
\[
    \text{State fitting:} \quad \mathcal{L}^\mathrm{s.f.} = \mathbb{E}_{\mathbf{x}_0, \mathbb{B}} \int_0^T \lVert\mathbf{x}(t) - \mathbf{x}^*(t)\rVert^2dt    
\]
where $\mathbf{u}^*: \mathbb{R}^d\rightarrow \mathbb{R}^d$ and $\mathbf{x}^*:\mathbb{R}\rightarrow \mathbb{R}^d$ are the target vector fields and states respectively. These two cases are particularly illustrative as generative flow and score matching are special cases of the former, while variational inference and FFJORD on CNFs are closely related to the latter. Because the drift fitting case is completely independent of $\mathbf{x}(t)$ the solution of the SDE, this is essentially a regression problem. We focus on the first layer's gradient, since it encodes the learned non-linear features. In that case:
\[
    \mathbf{h}_\mathbf{f} = \mathbf{h}_G = 0 \qquad \mathbf{h}_\ell = \mathbf{\bar x}(t) \qquad \mathbf{a}_\ell = \mathrm{diag}(\phi'(W_1 \mathbf{\bar x}(t)))W_2^\top \mathbf{r}(\mathbf{\bar x}(t))
\]
where $\mathbf{r}(\mathbf{\bar x}(t)) = W_2\phi(W_1\mathbf{\bar x}(t)) - \mathbf{u}^*(\mathbf{\bar x}(t))$ is the residual,
so that the gradient is given by:
\[
    \nabla_{W_1} \mathcal{L}^\mathrm{d.f.} = \mathbb{E}\left[ \int_0^T \left(\mathrm{diag}(\phi'(W_1 \mathbf{\bar x}(t)))W_2^\top \mathbf{r}(\mathbf{\bar x}(t))\right) \otimes \mathbf{\bar x}(t)dt\right] = A_\ell \circ H_\ell  
\]
In the state fitting case:
\[
    \mathbf{h}_\ell = \mathbf{h}_G = 0, \qquad \mathbf{h}_\mathbf{f} = \mathbf{x}(t) \qquad \mathbf{a}_\mathbf{f} = \mathrm{diag}(\phi'(W_1 \mathbf{x}(t)))W_2^\top \mathbf{a}(t)
\]
where $\mathbf{a}(t)$ is the state adjoint, which evolves as:
\[
    \mathbf{da} = -\left( \mathbf{x}(t) - \mathbf{x}^*(t) + W_1^\top \Big( \mathrm{diag}(\phi'(W_1\mathbf{x}(t))) \big(W_2^\top \mathbf{a}(t)\big) \Big) \right) dt
\]
The gradient is given by:
\[
    \nabla_{W_1} \mathcal{L}^\mathrm{s.f.} = \mathbb{E}\left[ \int_0^T \left(\mathrm{diag}(\phi'(W_1 \mathbf{x}(t)))W_2^\top \mathbf{a}(t)\right) \otimes \mathbf{x}(t)dt\right] = A_\mathbf{f} \circ H_\mathbf{f}  
\]
we will return to closely related examples in our variational inference and flow matching sections.

\subsection{Proof of theorem \ref{thm:srank_bound}}\label{proof:srank_bound}

We've demonstrated above that, even with highly non-linear models with general losses, the gradient of weight parameters can be written as a sum of compositions of linear operators. 
\[
    \nabla_W \mathcal{L} = A_\ell \circ H_\ell + A_\mathbf{f} \circ H_\mathbf{f} + A_G \circ H_G
\]
where $A_{(\cdot)}$ are \textit{adjoint} operators and $H_{(\cdot)}$ \textit{state} operators. In case of linear models as used in the main text, the state operator is the state of the system itself. The adjoint operator is directly dependent on the residual. As these operators are linear, and because they are maps from or to a finite dimensional space (the state-space) they are compact. They therefore admit singular value decompositions.
\rd{\begin{proposition}
    The adjoint and state operators admit the singular value decompositions:
    \[
        A_{(\cdot)} = U^a \circ S^a \circ V^a \qquad H_{(\cdot)} = U^h \circ S^h \circ V^h
    \]
    where:
    \[
        U^a\in\mathbb{R}^{k \times k}, \quad S^a\in\mathbb{R}^{k}, \quad V^a:L^2(\mathbb{R})\rightarrow \mathbb{R}^k
    \]
    and:
    \[
        U^h:\mathbb{R}^q\rightarrow L^2(\mathbb{R}), \quad S^h\in\mathbb{R}^{q}, \quad V^h\in\mathbb{R}^{q\times q}
    \]
    such that their composition is a linear map $\nabla_W \mathcal{L} \in \mathbb{R}^{k \times q}$:
    \[
        \nabla_W \mathcal{L}:\mathbb{R}^q
        \overset{V^h}{\longrightarrow}
        \mathbb{R}^q
        \overset{S^h}{\longrightarrow}
        \mathbb{R}^q
        \overset{U^h}{\longrightarrow}
        L^2(\mathbb{R})
        \overset{V^a}{\longrightarrow}
        \mathbb{R}^k
        \overset{S^a}{\longrightarrow}
        \mathbb{R}^k
        \overset{U^a}{\longrightarrow}
        \mathbb{R}^k
    \]
\end{proposition}}

\textbf{Stable rank.} 
A natural consequence of this structure in the gradient is that if either operator is low rank then the gradient will be low rank as it is generally true that:
\[
    \mathrm{Rank}(X \circ Y) \leq \mathrm{min}(\mathrm{Rank}(X), \mathrm{Rank}(Y))
\]
This can for example occur when the model weights are already low rank. In both examples above, we had that the adjoint of the $W_1$ matrix had the form:
\[
    \mathbf{a}_{(\cdot)}(t) = \mathrm{diag}(\phi'(W_1 \mathbf{x}(t)))W_2^\top (\,\cdot\,)
\]
Thus the rank of the operator is determined by the $W_2$ weights. For linear networks, as considered in the main text, a consequence of this is that the rank of the gradient of model weights bound each other:
\[
    \mathrm{Rank}(A_{(\cdot)}) \leq \mathrm{Rank}(W_2)\quad \Longrightarrow \quad \mathrm{Rank}(\nabla_{W_1}\mathcal{L}) \leq \mathrm{Rank}(W_2)
\]
this can cause an entrainment of the weight learning dynamics towards lower-rank solutions.

In practice weights are never exactly low-rank. However, we can characterise their \textit{effective} rank via the stable rank.
\rd{\begin{definition}
    The \textit{stable rank} is given by:
    \[
    \mathrm{SRank}(W) = \frac{\lVert W \rVert_*}{\lVert W \rVert_2} \quad = \sum_{i=1}^{\min(m, k)} \frac{s_i}{s_1}
    \]
\end{definition}}

Other notions of effective rank include the participation ratio or the energy rank\cite{thibeault2024low}. In general, the stable rank of the product of two matrices can be expressed in terms of the stable rank of each individual matrices. 
\rd{\begin{definition}
    Define the Schatten norm of a linear operator:
    \[
        \lVert X \rVert_{\mathrm{Sc}(r)} = {\mathrm{Tr}(S^r)}^{1/r}
    \]
    where $S$ is that of the SVD of $X$. 
\end{definition}}
\begin{proof}[\ap{Proof of theorem \ref{thm:srank_bound}}]
For all $r^{-1}\leq p^{-1}+q^{-1}$ the Schatten norm of the composition of linear operators satisfies:
\[
    \lVert X \circ Y \rVert_{\mathrm{Sc}(r)} \leq \lVert X \rVert_{\mathrm{Sc}(q)} \lVert Y \rVert_{\mathrm{Sc}(p)}
\]
In particular, taking $r=1$ and $p=\infty$ (resp. $1$) and $q=1$ (resp. $\infty$), yields the bound:
\[
    \lVert X \circ Y \rVert_* \leq \min \{\lVert Y \rVert_2 \lVert X \rVert_*, \lVert X \rVert_2 \lVert Y \rVert_*\}
\]
multiplying and dividing the first and second argument of the $\min$ by the spectral norm of $Y$ and $X$ respectively gives:
\[
     \lVert X \circ Y \rVert_* \leq \lVert Y \rVert_2 \lVert X \rVert_2\min \{\mathrm{SRank}(X), \mathrm{SRank}(Y)\}
\]
or:
\[
    \mathrm{SRank}(X\circ Y) \leq \frac{\lVert X \rVert_2 \lVert Y \rVert_2}{\lVert X\circ Y\rVert_2}\min \{\mathrm{SRank}(X), \mathrm{SRank}(Y)\}
\]
\rd{Thus, writing for the weight gradient we arrive at Theorem~\ref{thm:srank_bound}:}
\[
    \operatorname{SRank}(\nabla_W \mathcal{L}) \leq \frac{\lVert A \rVert_2 \lVert H \rVert_2}{\lVert A\circ H\rVert_2}\min \left\{\operatorname{SRank}(A), \operatorname{SRank}(H)\right\}.
\]
\end{proof}
Thus, in the examples above, the rank of the gradient of the hidden layer weights are bounded by the rank of the decoder:
\[
    \operatorname{SRank}(\nabla_{W_1} \mathcal{L}) \leq \gamma\,\mathrm{SRank}(W_2)
\]
where $\gamma$ depends on the spectral norms of the operators.

\rd{The stable rank bound in Theorem~\ref{thm:srank_bound} depends on the prefactor:
\[
    \gamma(A,H)=\frac{\lVert A \rVert_2 \lVert H\rVert_2}{\lVert A\circ H\rVert_2}
\]
where $\lVert \cdot \rVert_2$ denotes the spectral norm. Therefore, the bound is only tight when $\gamma(A,H)\approx1$. This occurs when the left singular vectors of $H$ and the right singular vectors of $A$ are aligned.} 

\rd{Using the fact that:
\[
    \lVert A\circ H\rVert_2 \ge \sigma_1^A\sigma_1^H\lvert \langle \mathbf{v}_1^A, \mathbf{u}_1^H\rangle\rvert
\]
where $\sigma_1^{(\cdot)}$ are the top singular values, and $\mathbf{v}_1^{(\cdot)}$ and $\mathbf{u}_1^{(\cdot)}$ are the right and left singular vectors of an operator $(\cdot)$ respectively, the prefactor is bounded by:
\[
    \gamma(A,H)\le\lvert \langle \mathbf{v}_1^A,\mathbf{u}_1^H\rangle\rvert^{-1}
\]}

% =================================
\section{Collapse in continuous normalising flow}\label{apd:adjoint_CNFs}

\subsection{Linear CNF adjoint}

\begin{proof}[\rd{Proof of theorem \ref{thm:CNF_gradient}}]\label{proof:CNF_gradient} 
\rd{In the notation of App.~\ref{apd:adjoint_derivation}, $\Phi=-\log p^*$ and the log-determinant gives the running loss $\ell=-\text{Tr}W$, contributing $\nabla_W \ell=-I$ which is constant and full rank. Here we consider just the drift operators.} Following from theorem~\ref{thm:operator}, the gradient flow can be written as
\[
    \nabla_W \mathcal{L} = A \circ H = \int_0^1 \mathbf{a}(t) \otimes \mathbf{x}(t) dt
\]
As previously shown, for a linear model the adjoint has dynamics
\[
    \frac{d}{dt}\mathbf{a}(t) = -W^\top \mathbf{a}(t)
\]
This admits the solution
\[
    \mathbf{a}(t)=\exp(W^\top (1-t)) \mathbf{a}(1)
\]
where the terminal condition is given by $\mathbf{a}(1) =-\nabla_{\mathbf{x}} \log p^*(\mathbf{x}(1))$. The state of the system at time $t$ for the linear model is
\[
\mathbf{x}(t)=\exp(Wt)\mathbf{x}_0
\]
Putting these together, we obtain
\[A\circ H = \int_0^1 \exp(W^\top (1-t)) (\mathbf{a}(1)\otimes \mathbf{x}(0)) \exp(W^\top t) dt.\]
\end{proof}

\rd{\subsection{Model details}\label{apd:CNF_train_details}}
\rd{\textbf{Training.}} In section~\ref{sec:CNF_results} we used a linear continuous normalising flow model. The model is trained to minimize the reverse-KL divergence:
\[
    D_{\text{KL}}(p_1(\mathbf{x})||p^*(\mathbf{x}))=\mathbb{E}_{\mathbf{x}\sim p_1}\left[\log p_1(\mathbf{x}) - \log p^*(\mathbf{x})\right]
\]
Following previous work\cite{grathwohl2018scalable}, we define a parameterisation of the velocity field, $\frac{d\mathbf{z}}{dt}=f(\mathbf{z}(t), t)$. We can rewrite this using the learned distribution at time $t$, integrated from $t\in[0,1]$:
\begin{align*}
    &\mathbb{E}_{\mathbf{z}(0)\sim p_0}\left[\log p_0(\mathbf{z}(0)) + \int_0^1 \frac{d}{dt}\log p_t(\mathbf{z}(t))dt - \log p^*(\mathbf{z}(1)) \right] \\
    &\stackrel{(1)}{=} \mathbb{E}_{\mathbf{z}(0)\sim p_0}\left[\log p_0(\mathbf{z}(0)) - \int_0^1 \nabla \cdot f(\mathbf{z}(t), t)dt - \log p^*(\mathbf{z}(1)) \right]
\end{align*}

where $p_t$ and $\mathbf{z}(t)$ are the learned distribution and pushed forward initial sample at time $t$, and in $(1)$ we used the instantaneous change of variables formula\cite{chen_neural_2018}. To compute gradients with respect to the loss, we define an augmented ODE:
\[
    \frac{d\mathbf{x}(t)}{dt}=\begin{bmatrix}
        f(\mathbf{z}(t)) \\
        -\nabla\cdot f(\mathbf{z}(t), t)
    \end{bmatrix} \qquad \mathbf{x}(0)=\begin{bmatrix}
        \mathbf{z}(0) \\
        0
    \end{bmatrix}
\]
and integrate forward in time using a differentiable ODE solver to obtain $\mathbf{z}(1)$ and the accumulated log-density of the learned distribution. Backpropagating through the ODE solver obtains the gradients.

We initialised weights $w_{ij}\sim\mathcal{N}(0,\sigma_{\text{init}}/\sqrt{n})$ where the dimension $n=2m$ and $m$ is the number of initial and target modes. For simulations we used a differentiable ODE solver with the Tsit5 method and  adaptive step size. To optimize the system we used SGD in Optax. The hyperparameters are as follows:

\begin{center}
    \begin{tabular}{cccccccc}
    \hline
    $m$ & $\sigma_{\text{init}}$ & lr & Iterations & Batch size & Optimizer & ODE solver & PID; rtol, atol \\ \hline
    3-10  & $1e^{-4}$   & $1e^{-3}$ & 30k & 20k & SGD & Tsit5 & $1e^{-3}$, $1e^{-6}$    \\ \hline
    \end{tabular}
\end{center}

\rd{\textbf{Mode geometry.} We construct our mode setup by centring $m$ orthogonal modes of norm $R$. Let $\mathbf{e}_1, \mathbf{e}_2, \dots, \mathbf{e}_m$ and $\mathbf{f}_1, \mathbf{f}_2,\dots,\mathbf{f}_m$ be orthonormal vectors. The initial and target modes are
\[
    \boldsymbol{\mu}_i=R(\mathbf{e}_i - \bar{\mathbf{e}}) , \qquad \boldsymbol{\mu}_i^*=R(\mathbf{f}_i-\bar{\mathbf{f}})
\]
where $\bar{\mathbf{e}}=\frac{1}{m}\sum_i \mathbf{e}_i$ and $\bar{\mathbf{f}}=\frac{1}{m}\sum_i \mathbf{f}_i$. They satisfy:
\begin{align*}
    \boldsymbol{\mu}_i \cdot \boldsymbol{\mu}_j&=R^2 (\mathbf{e}_i - \bar{\mathbf{e}})\cdot (\mathbf{e}_j - \bar{\mathbf{e}}) \\
    &= R^2 \left( \delta_{ij} - \frac{1}{m} \right)
\end{align*}
and additionally
\[
    \sum_i \boldsymbol{\mu}_i =0, \qquad \lVert \boldsymbol{\mu}_i\rVert = R\sqrt{\frac{m-1}{m}}, \qquad \boldsymbol{\mu}_i \cdot \boldsymbol{\mu}_j^*=0 
\]
and likewise for $\boldsymbol{\mu}_i^*$.}

\rd{The initial means span an $(m-1)$-dimensional subspace $S$ and target means span an $(m-1)$-dimensional subspace $T\perp S$. The remaining $d-2(m-1)$ directions $\perp$ contain no modes.}

\rd{\subsection{Analyses}\label{apd:CNF_analyses}}
Here, we briefly describe the numerical analyses performed on the linear CNF model.

\textbf{Identifying collapsed modes.} 
\rd{In order to see early in training which solution a model will express at convergence, it is useful to consider scalar quantities which appear in the loss and which can tell us about the state of the model. The loss contains
\[
    \log p^*(\mathbf{x}(1))=\log \sum_j \mathcal{N}(\mathbf{x}(1); \boldsymbol{\mu}^*_j, \Sigma)
\]
which gives rise to terms of the form
\begin{align*}
    \lVert \mathbf{x}(1)-\boldsymbol{\mu}_j^*\rVert^2&=\lVert e^W (\boldsymbol{\mu}_i + \mathbf{z}) - \boldsymbol{\mu}_j^* \rVert^2 \\
    &= \lVert e^W \boldsymbol{\mu}_i -\boldsymbol{\mu_j^*\rVert}^2 +\text{terms dependent on }\mathbf{z}
\end{align*}
for a particular initial mode $i$, and for noise sample $\mathbf{z}\sim\mathcal{N}(\mathbf{0}, I)$. Normalising this term gives us our summary statistic:
\[
    D_{ij}=\frac{\lVert e^W \boldsymbol{\mu}_i - \boldsymbol{\mu}_j^* \rVert^2}{R^2}
\]
This quantifies the distance between pushed forward initial modes and target modes, and provides a useful diagnostic for where initial modes are routed by the model. We can equivalently look at
\[
    s_{ij}=\frac{(e^W \boldsymbol{\mu}_i)\cdot\boldsymbol{\mu}_j^*}{R^2}=-\frac{1}{2}D_{ij}+ c_i
\]
where $c_i$ is constant across row $i$.
Since at convergence, each initial mode is transported to a target mode, the final state of the statistic is a matrix with one entry per row significantly smaller than the rest (indicating which target each initial mode is sent to). This allows us to compute the assignment of initial modes: 
\[
    j_0 (i) = \text{argmax}_j s_{ij}
\]
which gives an $m$-dimensional vector of integers corresponding to the index of the target mode that each initial mode is assigned. For the perfect solution, this is a permutation of $j_0(i)=(1, \dots, m)$, whereas collapsed solutions will have repeated indices.}

\rd{By computing $j_0(i)$ at different points across training, we can see which assignment pattern is expressed by the weights. If at train time $\tau_{\text{lock}}$ the $D$ matrix predicts the solution at convergence, then $j_0(i)$ should be constant from time $\tau_{\text{lock}}$.}

\begin{figure}[h]
    \centering
    \includegraphics[width=0.7\linewidth]{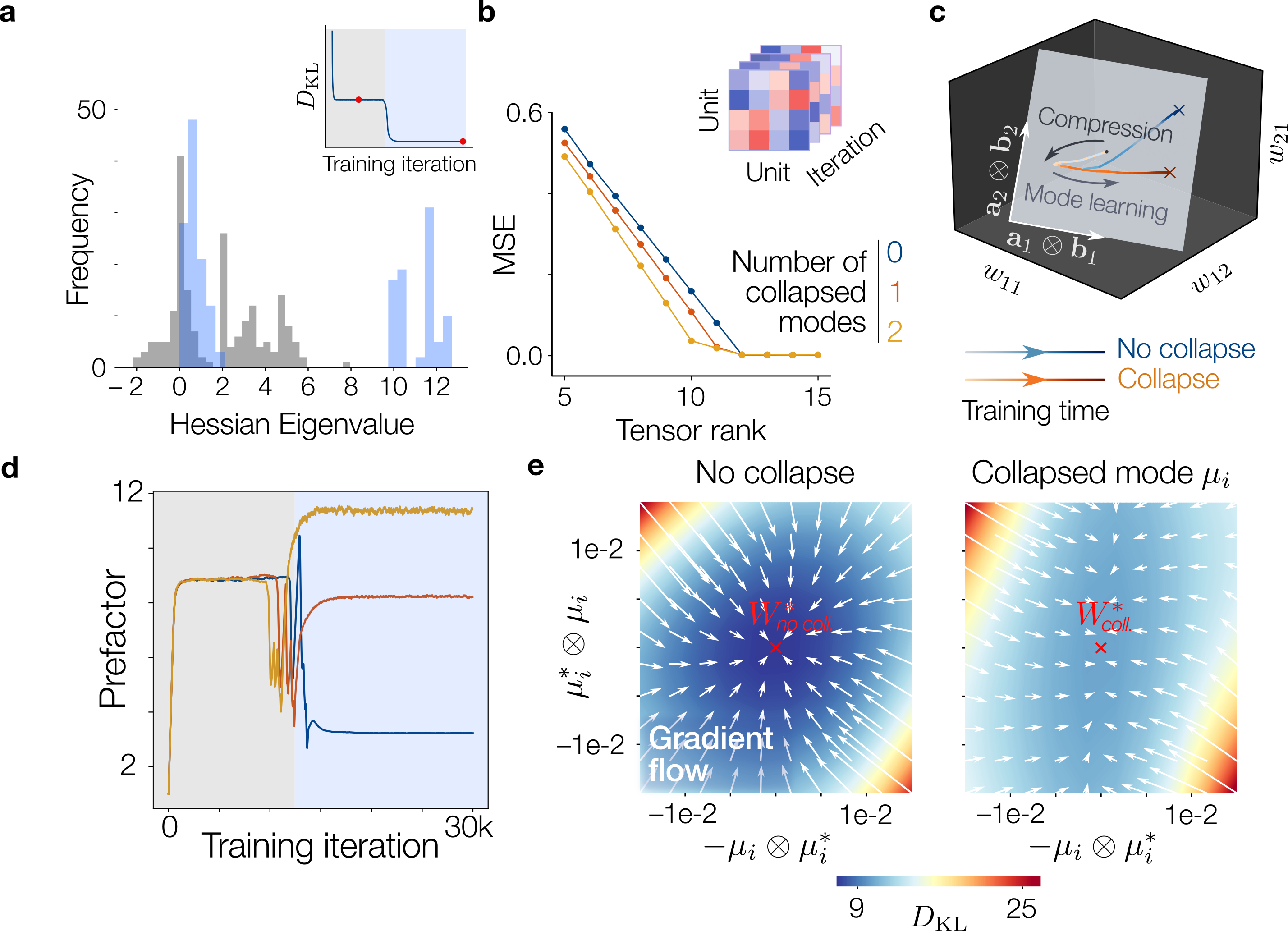}
    \caption{\textbf{Low-rank learning dynamics of CNF model.} \textbf{a.} Empirical distributions of the eigenvalues of the Hessian, evaluated at the compression slow-point (grey) and at the end of training (blue) for a non-collapsed model. \textbf{b.} MSE of a rank-$k$ \rd{CP} decomposition of the weight tensor\cite{pellegrino2023low} formed by stacking $W$ over training. \textbf{c.} Low-tensor-rank subspace of the weight space highlighting how the trajectory of the collapsed and non-collapsed models split during the mode alignment phase.
    \rd{\textbf{d.} Prefactor during learning for collapsed and non-collapsed models. 
    \textbf{e.} Visualisation of gradient flow near collapsed and non-collapsed fixed points. }
}
    \label{sup_fig:cnf}
\end{figure}

\textbf{Visualising the gradient flow near fixed points.} In figure~\ref{sup_fig:cnf}e we plot the gradient flow around fixed points of the weights for a collapsing and non-collapsing model. We do this by perturbing the converged weights $W^*$ in directions $-\boldsymbol{\mu}_i\otimes\boldsymbol{\mu}_j^*$ and $\boldsymbol{\mu}_j^*\otimes\boldsymbol{\mu}_i$, corresponding to directions away from and towards the optimal solution respectively, for a given mode pairing $i,j$. In the collapsed model the mode indices $i$ and $j$ are chosen such that they correspond to a collapsing initial mode and an unassigned-to target mode. In the non-collapsed model they correspond to an initial mode and its assigned target. The gradient is then computed via auto-differentiation in JAX on a grid of weights along these two basis vectors and then visualised as a flow field.

\textbf{Eigenvalues of the Hessian.} Over training, Hessians are computed with respect to the vectorised weights: $\frac{\partial^2 \mathcal{L}}{\partial \text{vec}(W_{ij})\text{vec}(W_{kl})}\in\mathbb{R}^{n^2\times n^2}$, using auto-differentiation through the ODE solver in JAX. This yields $n^2$ eigenvalues with $n^2$ corresponding eigenvectors of shape $n\times n$. The histogram of eigenvalues at the two fixed points of a non-collapsed model is plotted in figure~\ref{sup_fig:cnf}a.

\textbf{Tensor decomposition of weights.} In figure~\ref{sup_fig:cnf}b-c we decomposed the weights over training into a sum of rank-1 tensor products $\boldsymbol{W} = \textstyle\sum_{i=1}^r \mathbf{a}_i\otimes \mathbf{b}_i\otimes \mathbf{c}_i + \boldsymbol{\chi}$ where $\mathbf{a}_i\in\mathbb{R}^{k}$ tracks the evolution over training and $\mathbf{b}_i,\mathbf{c}_i\in\mathbb{R}^{n\times n}$ are spatial components. We performed the decomposition using tensorly parafac. We then choose two spatial component products of the non-collapsed weights, $\mathbf{b}_i\otimes\mathbf{c}_i$, to project the temporal components onto for both models. The projection components are chosen such that one of them collapses in the collapsed model, giving a separation between the two trajectories. We then plot the projected components over training in the 2-d subspace of the full $n\times n$ weight space.

\textbf{Computing the operator rank.} In figure~\ref{fig:f2}h-i we plot the stable rank of the \rd{state and adjoint} operators over training. From theorem~\ref{thm:operator} and theorem~\ref{thm:CNF_gradient}, we can write the gradient as a composition of two linear operators, $\nabla_W\mathcal{L}=A\circ H$, where:
\[
    A=-\mathbb{E}_{\mathbf{x}_0}\left[\int_0^T e^{(1-t)W^\top}\nabla_{\mathbf{x}_1}\log p^*(\mathbf{x}_1) dt\right], \qquad H=\mathbb{E}_{\mathbf{x}_0}\left[\int_0^1 e^{tW}\mathbf{x}_0dt\right]
\]
where $\mathbf{x}_1=e^W\mathbf{x}_0$. There is a fundamental theorem in functional analysis that says that these operators admit a singular value decomposition. Since $rank(A) \leq min(dim((R^n)^*)=n$ and $dim(L^2)=\infty$, it has at most $n$ non-zero singular values. By discretising the integral with $q$ initial samples and $T$ time points we can write $A \in (R^n)^* \otimes R^{T\times q}$, on which we can perform singular value decomposition and compute the stable rank for each training time. \rd{From this, it is trivial to compute the prefactor (Fig.~\ref{sup_fig:cnf}d).}

\rd{\subsection{CNF rank bound prefactor at initialisation}\label{apd:cnf_prefactor_proof}}
\rd{\begin{theorem}\label{thm:prefactor_initialisation}
    In the CNF model defined in Section~\ref{sec:CNF_results}, with initial modes $\boldsymbol{\mu}_i$ and target modes $\boldsymbol{\mu}_i^*$ satisfying $\boldsymbol{\mu}_i\cdot\boldsymbol{\mu}_j^*=0$, $\boldsymbol{\mu}_i\cdot\boldsymbol{\mu}_k=\boldsymbol{\mu}_i^*\cdot\boldsymbol{\mu}_k^*$ and with weights initialised to be zero and with well separated modes such that $R\gg\sigma$, the prefactor can be evaluated as exactly \[\gamma(A,H) = \frac{\lVert A\rVert_2 \lVert H\rVert_2}{\lVert A \circ H \rVert_2} =  1\]
\end{theorem}}

\rd{\textit{Proof of Theorem~\ref{thm:prefactor_initialisation}.}} 
\rd{We consider the well separated regime where mode radius dominates over noise, $R\gg \sigma$, and sampling noise is negligible. In this case, we have:
\[
    \mathbf{a}^{(i)}(1)=\boldsymbol{\mu}_i-\boldsymbol{\mu}_i^*  \equiv \Delta_i, \qquad \mathbf{x}^{(i)}(0)=\boldsymbol{\mu}_i
\]
which yield the discretised operators:
\[
    H=\frac{1}{\sqrt{m}} \begin{pmatrix} \mu_1^\top \\ \vdots \\ \mu_m^\top\end{pmatrix} \in \mathbb{R}^{m\times d}, \qquad A=\frac{1}{\sqrt{m}}(\Delta_1, \dots, \Delta_m)\in \mathbb{R}^{d\times m}
\]
To obtain the operator spectral norm we need to compute SVD on the covariance type matrices $HH^\top$ and $A^\top A$. To compute this, we note the following:}

\rd{\begin{lemma}\label{lemma:operator_eigvals}
    For any matrix B, $\lVert B \rVert^2_2 = \lambda_\text{max}(BB^\top)=\lambda_\text{max}(B^\top B)$ because $B^\top B$ and $B B^\top$ have the same non-zero eigenvalues.
\end{lemma}}

\rd{Now we note that
\[
    (HH^\top)_{ij}=\frac{1}{m}\mu_i\cdot\mu_j\equiv \frac{1}{m}G_{ij}, \qquad (A^\top A)_{ij} = \frac{1}{m}\Delta_i \cdot \Delta_j = \frac{2}{m}G_{ij}
\]
where both are proportional to $G$ due to the initial and target modes having the same geometry in orthogonal subspaces. This means that by lemma~\ref{lemma:operator_eigvals}:
\[
    \lVert H\rVert^2_2 = \lambda_\text{max}(H^\top H)=\frac{\lambda_\text{max}(G)}{m}, \qquad \lVert A \rVert^2_2=\lambda_\text{max}(AA^\top)=\frac{2\lambda_\text{max}(G)}{m}
\]
Now we need to find the eigenvalues of the composition of the operators, $A\circ H=AH$. The covariance-style matrix is $(AH)^\top (AH)=H^\top A^\top A H=P^\top P$ where $P=\left( \frac{2}{m} G \right)^{\frac{1}{2}} H$. We then construct the matrix: 
\[
    PP^\top = \left( \frac{2}{m} G \right)^{\frac{1}{2}} H H^\top \left( \frac{2}{m}G \right)^{\frac{1}{2}}=\frac{2}{m^2}G^{\frac{1}{2}} G G^{\frac{1}{2}}=\frac{2}{m^2}G^2
\]
From lemma~\ref{lemma:operator_eigvals} and the fact that for a symmetric positive semi-definite matrix $m$, $\lambda_\text{max}(m^2)=\lambda_\text{max}(m)^2$, we have that 
\[
    \lVert A\circ H \rVert^2_2 = \frac{2\lambda_\text{max}(G^2)}{m^2}=\frac{2\lambda_\text{max}(G)^2}{m^2}
\]
Now by defining $\lambda=\lambda_\text{max}(G)$, we can compute the prefactor:
\[
    \gamma(A,H)=\frac{\lVert A\rVert_2 \lVert H\rVert_2}{\lVert A \circ H \rVert_2}=\frac{\sqrt{2\lambda/m}\sqrt{\lambda/m}}{\sqrt{2\lambda^2/m^2}}=1
\]
}

\subsection{Fixed points of the gradient flow}\label{apd:CNF_fixed_points}
\rd{In this section we show that three important points during the training of the linear CNF are fixed points of the gradient flow, namely the optimal solution, collapsed solutions and the compression saddle point. We use an ansatz for each point derived from empirically observed weight matrix structures, showing that they satisfy the fixed point equation.}

\subsubsection{Setup}
\rd{\textbf{Form of the gradient.} First, note that the gradient has the form}:
\[
    \nabla_W \mathcal{L} = \nabla_{W} D_{\mathrm{KL}}(p_1, p^*) = \nabla_{W} \mathbb{E}\left[ \log p_1(\mathbf{x}(1)) -\log p^*(\mathbf{x}(1))\right]
\]
By the continuity equations\rd{, for the linear CNF}:
\[
    \log p(\mathbf{x}(1)) = \log p(\mathbf{x}(0)) - \int_0^1 \mathrm{div}(W\mathbf{x}(t))dt =  \log p(\mathbf{x}(0)) - \mathrm{Tr}(W)
\]
Hence, because $\mathbf{x}(0)$ is independent of $W$:
\[  
    \nabla_W \mathcal{L} = -\nabla_W \mathbb{E}\left[ \log p^*(\mathbf{x}_1) \right] - I
\]
\rd{Using Theorem~\ref{thm:CNF_gradient}, }the gradient is given by:
\[
    \nabla_W \mathcal{L} = \mathbb{E}\left[ \int_0^1 e^{W^\top (1-t)} \mathbf{a}(1) \otimes \mathbf{x}(0) e^{W^\top t}dt \right] - I
\]
where $\mathbf{a}(1) = -\nabla_{\mathbf{x}} \log p^*(\mathbf{x}(1))$ is the negative score of the data distribution.
We are interested in the fixed points $\nabla_W \mathcal{L} = 0$, that is:
\[
    I = \mathbb{E}\left[ \int_0^1 e^{W^\top (1-t)} \mathbf{a}(1) \otimes \mathbf{x}(0) e^{W^\top t}dt \right]
\]
\textbf{Data assumptions.} \rd{Our initial samples are drawn from $p_0=\sum_iw_i\mathcal{N}(\boldsymbol{\mu}_i,\sigma^2 I_d)$, and so we can write $\mathbf{x}(0)=\boldsymbol{\mu}_i+\sigma\mathbf{z}$ where $\mathbf{z}\sim \mathcal{N}(\mathbf{0}, I_d)$.
We use the centred mode setup detailed in App.~\ref{apd:CNF_train_details}, with 
\[
     \boldsymbol{\mu}_i\cdot \boldsymbol{\mu}_j=R^2\left( \delta_{ij} - \frac{1}{m} \right), \qquad\sum_i\boldsymbol{\mu}_i =0, \qquad \lVert\boldsymbol{\mu}_i\rVert = R\sqrt{\frac{m-1}{m}} 
\]
and likewise for $\boldsymbol{\mu}_i^*$, with $\boldsymbol{\mu}_i\cdot\boldsymbol{\mu}_j^*=0$. These are modes of a Gaussian mixture with $m$ modes, covariances $\Sigma_i$. The initial means span an $(m-1)$-dimensional subspace $S$ and target means span an $(m-1)$-dimensional subspace $T\perp S$. The remaining $d-2(m-1)$ directions $\perp$ contain no modes. Let $Q_S$ and $Q_T$ be $d\times (m-1)$ matrices whose columns form an orthonormal basis on $S$ and $T$ respectively. $Q_\perp$ is a $d\times d-2(m-1)$ matrix that spans the remaining dimensions. We define modes in their subspace basis as:
\[
    \boldsymbol{\alpha}_i=Q_S^\top \boldsymbol{\mu}_i, \quad \boldsymbol{\beta}_i=Q_T^\top \boldsymbol{\mu}_i^*, \quad \sum_i \boldsymbol{\alpha}_i=0,\quad \sum_i \boldsymbol{\alpha}_i\otimes \boldsymbol{\alpha}_i=R^2 I_{m-1}
\]
We will also deal with terms of the form $\sum_i \boldsymbol{\mu}_i \otimes \boldsymbol{\mu}_i$. Expanding gives: 
\begin{align*}
    \sum_i \boldsymbol{\mu}_i \otimes \boldsymbol{\mu}_i &= R^2\sum_i (\mathbf{e}_i - \bar{\mathbf{e}}) \otimes (\mathbf{e}_i - \bar{\mathbf{e}}) \\
    &= R^2 \left( \sum_i \mathbf{e}_i \otimes \mathbf{e}_i - m \bar{\mathbf{e}}\otimes \bar{\mathbf{e}} \right)
\end{align*}
The first term is a projection onto the span of $\{\mathbf{e}_1, \dots \mathbf{e}_m\}$ and the second is a projection onto mean. Together they project onto the subspace orthogonal to the mean directions. This is exactly the subspace $S$, which has projector $Q_S Q_S^\top$. Thus we have that: 
\[
    \sum_i \boldsymbol{\mu}_i \otimes \boldsymbol{\mu}_i=R^2 Q_S Q_S^\top
\]}

For a Gaussian Mixture, the negative score can be derived explicitly:
\[
    \mathbf{a}(1)=-\nabla_{\mathbf{x}} \log p^*(\mathbf{x}(1)) = -\nabla_{\mathbf{x}} \log \sum_{i=1}^m w_i \mathcal{N}(\mathbf{x}(1) ;  \boldsymbol{\mu}_i\rd{^*}, \Sigma_i)
\]
We make the additional assumptions that $w_i=1/m$ and $\Sigma_i=\sigma^2 I$. Then:
\[
    \nabla_{\mathbf{x}} \log p^*(\mathbf{x}(1)) = \sum_{i=1}^m \gamma_i(\mathbf{x}(1))(\boldsymbol{\mu}_i^*-\mathbf{x}(1))
\]
where:
\[
    \gamma_i(\mathbf{x}(1)) = \frac{\mathcal{N}(\mathbf{x}(1);\boldsymbol{\mu}_i^*, \sigma^2 I)}{\sum_{j=1}^m \mathcal{N}(\mathbf{x}(1);\boldsymbol{\mu}_j^*, \sigma^2 I)}
\]

\rd{\textbf{Fixed point equation for converged weights.} We assume that at convergence (either for the optimal or collapsed solutions),} $\mathbf{x}(1)$ should be near one of the data modes (otherwise the $D_\mathrm{KL}$ will be large), and if \rd{$R\gg\sigma$}, the density of all other Gaussians will be vanishingly small. \rd{Thus the softmax weighting in the score becomes a hard assignment}:
\[
    \gamma_i(\mathbf{x}(1)) \approx \delta_{i,j_0(i)}
\]
\rd{where $j_0(i)$ gives the index of the target mode that initial mode $i$ is transported to by the final weights}. Thus we focus on the case where:
\[
    \nabla_{\mathbf{x}} \log p^*(\mathbf{x}(1)) \approx \frac{1}{\sigma^2}(\boldsymbol{\mu}^*_{j_0(i)} - \mathbf{x}(1))
\]
\rd{for a sample transported from the $i^\text{th}$ initial mode}. Thus, the \rd{fixed point equation} can be written as:
\rd{\[
    -\sigma^2 I = \mathbb{E}\left[ \int_0^1 e^{W^\top (1-t)} (\boldsymbol{\mu}_{j_0(i)}^* - e^{W}\mathbf{x}(0))\otimes \mathbf{x}(0) e^{W^\top t}dt \right]
\]
We note the identity: 
\[
    I=\int_0^1 e^{W^\top (1-t)}e^{-W^\top} e^{W^\top t}dt
\]
which allows us to write the equation as:
\[
    \mathcal{F}\left( \mathbb{E}\left[ \left(\boldsymbol{\mu}^*_{j_0(i)}-e^W \mathbf{x}(0)\right)\otimes \mathbf{x}(0) \right] + \sigma^2 e^{-W^\top} \right)=0
\]
where $\mathcal{F}(B)=\int_0^1 e^{W^\top (1-t)} B e^{W^\top t}dt$ is the Fréchet derivative of the matrix exponential in a direction $B$. The fixed point equation is then: 
\begin{align*}
    \sigma^2 e^{-W^\top} &= e^W \mathbb{E}\left[ \mathbf{x}(0) \otimes \mathbf{x}(0) \right] - \mathbb{E}\left[\boldsymbol{\mu}_{j_0(i)}^* \otimes \mathbf{x}(0)  \right] \\
    \sigma^2 e^{-W^\top} &= \frac{e^W}{m} \sum_i \mathbb{E}\left[ (\boldsymbol{\mu}_i +\sigma\mathbf{z})\otimes(\boldsymbol{\mu}_i +\sigma\mathbf{z}) \right]  - \frac{1}{m}\sum_i \mathbb{E} \left[ \boldsymbol{\mu}_{j_0(i)}^*\otimes (\boldsymbol{\mu}_i + \sigma\mathbf{z}) \right]\\
    \sigma^2 e^{-W^\top} &= e^W \left(\frac{1}{m}\sum_i\boldsymbol{\mu}_i \otimes \boldsymbol{\mu}_i + \sigma^2\mathbb{E}[\mathbf{z}\otimes \mathbf{z}]\right) -\frac{1}{m} \sum_i \boldsymbol{\mu}_{j_0(i)}^* \otimes \boldsymbol{\mu}_i  \\
    \sigma^2 e^{-W^\top} &= e^W\left(\frac{R^2}{m}Q_S Q_S^\top + \sigma^2 I_d  \right) - \frac{1}{m}\sum_i \boldsymbol{\mu}_{j_0(i)}^* \otimes \boldsymbol{\mu}_i \\
    \sigma^2 e^{-W^\top} &= e^W \Sigma_0 - \mathcal{T}
\end{align*}
where $\Sigma_0$ is the second moment of the initial distribution and $\mathcal{T}$ is a mapping which sends initial subspace directions to target subspace directions.}

\subsubsection{Optimal solution}
\rd{Due to the symmetry of the initial and target mode geometry, an easy guess as to a non-collapsing solution is the rotation of the initial to target means. Write $\mathbf{s}_k$ and $\mathbf{t}_{k}$ as the $k^\text{th}$ column of $Q_S$ and $Q_T$ respectively, then such a solution can be written as:
\[
    W = U L U^\top
\]
where $U=[\mathbf{s}_1, \mathbf{t}_1,\dots,\mathbf{s}_{m-1}, \mathbf{t}_{m-1}, Q_\perp]$ (the columns of $Q_\perp$ are arbitrary since they have the same zero eigenvalue)}, and $L$ is given by:
\[
L=\text{diag}\left( B_1, \dots, B_{m-1}, \mathbf{0} \right)
, \qquad
B_i = \frac{\pi}{2}\begin{pmatrix} 0 & -1 \\ 1 & 0 \end{pmatrix}
\]
\rd{Then, 
\[
    e^{B_i} = \begin{pmatrix} 0 & -1 \\ 1 & 0 \end{pmatrix}
\]
and so $e^W \mathbf{s}_{i}= \mathbf{t}_{i}$ and $e^W \mathbf{t}_{i} = - \mathbf{s}_{i}$. We now take the RHS of the fixed point equation from above, noting that $e^W \boldsymbol{\mu}_i  = \boldsymbol{\mu}_{j_0(i)}^*$:
\begin{align*}
    \text{RHS}: \quad&e^W \left( \frac{R^2}{m} Q_S Q_S^\top + \sigma^2 I_d \right) - \frac{1}{m}\sum_i \boldsymbol{\mu}_{j_0(i)}^* \otimes \boldsymbol{\mu}_i \\
    &=e^W \left( \frac{R^2}{m} Q_S Q_S^\top + \sigma^2 I_d \right) - \frac{e^W}{m} \sum_i \boldsymbol{\mu}_i \otimes \boldsymbol{\mu}_i \\
    &= e^W \left( \frac{R^2}{m} Q_S Q_S^\top + \sigma^2 I_d \right) - e^W \frac{R^2}{m} Q_S Q_S^\top \\
    &= \sigma^2 e^W \\
    &=\sigma^2 e^{-W^\top}= \text{LHS}
\end{align*}
where the last equality is true since $L$ is antisymmetric. Thus, the fixed point equation is satisfied for the perfect solution.}

\subsubsection{Collapsed solution}
\rd{Empirically, we observe collapsed solutions to occur when $n$ pairs of initial modes are each sent to one target mode, with one target mode ignored/unassigned per collapsing pair. The remaining initial modes are sent to one target per mode each, as in the perfect solution. For simplicity, we will consider the case of 1 collapsed pair, with $\boldsymbol{\mu}_1$ and $\boldsymbol{\mu}_2$ both being sent to $\boldsymbol{\mu}_1^*$, and $\boldsymbol{\mu}_2^*$ being ignored. Without loss of generality we assume that the $i^\text{th}$ non-collapsing initial mode is sent to the $i^\text{th}$ target mode. The extension to $n>1$ is discussed at the end.}

\rd{We begin by evaluating the terms in the fixed point equation for this case:
\[
     \Sigma_0 = \frac{R^2}{m}Q_S Q_S^\top + \sigma^2 I_d \qquad \mathcal{T}= \frac{1}{m}\left( \boldsymbol{\mu}_1^* \otimes(\boldsymbol{\mu}_1 + \boldsymbol{\mu}_2) + \sum_{i\ge 3}\boldsymbol{\mu}_i^* \otimes \boldsymbol{\mu}_i \right)
\]
We notice that $\Sigma_0$ acts on subspace $S$ as $(\frac{R^2}{m}+\sigma^2)I_S$ and subspace $T$ as $\sigma^2I_T$, and thus does not mix the subspaces. On the other hand, $\mathcal{T}$ mixes the subspaces by sending $\boldsymbol{\mu}_1$ and $\boldsymbol{\mu}_2$ to $\boldsymbol{\mu}_1^*$, and all other initial modes go to their respective target modes. This structure informs our guess of basis vectors:
\[
    \mathbf{p}=\frac{\boldsymbol{\mu}_1+\boldsymbol{\mu}_2}{\lVert \boldsymbol{\mu}_1+\boldsymbol{\mu}_2\rVert}, \quad \mathbf{q}=\frac{\boldsymbol{\mu}_1-\boldsymbol{\mu}_2}{\lVert \boldsymbol{\mu}_1-\boldsymbol{\mu}_2\rVert}, \quad \mathbf{s}_1, \dots\mathbf{s}_{m-3} \perp \mathbf{p}, \mathbf{q}
\]
where $\lVert \boldsymbol{\mu}_1+\boldsymbol{\mu}_2\rVert^2=\frac{2R^2(m-2)}{m}$, $\lVert \boldsymbol{\mu}_1 - \boldsymbol{\mu}_2\rVert^2=2R^2$, and $\mathbf{p}\perp \mathbf{q}$. $\mathbf{s}_k$ are any orthonormal basis of the directions orthogonal to $\boldsymbol{\mu}_1$ and $\boldsymbol{\mu}_2$.
Together, these form an orthonormal basis on the initial subspace $S$. We guess that the weights act on these directions independently. To solve for the form of the weights, we will make an ansatz and \textbf{solve the fixed point equation in each of the basis directions separately}, yielding scalar equations. Now we compute the image of each vector under the transformation $\mathcal{T}$.}

\rd{\textbf{Shared direction $\mathbf{p}$ under $\mathcal{T}$.} We use the fact that $\boldsymbol{\mu}_i\cdot (\boldsymbol{\mu}_1+\boldsymbol{\mu}_2)= R^2 \frac{m-2}{m}$ for $i=1,2$ and $-\frac{2R^2}{m}$ for $i\ge3$ to compute:
\begin{align*}
    \mathcal{T}(\boldsymbol{\mu}_1 + \boldsymbol{\mu}_2)&=\frac{2R^2}{m^2}\left[ (m-2)\boldsymbol{\mu}_1^* - \sum_{i\ge 3} \boldsymbol{\mu}_i^* \right] \\&= \frac{2R^2(m-1)}{m^2} \left[ \boldsymbol{\mu}_1^*+\frac{1}{m-1}\boldsymbol{\mu}_2^* \right]
\end{align*} 
Normalising gives $\mathcal{T}\mathbf{p}=\tau \mathbf{u}$ where
\[
    \mathbf{u}=\frac{\boldsymbol{\mu}_1^*+\frac{1}{m-1}\boldsymbol{\mu}_2^*}{\lVert \boldsymbol{\mu}_1^*+\frac{1}{m-1}\boldsymbol{\mu}_2^*\rVert}, \qquad \tau=\frac{R^2}{m} \sqrt{\frac{2(m-1)}{m}}
\]
and $\lVert \boldsymbol{\mu}_1^* + \frac{1}{m-1}\boldsymbol{\mu}_2^* \rVert^2=R^2 \frac{m-2}{m-1}$.
Note that $\mathbf{u}$ corresponds to $\boldsymbol{\mu}_1^*$ with its component in the $\boldsymbol{\mu}_2^*$ direction removed since $\mathbf{u}\cdot \boldsymbol{\mu}_2^*=0$.}

\rd{\textbf{Difference direction $\mathbf{q}$ under $\mathcal{T}$.} Since $\mathbf{q}$ is orthogonal to $\mathbf{p}$ and all other $\boldsymbol{\mu}_{i\ge3}$, we compute that:
\[
    \mathcal{T}(\boldsymbol{\mu}_1-\boldsymbol{\mu}_2)=0
\]
and so no target pulls on the difference direction (i.e. it stays in subspace $S$).}

\rd{\textbf{Basis directions under $\mathcal{T}$.} Since we have chosen $\mathbf{s}_k \cdot (\boldsymbol{\mu}_1+\boldsymbol{\mu}_2)=0$, the transformation becomes:
\[
    \mathcal{T}\mathbf{s}_k=\frac{1}{m}\sum_{i\ge3} (\boldsymbol{\mu}_i^* \otimes\boldsymbol{\mu}_i) \mathbf{s}_k
\]
Since in these directions modes are sent to a single target, $\mathcal{T}$ (in these directions) is identical to in the optimal solution case. We note that $e^{W_\text{opt}}\boldsymbol{\mu}_i=\boldsymbol{\mu}_i^*$, we can write:
\[
    \mathcal{T} \mathbf{s}_k = e^{W_\text{opt}}\frac{1}{m} \left(\sum_i \boldsymbol{\mu}_i \otimes \boldsymbol{\mu}_i\right) \mathbf{s}_k=\frac{R^2}{m}e^{W_\text{opt}} \mathbf{s}_k
\]
and so $\mathcal{T}\mathbf{s}_k=\frac{R^2}{m} \mathbf{t}_k$ where $\mathbf{t}_k$ form an orthogonal basis on $T$.}

\rd{\textbf{Ansatz.} We showed that $\{\mathbf{p}, \mathbf{q}, \mathbf{s}_1,\dots \mathbf{s}_{m-3}\}$ form an orthonormal basis on $S$. Under the transformation $\mathcal{T}$, $\mathbf{p}$ is sent to $\mathbf{u}\propto \boldsymbol{\mu}_1^*+\frac{1}{m-1}\boldsymbol{\mu}_2^*$ and $\mathbf{s}_k$ is sent to $\mathbf{t}_k$, both of which lie in the target subspace. To complete the basis in the target subspace, we define $\mathbf{v}=\boldsymbol{\mu}_2^* /\lVert\boldsymbol{\mu}_2^* \rVert$ giving $\{\mathbf{u}, \mathbf{v}, \mathbf{t}_1,\dots,\mathbf{t}_{m-3}\}$.}

\rd{We choose the ansatz $e^W=Ue^L U^\top$ where $U=[\mathbf{s}_1, \mathbf{t}_1, \dots, \mathbf{s}_{m-3}, \mathbf{t}_{m-3}, \mathbf{p}, \mathbf{u}, \mathbf{q}, \mathbf{v},Q_\perp]$ is constructed from our orthonormal bases with $Q_\perp$ forming a basis on the directions perpendicular to $S$ and $T$. We also choose: 
\[
    e^L=\text{diag}(E_1, \dots, E_{m-3}, E_{\text{sh}}, b, 1,\mathbf{1}), \quad E_i=\begin{pmatrix} 0 & -1\\1&0 \end{pmatrix}, \quad E_\text{sh}=\begin{pmatrix} 0 & -\beta \\ a & 0 \end{pmatrix}
\]
where $a,b,\beta$ are unknowns to solve for. Intuitively, $E_i$ rotate the non-collapsing initial modes to their assigned targets, and $E_\text{sh}$ performs some mapping from shared initial direction $\mathbf{p}$ to the collapsed-to target direction $\mathbf{u}$, $\mathbf{q}$ is scaled by $b$ (since it stays in the initial subspace), and the other directions are ignored (scaled by $1$).} 

\rd{The ansatz in the directions mapping $\mathbf{s}_k\to \mathbf{t}_k$ automatically satisfies the fixed point equation since it is identical to the optimal solution. The directions scaled by $1$ satisfy $\sigma^2I=\sigma^2I$ since $\mathbf{v},Q_\perp\perp S$. For the other directions, we must solve the fixed point equation.}

\rd{To compute $\beta$ and $a$, we solve the fixed point equation in the $\mathbf{p}$ and $\mathbf{u}$ subspaces, representing the mapping of two initial modes to the same target. For generic vectors $\mathbf{s}$ and $\mathbf{t}$ in the initial and target subspaces we have $E_\text{sh}\mathbf{t}=-\beta\mathbf{s}$, $E_{\text{sh}}^{-\top}\mathbf{t}=-\beta^{-1}\mathbf{s}$, $E_\text{sh}\mathbf{s}=a\mathbf{t}$, and $E^{-\top}_\text{sh}\mathbf{s}=a^{-1}\mathbf{t}$. Acting on the $\mathbf{u}$ subspace, the equation becomes:
\begin{align*}
    \sigma^2 E_{\text{sh}}^{-\top} \mathbf{t} &=E_\text{sh}\Sigma_0 \mathbf{t} - \mathcal{T}\mathbf{t} \\
    -\sigma^2 \beta^{-1} \mathbf{s}&= -\sigma^2 \beta \mathbf{s} - 0 \\
    \implies \beta&=1
\end{align*}
Now acting on the $\mathbf{p}$ subspace:
\begin{align*}
    &\sigma^2 E_\text{sh}^{-\top} \mathbf{s} = E_\text{sh}\Sigma_0 \mathbf{s} - \mathcal{T}\mathbf{s} \\ 
    &\sigma^2 a^{-1} \mathbf{t}=\left( \frac{R^2}{m} +\sigma^2 \right) a\mathbf{t} - \tau \mathbf{t}\\
    &\implies a=\frac{\tau+\sqrt{\tau^2 +4\sigma^2\left(\sigma^2 +\frac{R^2}{m}\right)}}{2\left( \sigma^2 + \frac{R^2}{m} \right)}
\end{align*}
with $\tau$ defined above.}

\rd{Finally, to compute $b$ we solve the same equation for the difference direction $\mathbf{q}$:
\begin{align*}
    &\sigma^2 e^{-W^\top}\mathbf{q} = e^W\Sigma_0 \mathbf{q} - \mathcal{T}\mathbf{q} \\
    & \sigma^2 b^{-1} = \left( \sigma^2 + \frac{R^2}{m} \right) b \\
    & \implies b=\left( \frac{\sigma^2}{\sigma^2+\frac{R^2}{m}} \right)^{\frac{1}{2}}
\end{align*}}

\rd{In sum, we constructed an ansatz for a solution with $1$ collapsing pair and solved the fixed point equation to find the specific form of the solution, thus proving it is a fixed point. Table~\ref{tab:compression_vals} shows the values of $a,b, \beta$ obtained from the derived expressions and from trained weights, showing a close match. The derivation can be extended to $n>1$ by constructing initial subspace bases as shared directions $\mathbf{p}_{ab}=\frac{\boldsymbol{\mu}_a+\boldsymbol{\mu}_b}{\lVert \boldsymbol{\mu}_a+\boldsymbol{\mu}_b\rVert}$ and difference directions $\mathbf{q}_{ab}=\frac{\boldsymbol{\mu}_a-\boldsymbol{\mu}_b}{\lVert \boldsymbol{\mu}_a - \boldsymbol{\mu}_b\rVert}$ for pairs of collapsing initials $(\boldsymbol{\mu}_a,\boldsymbol{\mu}_b)$. Then the target subspace bases are $\mathbf{u}_{\alpha, \beta}=\frac{\boldsymbol{\mu}_\alpha^* +\frac{1}{m-1}\boldsymbol{\mu}_\beta^*}{\lVert \boldsymbol{\mu}_\alpha^* +\frac{1}{m-1}\boldsymbol{\mu}_\beta^* \rVert}$ for collapsed-to target mode $\boldsymbol{\mu}_\alpha^*$ and ignored target mode $\boldsymbol{\mu}_\beta^*$ with scaling $\tau_{\alpha\beta}$ under the transformation $\mathcal{T}$.}

\begin{table}[htbp]
    \centering
    \begin{tabular}{l*{3}{r}}
    \toprule
     & Analytic & Computed\\
    \midrule
    $a$ & 1.2750 & 1.2748  \\
    $b$ & 0.2558 & 0.2556  \\
    $\beta$ & 1.0000 & 0.9846 \\
    \bottomrule
    \end{tabular}
    \caption{Analytically derived vs numerically computed constants used in collapsed weight ansatz.}
    \label{tab:compression_vals}
\end{table}

\subsubsection{Saddle/slow compression point}
\rd{During training, we observe the weight trajectories pass through an extended period where they act independently on the initial and target subspaces. We call this the "compression phase". From observations, we use the following ansatz for the weight structure:
\[
    W=ULU^\top
\]
where $U=[Q_S, Q_T, Q_\perp]$ with $Q_T$ and $Q_S$ the $d\times(m-1)$ matrices whose orthonormal columns span $\{\mu_j^*\}$ and $\{\mu_i\}$ respectively, and $Q_\perp$ an orthonormal basis of the $\perp$ subspace. We also define:
\[
    L=\text{diag}(\lambda_S I_{m-1}, \lambda_T I_{m-1}, \mathbf{0})
\]
where $\lambda_T>0$ and $\lambda_S<0$ are the scaling factors in the target and initials subspace respectively. In the rest of the section we will solve the fixed point equation for the values of these scaling factors.}

\rd{\textbf{The fixed point equation.} Since at the compression point, the initial modes are not assigned to their targets, we cannot use the hard assignment limit to simplify the fixed point equation as in the previous sections. We must use the full terminal adjoint which has the form $\mathbf{a}(\mathbf{x}(1))=\frac{1}{\sigma^2}\left( \mathbf{x}(1)-\bar{\textbf{m}}(\mathbf{x}(1)) \right)$ where:
\[
    \bar{\textbf{m}}(\mathbf{x})=\sum_{j=1}^m \gamma_j(\mathbf{x})\boldsymbol{\mu}_j^*, \qquad \gamma_j(\mathbf{x})= \frac{\mathcal{N}(\mathbf{x}; \boldsymbol{\mu}_j^*,\sigma^2I_d)}{\sum_k \mathcal{N}(\mathbf{x}; \boldsymbol{\mu}_k^*,\sigma^2I_d)}
\]}

\rd{Similarly to before, we can write the full fixed point equation with the form of the score substituted in (defining $\mathbf{x}_t \equiv \mathbf{x}(t)$):
\begin{align*}
    -\sigma^2I_d&=\mathbb{E}\left[ \int_0^1 e^{W^\top(1-t)} \left( \mathbf{m}(\mathbf{x}_1) - e^W\mathbf{x}_0 \right)\otimes \mathbf{x}_0 e^{W^\top t} dt\right] \\
    -\sigma^2 I_d&= \mathbb{E}\left[ \int_0^1 Ue^{L(1-t)}U^\top \left( \mathbf{m}(\mathbf{x}_1) - Ue^L U^\top\mathbf{x}_0 \right) \otimes \mathbf{x}_0  Ue^{Lt} U^\top dt\right] \\ 
    -\sigma^2 U^\top U&=-\sigma^2 I_d =\mathbb{E}\left[ \int_0^1 e^{L(1-t)}U^\top \left( \mathbf{m}(\mathbf{x}_1) - Ue^L U^\top\mathbf{x}_0 \right) \otimes \mathbf{x}_0  Ue^{Lt} dt\right]
\end{align*}
where in the second line we use the ansatz and in the third line we use the fact that $U$ is orthogonal. Now we can write $\mathbf{x}_0=\boldsymbol{\mu}_i+\sigma\mathbf{z}$ where $\mathbf{z}\sim \mathcal{N}(\mathbf{0}, I_d)$. Splitting the expectation into a sum over initial modes we get:
\[
    -\sigma^2 I_d = \frac{1}{m}\sum_i \mathbb{E}\left[ \int_0^1 e^{L(1-t)}U^\top \left( \mathbf{m}(\mathbf{x}_1) - Ue^L U^\top (\boldsymbol{\mu}_i+\sigma\mathbf{z}) \right) \otimes (\boldsymbol{\mu}_i+\sigma\mathbf{z})  Ue^{Lt} dt\right]
\]
We also introduce coordinates in the rotated basis, such that
\[
    U^\top \boldsymbol{\mu}_i=\begin{pmatrix} \boldsymbol{\alpha}_i\\0\\0 \end{pmatrix},\quad U^\top \boldsymbol{\mu}_i^* =\begin{pmatrix} 0\\ \boldsymbol{\beta}_i\\0 \end{pmatrix}, \quad U^\top \mathbf{z}=\tilde{\mathbf{z}}=\begin{pmatrix} \mathbf{z}_S \\ \mathbf{z}_T\\\ \mathbf{z}_\perp \end{pmatrix}, \quad U^\top \mathbf{m}(\mathbf{x}_1)=\begin{pmatrix} 0\\ \mathbf{m}_T \\0 \end{pmatrix}
\]
where $\tilde{\mathbf{z}} \sim\mathcal{N}(\mathbf{0},I_d)$ since $U$ is orthogonal and $\mathbf{m}_T=\sum_j \gamma_j(\mathbf{x}_1) \boldsymbol{\beta}_j$. We also note that $e^{cL}=\text{diag}(e^{c\lambda_S} I_{m-1}, e^{c\lambda_T}I_{m-1}, 1)$. Now the fixed point equation becomes:
\begin{align*}
    -\sigma^2 I_d &= \frac{1}{m}\sum_i \mathbb{E}\left[ \int_0^1 e^{L(1-t)} \left( U^\top \mathbf{m}(\mathbf{x}_1) - e^L U^\top (\boldsymbol{\mu}_i+\sigma\mathbf{z}) \right) \otimes \left( U^\top (\boldsymbol{\mu}_i +\sigma\mathbf{z})  \right) e^{Lt} \right] \\
    &=\frac{1}{m}\sum_i \mathbb{E}\left[ \int_0^1 \begin{pmatrix} - e^{(1-t)\lambda_S}e^{\lambda_S} (\boldsymbol{\alpha_i} + \sigma\mathbf{z}_S) \\ e^{(1-t)\lambda_T}(\mathbf{m}_T - e^{\lambda_T}\sigma\mathbf{z}_T)\\-\sigma\mathbf{z}_\perp \end{pmatrix} \otimes \begin{pmatrix} e^{t\lambda_S}(\boldsymbol{\alpha}_i+\sigma\mathbf{z}_S)\\e^{t\lambda_T}\sigma\mathbf{z}_T \\ \sigma\mathbf{z}_\perp\end{pmatrix}dt\right]
\end{align*}
which allows us to split the computation into blocks of $[S,T,\perp]\otimes[S,T,\perp]$. We will use the identity:
\[
    \int_0^1e^{(1-t)\lambda_a}e^{t\lambda_b}dt=\begin{cases}
    \kappa_{ab}\equiv\frac{e^{\lambda_a} - e^{\lambda_b}}{\lambda_a-\lambda_b} & \lambda_a\neq\lambda_b \\
    e^{\lambda_a} & \lambda_a=\lambda_b
\end{cases}
\]}

\rd{\textbf{$S\times S$ block}. This yields:
\begin{align*}
    -\sigma^2 I_{m-1}&=-\frac{1}{m}\sum_i \mathbb{E}\left[ \int_0^1 e^{(1-t)\lambda_S}e^{t\lambda_S} e^{\lambda_S}(\boldsymbol{\alpha}_i +\sigma\mathbf{z}_S)(\boldsymbol{\alpha}_i +\sigma\mathbf{z}_S)^\top dt\right] \\
    &= -e^{2\lambda_S}\left( \frac{1}{m}\sum_i \boldsymbol{\alpha}_i \boldsymbol{\alpha}_i^\top +\frac{\sigma}{m}\sum_i \mathbb{E} \left[\boldsymbol{\alpha}_i \mathbf{z}_S^\top +\mathbf{z}_S \boldsymbol{\alpha}_i^\top\right]+\sigma^2\mathbb{E}\left[ \mathbf{z}_S \mathbf{z}_S^\top \right]\right) \\
    &=-e^{2\lambda_S}\left( \frac{R^2}{m}+ \sigma^2\right)I_{m-1}\\
    &\implies \lambda_S=-\frac{1}{2}\ln\left( \frac{\frac{R^2}{m}+\sigma^2}{\sigma^2} \right) <0
\end{align*}}

\rd{\textbf{$T\times T$ block.} The fixed point equation yields:
\begin{align*}
    -\sigma^2I_{m-1} &= \frac{1}{m}\sum_i \mathbb{E}\left[ \int_0^1 e^{(1-t)\lambda_T} e^{t\lambda_T} \left( \mathbf{m}_T - e^{\lambda_T}\sigma \mathbf{z}_T \right)\otimes (\sigma\mathbf{z}_T) dt\right] \\ 
    &=\sigma e^{\lambda_T} \mathbb{E}\left[ \mathbf{m}_T \otimes \mathbf{z}_T \right] - \sigma^2 e^{2\lambda_T}\mathbb{E}\left[ \mathbf{z}_T \otimes \mathbf{z}_T \right] \\
    &=\sigma e^{\lambda_T} \mathbb{E}\left[ \mathbf{m}_T \otimes \mathbf{z}_T  \right] - \sigma^2 e^{2\lambda_T} I_{m-1}
\end{align*}
Since $\mathbf{m}_T=\sum_j \gamma_j(\mathbf{x}_1) \boldsymbol{\beta}_j$ depends on $\mathbf{z}_T$ through $\mathbf{x}_1$, we cannot factorise the expectation. Instead, we must use Stein's lemma which states that for $\mathbf{z}\sim\mathcal{N}(\mathbf{0},I)$:
\[
    \mathbb{E}\left[ f(\mathbf{z})\mathbf{z}^\top \right]=\mathbb{E}\left[ \frac{\partial f}{\partial \mathbf{z}} \right]^\top
\]
for some smooth, bounded function $f$. Using standard results for derivatives of the softmax function and the chain rule through the logits, we can compute:
\[
    \frac{\partial \mathbf{m}_T}{\partial \mathbf{z}_T}=\frac{e^{\lambda_T}}{\sigma} \left( \sum_j \gamma_j \boldsymbol{\beta}_j \boldsymbol{\beta}_j^\top -\mathbf{m}_T \mathbf{m}_T^\top \right)
\]
This expression has the form of a covariance over the targets, weighted by the probability distribution over the targets $\gamma$. We do not compute it but simply note that since the means and covariances of the target modes are symmetric, and noise $\mathbf{z}_T$ is isotropic, the expected weighted covariance has no preferred direction and so is proportional to the identity. We can then write $C=\mathbb{E}\left[ \sum_j \gamma_j\boldsymbol{\beta}_j \boldsymbol{\beta}_j  - \mathbf{m}_T \mathbf{m}_T^\top\right]=cI_{m-1}$. Thus the fixed point equation becomes:
\begin{align*}
    -\sigma^2I_{m-1}&= e^{2\lambda_T} c I_{m-1} - \sigma^2 e^{2\lambda_T} I_{m-1} \\
    &e^{2\lambda_T}\left( \sigma^2 -c \right) = \sigma^2
\end{align*}
We cannot directly solve this since $c$ depends on $\lambda_T$ through the softmax weightings $\gamma$. Instead we use a numerical solver, and show at the end of the section that the obtained value closely matches that from training a model.}

\rd{\textbf{$S \times T$ block.} We have:
\[
    \frac{1}{m}\sum_i \mathbb{E}\left[ \left( \dots \right)\otimes \mathbf{z}_T \right] = \frac{1}{m}\sum_i \mathbb{E}\left[ \dots \right]\otimes\mathbb{E}\left[ \mathbf{z}_T \right]=0
\]
because $\mathbf{z}_T$ is noise independent from all $S$ components.}

\rd{\textbf{$T\times S$ block.} We have:
\[
    \frac{1}{m}\sum_i \mathbb{E}\left[ \left( \dots \right)\otimes \left( \boldsymbol{\alpha}_i + \sigma\mathbf{z}_S \right) \right]=\mathbb{E}\left[ \dots \right]\otimes \left( \frac{1}{m}\sum_i \boldsymbol{\alpha}_i +\sigma\mathbb{E}\left[ \mathbf{z}_S \right] \right)=0
\]
since the modes are centred and $\mathbf{z}_S$ is noise independent of all $T$ components.}

\rd{\textbf{$\perp\times\perp$ block.} This block is simply:
\[
    \sigma^2 \mathbb{E}\left[ -\mathbf{z}_\perp \otimes\mathbf{z}_\perp \right]=-\sigma^2 I_{d-2(m-1)}
\]
which immediately satisfies the fixed point equation.}

\rd{\textbf{Cross blocks involving $\perp$.} Since $\mathbf{z}_\perp$ is noise independent from all other blocks, all terms involving it evaluate to zero due to the expectation.}

\rd{We have thus verified that our ansatz satisfies the fixed point equation in each block, and either analytically or numerically found a value for the diagonal scalings which closely match the values found from training the models (Table~\ref{tab:saddle_eigs}).}

\begin{table}[htbp]
    \centering
    \begin{tabular}{l*{3}{r}}
    \toprule
     & Analytic & Computed\\
    \midrule
    $\lambda_S$ & -1.3635 & -1.3629  \\
    $\lambda_T$ & 0.9674 & 0.9669  \\
    $\lambda_\perp$ & 0 & -0.0002\\
    \bottomrule
    \end{tabular}
    \caption{Analytically derived vs numerically computed eigenvalues of the saddle compression point weights for $m=7$, $R=10$. }
    \label{tab:saddle_eigs}
\end{table}

\rd{\subsection{Response of statistics to perturbations around compression saddle point}\label{apd:CNF_statistics_perturbation}}

\rd{In order to study the behaviour of the training dynamics near the compression saddle, we study how the distance statistics defined in App.~\ref{apd:CNF_analyses} respond to perturbations around the saddle weights, in the direction of the optimal or collapsed solutions.}

\rd{We consider the compression saddle setup detailed in App.~\ref{apd:CNF_fixed_points}, with saddle weights:
\[
    W_s=ULU^\top
\]
where $U=[Q_S, Q_T, Q_\perp]$ has a block structure where $Q_S,Q_T$ are $d\times (m-1)$ dimensional with orthonormal columns spanning $\{\boldsymbol{\mu}_i\}$ and $\{\boldsymbol{\mu}_i^*\}$ respectively and $Q_\perp$ forms an orthonormal basis on the remaining $d-2(m-1)$ dimensions. We refer to $S$ and $T$ as the initial and target subspaces spanning $\{\boldsymbol{\mu}_i\}$ and $\{\boldsymbol{\mu}_i^*\}$ respectively. $L$ has the block structure: 
\[
    L=\text{diag}
        (\lambda_S I_{m-1},\lambda_T I_{m-1} , \mathbf{0})
\]
For initial and target modes of constant radius, we can write $D_{ij}=-2s_{ij}+c_i$ where 
\[
    s_{ij}=\frac{\langle e^W \boldsymbol{\mu}_i, \boldsymbol{\mu}_j^*\rangle}{R^2}
\]
and $c_i$ is a constant across row $i$. Next, we consider perturbations of the form 
\[
    \delta W=\epsilon (\tilde B - \tilde B^\top), \qquad \tilde B = Q_T B Q_S^\top
\]
where $B$ is an $(m-1)\times (m-1)$ matrix. $\delta W$ explicitly mixes the initial and target subspaces, and so depending on the choice of $B$, perturbs the saddle towards either collapsed or non-collapsed solutions. In practice, we construct the perturbation as:
\[
    \tilde B = \frac{\sum_i \boldsymbol{\mu}_{j_0(i)}^* \boldsymbol{\mu}_i^\top}{\left\lVert \sum_i \boldsymbol{\mu}_{j_0(i)}^* \boldsymbol{\mu}_i^\top \right\rVert_F}
\]
where $j_0(i)$ is the chosen assignment pattern. Fig.~\ref{fig:f2}g shows that training a model from saddle weights perturbed by $\delta W$ constructed by different assignment patterns $j_0$ can induce different solutions. }

\rd{At the saddle compression weights $s_{ij}=0$, since $W_s$ acts in the initial and target subspaces independently. Thus, any change to the statistics will come from $\delta W$ mixing initial and target subspaces. We can compute the first order effect of the perturbation on the matrix exponential using the Fréchet derivative in the direction of $\delta W$, at $W_s$: 
\begin{align*}
    \delta(e^W)&=\int_0^1 e^{(1-s)W_s} \delta W e^{sW_s} ds  \\
    &= \epsilon \int_0^1 e^{(1-s)W_s} \left( Q_T B Q_S^\top - Q_S B^\top Q_T^\top \right) e^{sW_s} ds \\
    &= \epsilon\left[ \int_0^1 e^{(1-s)\lambda_T}e^{s\lambda_S}ds Q_T B Q_S^\top - \int_0^1 e^{(1-s)\lambda_S}e^{s\lambda_T}ds Q_S B^\top Q_T^\top \right] \\
    &= \epsilon \frac{e^{\lambda_T} - e^{\lambda_S}}{\lambda_T - \lambda_S} (Q_T B Q_S^\top - Q_S B^\top Q_T^\top) \\ 
    &= \epsilon \kappa_B (\tilde B - \tilde B^\top)
\end{align*}
where $\kappa_B=\frac{e^{\lambda_T} - e^{\lambda_S}}{\lambda_T - \lambda_S}$ is a constant determined by the saddle compression eigenvalues. }

\rd{We now want to know how the initial modes respond to this perturbation. At the saddle: 
\[
    e^{W_s} \boldsymbol{\mu}_i = e^{\lambda_S} \boldsymbol{\mu}_i
\]
with no component in the target direction. The only target component comes from the perturbation which mixes initial into target subspace:
\begin{align*}
    \delta(e^W) \boldsymbol{\mu}_i &= \epsilon \kappa_B (\tilde B - \tilde B^\top)\boldsymbol{\mu}_i  \\
    &= \epsilon \kappa_B Q_T B Q_S^\top \boldsymbol{\mu}_i \\
\end{align*}
where the second term drops since $\tilde B^\top \boldsymbol{\mu}_i = Q_S B^\top Q_T^\top \boldsymbol{\mu}_i=0$ because $\boldsymbol{\mu}_i\perp T$. Then we can compute the overlaps to first order:
\[
    s_{ij}=\frac{\langle e^W\boldsymbol{\mu}_i, \boldsymbol{\mu}_j^*\rangle}{R^2}\approx\epsilon\kappa_B\frac{\langle Q_TBQ_S^\top \boldsymbol{\mu}_i, \boldsymbol{\mu}_j^*\rangle}{R^2}=\epsilon \kappa_B \frac{\langle B\boldsymbol{\alpha}_i,\boldsymbol{\beta}_j\rangle}{R^2}
\]
where the penultimate expression is in ambient coordinates and the final expression is in $T$ coordinates with $\boldsymbol{\alpha}_i=Q_S^\top \boldsymbol{\mu}_i$ and $\boldsymbol{\beta}_i=Q_T^\top \boldsymbol{\mu}_i^*$. Hence, to first order, the overlaps are bilinear in $\epsilon$ and $B$. We confirm that the overlaps respond linearly to perturbation strength $\epsilon$ for different $B$ matrices in Fig.~\ref{fig:apd_perturbations}a by measuring the change in $\lVert s\rVert_F$ for varying perturbation strength $\epsilon$. }

\rd{\subsection{Response of operator ranks to perturbations around saddle}}

We would like to measure how the rank of the state and adjoint operators change in response to perturbations around the saddle point. Similarly to the response of the summary statistics, we take the saddle weight structure as $W_s=ULU^\top$ with $U=[Q_S, Q_T, Q_\perp]$ and $L=\text{diag}(\lambda_S I_{m-1}, \lambda_T I_{m-1}, \mathbf{0})$. The perturbations are of the form $\delta W = \epsilon(\tilde B - \tilde{B}^\top)$ with 
\[
    \tilde{B} = \frac{\sum_i \boldsymbol{\mu}_{j_0(i)}^* \boldsymbol{\mu}_i^\top}{\left\lVert \sum_i \boldsymbol{\mu}_{j_0(i)}^* \boldsymbol{\mu}_i^\top\right\rVert} 
\]
where $j_0(i)$ is the chosen assignment pattern. In the previous section we compute that the matrix exponential of $W=W_s +\delta W$ to first order becomes:
\begin{align*}  
    e^{W_s+\delta W}&=e^{W_s} + \epsilon\kappa_B (\tilde{B}- \tilde{B}^\top) \\
    &= e^{W_s}+\epsilon\kappa_B(Q_T B Q_S^\top - Q_SB^\top Q_T^\top)
\end{align*}
where $\kappa_B=\frac{e^{\lambda_T} - e^{\lambda_S}}{\lambda_T - \lambda_S}$. Thus, we can compute $\mathbf{x}_1=e^W\mathbf{x}_0$ to first order as:
\begin{align*}
    \mathbf{x}_1&=e^{W_s+\delta W} (\boldsymbol{\mu}_i+\mathbf{z}) + \mathcal{O}(\epsilon^2)\\
    &=e^{W_s}(\boldsymbol{\mu}_i+\mathbf{z})+\epsilon \kappa_B (Q_TBQ_S^\top - Q_S B^\top Q_T^\top)(\boldsymbol{\mu}_i +\mathbf{z})+ \mathcal{O}(\epsilon^2) \\ &= \begin{pmatrix} e^{\lambda_S} (\boldsymbol{\alpha}_i + \mathbf{z}_S) \\ e^{\lambda_T}\mathbf{z}_T \\ \mathbf{z}_\perp \end{pmatrix} + \epsilon\kappa_B \begin{pmatrix} -B^\top \mathbf{z}_T \\ B(\boldsymbol{\alpha}_i+\mathbf{z}_S) \\ \mathbf{0} \end{pmatrix} + \mathcal{O}(\epsilon^2)\\
    &= \mathbf{x}_1^{(0)} +\epsilon \mathbf{x}^{(1)}+ \mathcal{O}(\epsilon^2)
\end{align*}
where in the penultimate line we split into $[S, T, \perp]$ subspaces.

\subsubsection{State operator response}
For the linear CNF the state operator covariance is
\[
    H^\top H= \mathbb{E}\left[ \int_0^1 \mathbf{x}(t) \otimes \mathbf{x}(t) dt \right]
    =\int_0^1 e^{tW} C_H e^{tW^\top}dt, \quad C_H=\mathbb{E}\left[ \mathbf{x}_0 \otimes \mathbf{x}_0 \right]
\]
where $\mathbf{x}(t)=e^{Wt}\mathbf{x}_0=e^{Wt}(\boldsymbol{\mu}_i+\mathbf{z})$ is a transported initial sample and $\mathbf{z}\sim\mathcal{N}(\mathbf{0}, I)$. We use the above form of the expansion of $\mathbf{x}_1$ to first order to compute the covariance $C_H$. The stable rank depends on the eigenvalues of the state covariance, and so we must propagate the effect of the perturbation $W=W_s+\delta W$ to the eigenvalues. We first consider the effects on the terminal covariance $C_H$, arguing that due to the structure of $W_s$, the same qualitative result holds for the integrated covariance $H^\top H$.

\textbf{Zeroth order.} $\boldsymbol{\alpha}_i+\mathbf{z}_S$, $\mathbf{z}_T$ and $\mathbf{z}_\perp$ are independent and have zero mean (the initial modes are centred), so the $S\times T$ and $T\times S$ blocks are zero. The $S\times S$ block is:
\begin{align*}
    C_{H,SS}^{(0)}&= \frac{e^{2\lambda_S}}{m}\sum_i\mathbb{E}\left[ (\boldsymbol{\alpha}_i +\mathbf{z}_S)\otimes(\boldsymbol{\alpha}_i + \mathbf{z}_S)\right] \\&= e^{2\lambda_S}\left( \frac{R^2}{m}I_{m-1} +I_{m-1}\right)\\
    &=e^{2\lambda_S}\left( \frac{R^2}{m}+1 \right)I_{m-1}
\end{align*}
The $T\times T$ block is:
\begin{align*}
    C_{H,TT}^{(0)}&=e^{2\lambda_T}\mathbb{E}\left[ \mathbf{z}_T \otimes \mathbf{z}_T \right] \\&= e^{2\lambda_T}I_{m-1}
\end{align*}
All cross blocks with $\perp$ components are zero because $\mathbf{z}_\perp$ is noise independent of other subspaces. The $\perp\times\perp$ block is $I_{d-2(m-1)}$.

\textbf{First order.} The first order contains the terms:
\[
    C_H^{(1)}=\mathbb{E}\left[ \mathbf{x}^{(0)}\otimes \mathbf{x}^{(1)} +\mathbf{x}^{(1)}\otimes \mathbf{x}^{(0)}\right]
\]
The $S\times S$ and $T\times T$ blocks are zero since $\mathbf{z}_T$ is independent of $\boldsymbol{\alpha}_i$ and $\mathbf{z}_S$. The $S\times T$ block is:
\begin{align*}
    C_{H,ST}^{(1)}&=\mathbb{E}\left[ e^{\lambda_S}(\boldsymbol{\alpha}_i+\mathbf{z}_S)\otimes (B(\boldsymbol{\alpha}_i+\mathbf{z}_S)) - e^{\lambda_T}(B^\top\mathbf{z}_T)\otimes \mathbf{z}_T \right] \\
    &=  e^{\lambda_S}\left( \frac{R^2}{m}+1 \right)I_{m-1}B^\top - e^{\lambda_T}B^\top I_{m-1}\\ &=\left( e^{\lambda_S}\left( \frac{R^2}{m}+1\right)-e^{\lambda_T} \right)B^\top
\end{align*}
The $T\times S$ block is:
\begin{align*}
    C_{H,TS}^{(1)}&=\mathbb{E}\left[ -e^{\lambda_T}\mathbf{z}_T \otimes (B^\top \mathbf{z}_T) +e^{\lambda_S} B(\boldsymbol{\alpha}_i +\mathbf{z}_S)\otimes (\boldsymbol{\alpha}_i+\mathbf{z}_S) \right] \\
    &= B e^{\lambda_S}\left( \frac{R^2}{m}+1 \right)I_{m-1} - e^{\lambda_T}I_{m-1}B \\
    &= \left( e^{\lambda_S}\left( \frac{R^2}{m}+1\right) -e^{\lambda_T} \right)B\\
    &=\left(C_{H,ST}^{(1)}\right)^\top
\end{align*}

\begin{figure}
    \centering
    \includegraphics[width=0.8\linewidth]{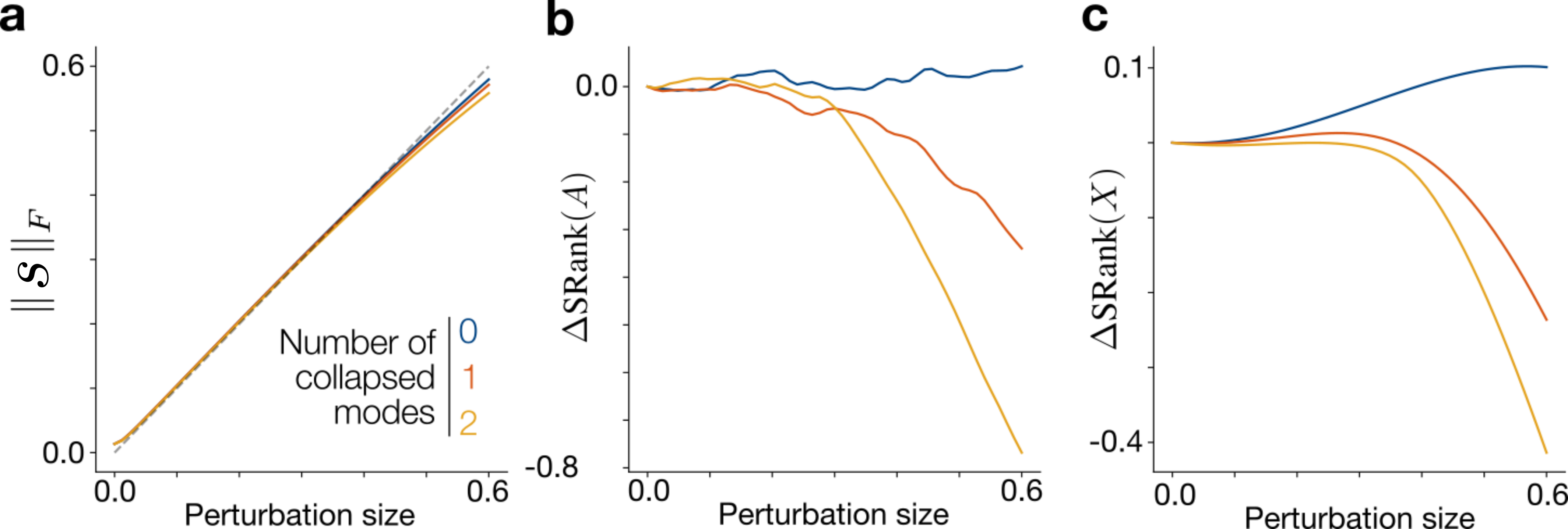}
    \caption{\textbf{Response to perturbations around the compression saddle weights. }\textbf{a.} Magnitude of distance statistics matrix in response to perturbation strength around saddle point. Closely matches linear trend. \textbf{b-c.} Change in rank of the adjoint and state operators in response to different perturbation strengths.}
    \label{fig:apd_perturbations}
\end{figure}

\textbf{Putting together.} Now we have the form of the covariance $C_H=C_H^{(0)}+\epsilon C_H^{(1)}+\mathcal{O}(\epsilon^2)$ where:
\[
    C_H^{(0)}=\begin{pmatrix} C_{H,SS}^{(0)} & \mathbf{0} \\
    \mathbf{0} & C_{H,TT}^{(0)}\end{pmatrix}, \qquad C_H^{(1)}=\begin{pmatrix} \mathbf{0} & C_{H,ST}^{(1)}\\
    \left(C_{H,ST}^{(1)}\right)^\top & \mathbf{0}\end{pmatrix}
\]
Hence the block-diagonal part of $C_H$ equals $C_H^{(0)}+\mathcal{O}(\epsilon^2)$, and by Weyl's inequality its eigenvalues differ from $C_H^{(0)}$ eigenvalues by $\mathcal{O}(\epsilon^2)$.
The off-diagonal part is $O(\epsilon)$, and since the two diagonal blocks have disjoint spectra (eigenvalues $\approx1$ for $S$ and $\approx 6.92$ for $T$), it shifts each eigenvalue by at most $2\|\Delta\|^2/(\eta+\sqrt{\eta^2+4\|\Delta\|^2})=O(\epsilon^2)$ where $\eta$ is the gap in the eigenvalues~\cite{li2005note}. Every eigenvalue of $C_H$, and therefore its stable rank, thus changes only at $O(\epsilon^2)$ and so there is no linear response. The same argument applies to $H^\top H$, since at the saddle $e^{sW_s}$ is block diagonal and the first-order change in $e^{sW}$ has only off-diagonal blocks, so $H^\top H^{(1)}$ also has vanishing diagonal blocks. The state rank response is shown in Fig.~\ref{fig:apd_perturbations}c.  

\rd{\subsection{Response of adjoint to perturbations around saddle}\label{apd:CNF_adjoint_perturbation}}

\rd{Next, we would like to measure how the rank of the adjoint operator changes in response to the same perturbations.
Analytically, the adjoint covariance is 
\[
    AA^\top=\int_0^1 e^{sW^\top} C_A e^{sW} ds,\qquad C_A=\mathbb{E}\left[\mathbf{a}_1 \mathbf{a}_1^\top\right], \qquad \mathbf{a}_1=\mathbf{x}_1-\sum_j S_j(\mathbf{x}_1) \boldsymbol{\mu}_j^*=\mathbf{x}_1-\bar{\mathbf{m}}(\mathbf{x}_1)
\]
where $\mathbf{a}_1=-\nabla_{\mathbf{x}_1} \log p^*(\mathbf{x}_1)$ is the adjoint terminal state and $\mathbf{x}_1=e^W \mathbf{x}_0=e^W (\boldsymbol{\mu}_i + \mathbf{z})$ is a pushed forward initial sample from the $i^\text{th}$ initial mode. Similarly to before, the stable rank depends on the eigenvalues of the adjoint covariance, and so we must propagate the effect of the perturbation $W=W_s+\delta W$ to the eigenvalues through the adjoint.}

\rd{Using the definition of the expansion of $\mathbf{x}_1$ in the last section, we compute the adjoint terminal state, starting with $\mathbf{a}_1=\mathbf{a}_1^{(0)} + \mathbf{a}_1^{(1)} + \mathcal{O}(\epsilon^2)$. From the chain rule we have that $\mathbf{a}_1^{(1)} = \frac{\partial \mathbf{a}_1}{\partial \mathbf{x}_1} \mathbf{x}_1^{(1)}=(I-J(\mathbf{x}_{1,T})) \mathbf{x}_1^{(1)}$ where $J(\mathbf{x}_{1,T})=\frac{\partial \bar{\mathbf{m}}(\mathbf{x}_{1,T})}{\partial \mathbf{x}_1}=\sum_j S_j \boldsymbol{\mu}_j^* \boldsymbol{\mu}_j^{*\top} - \bar{\mathbf{m}}(\mathbf{x}_{1,T}) \bar{\mathbf{m}}(\mathbf{x}_{1,T})^\top$ which acts only in the $T$ subspace. We can then write the components of the adjoint as:
\[
    \mathbf{a}_1^{(0)} = \begin{pmatrix} e^{\lambda_S}(\boldsymbol{\alpha}_i+\mathbf{z}_S)\\ e^{\lambda_T} \mathbf{z}_T-\bar{\mathbf{m}}(e^{\lambda_T} \mathbf{z}_T)\\ \mathbf{z}_\perp \end{pmatrix}, \qquad \mathbf{a}_1^{(1)}= \epsilon\kappa_B\begin{pmatrix} - B^\top \mathbf{z}_T \\ (I-J(e^{\lambda_T} \mathbf{z}_T)) B(\boldsymbol{\alpha}_i+\mathbf{z}_S) \\ \mathbf{0} \end{pmatrix}
\]
Now, a similar expansion of the adjoint covariance yields $C=C^{(0)}+C^{(1)} + \mathcal{O}(\epsilon^2)$. First we consider the zeroth-order term $C^{(0)}$, noting that the off-diagonal blocks are zero: 
\[  
    C^{(0)}_{ST}=\mathbb{E}\left[ \mathbf{a}_{1,S}^{(0)} \mathbf{a}_{1,T}^{(0)\top} \right]=\mathbb{E}\left[ e^{\lambda_S}(\boldsymbol{\alpha}_i + \mathbf{z}_S)\left( e^{\lambda_T}\mathbf{z}_T - \bar{\mathbf{m}}(e^{\lambda_T} \mathbf{z}_T) \right)^\top \right]=\mathbf{0}
\]
since $\mathbb{E}[\boldsymbol{\alpha}_i+\mathbf{z}_S]=\mathbf{0}$. We are interested in the first order response to the perturbation, so we consider the terms:
\[
    C^{(1)}=\mathbb{E} \left[ \mathbf{a}^{(0)} \mathbf{a}^{(1)\top} + \mathbf{a}^{(1)}\mathbf{a}^{(0)\top} \right]
\]
Using the previous results, we can compute this quantity block by block. }

\rd{\textbf{The initial subspace block:}
\[
    \mathbb{E} \left[ \mathbf{a}_S^{(0)} \mathbf{a}_S^{(1)\top} \right]=-\frac{\epsilon\kappa_B}{m}\sum_i \mathbb{E} \left[ e^{\lambda_S} (\boldsymbol{\alpha}_i + \mathbf{z}_S) \mathbf{z}_T^\top B  \right] =\mathbf{0}
\]
due to the fact that $\mathbf{z}_T$ is zero-centred.}

\rd{\textbf{The target subspace block:}
\[
    \mathbb{E}\left[ \mathbf{a}^{(0)}_T \mathbf{a}_T^{(1)\top}  \right] = \frac{1}{m}\sum_i\mathbb{E}\left[\epsilon\kappa_B \left(e^{\lambda_T} \mathbf{z}_T - \bar{\mathbf{m}}(e^{\lambda_T} \mathbf{z}_T)\right) (\boldsymbol{\alpha}_i + \mathbf{z}_S)^\top B^\top \left(I - J(e^{\lambda_T} \mathbf{z}_T)\right)^\top\right] =\mathbf{0}
\]
due to the centring of $\boldsymbol{\alpha}_i$ and independence of noise in initial and target subspaces.}

\rd{\textbf{Blocks involving $\perp$ dimensions:} Trivially zero since $\mathbf{z}_\perp$ is zero-mean and independent of all other components.}

\rd{\textbf{Initial-target cross blocks:} In general non-zero due to terms like $\mathbb{E}\left[ (\boldsymbol{\alpha}_i+\mathbf{z}_S)(\boldsymbol{\alpha}_i+\mathbf{z}_S)^\top \right]$, and proportional to $\epsilon$. Numerically computed to be small.}

\rd{The first order change in stable rank of the adjoint operator will depend on the first order change in the eigenvalues of $C^{(1)}$. This has a block structure with zero blocks on the diagonal and small, non-zero off-diagonal blocks proportional to $\epsilon$. Since we have shown that $C^{(0)}$ is block diagonal, we have the following structure:
\[
    C^{(0)}=\begin{pmatrix} C_S & \mathbf{0}\\ \mathbf{0} & C_T \end{pmatrix}, \qquad C^{(1)}=\begin{pmatrix} \mathbf{0} & \Delta \\ \Delta^\top & \mathbf{0} \end{pmatrix}, \qquad \Delta\propto\epsilon
\]
By the same argument as the state operator, the stable rank responds at $\mathcal{O}(\epsilon^2)$. The non-linear response of the maximum eigenvalues of $C$ and the rank of the adjoint operator can be seen empirically in Fig.~\ref{fig:apd_perturbations}b.}

\section{Effective collapse in flow matching models}\label{apd:FM}
\rd{The goal of flow matching is to learn some function $\mathbf{u}_t^\theta(\mathbf{x})\equiv\mathbf{u}^\theta(\mathbf{x}, t)$ which transports samples $\mathbf{x}(0)\sim p_0$, where $p_0$ is a simple distribution (typically Gaussian), to $\mathbf{x}(1)\sim p^*$ via integration:
\[
    \mathbf{x}_1= \mathbf{x}_0+\int_0^1 \mathbf{u}_t^\theta(\bar{\mathbf{x}}(t))dt
\]
where $\bar{\mathbf{x}}(t)$ follows a chosen ``interpolant trajectory''. To do this, we define the marginal probability path $p_t$, which smoothly interpolates between the initial and data distribution, satisfying $p_{t=0}=p_0$ and $p_{t=1}=p^*$. The flow matching objective can then be written generally as 
\[
    \mathcal{L}_{\text{FM}}=\int_0^1\mathbb{E}_{\mathbf{x}\sim p_t}\left[||\mathbf{u}_t^*(\mathbf{x})-\mathbf{u}_t^\theta(\mathbf{x})||^2 \right]dt
\]
where $\mathbf{u}_t^*(\mathbf{x})$ is the optimal velocity field that generates $p_t$. In general, the optimal velocity is intractable, and so in practice the conditional flow matching objective is used:
\[
    \mathcal{L}_\text{CFM}=\int_0^1 \mathbb{E}_{\mathbf{z}\sim p^*, \mathbf{x}\sim p_t(\cdot\mid \mathbf{z})}\left[||\mathbf{u}_t^*(\mathbf{x}\mid \mathbf{z}) - \mathbf{u}_t^\theta(\mathbf{x})||^2\right]dt
\]
where $p_t(\cdot\mid \mathbf{z})$ is the conditional probability path which is related to the marginal by $p_t(\mathbf{x})=\int p_t(\mathbf{x}\mid \mathbf{z}) p^*(\mathbf{z})d\mathbf{z}$, and similarly $\mathbf{u}_t^*(\mathbf{x})=\mathbb{E}\left[ \mathbf{u}_t^*(\mathbf{x}\mid \mathbf{z}) \mid  \bar{\mathbf{x}}(t)=\mathbf{x} \right]$ relates the marginal and conditional velocity fields. There is a fundamental theorem in flow matching which says that the conditional flow matching loss has the same gradients as the flow matching loss\cite{lipman2023flow}. 
From the conditional probability path we can easily compute the target vector field by taking the time derivative, $\mathbf{u}_t^*(\bar{\mathbf{x}}(t)\mid \mathbf{z})=\dot{\mathbf{x}}(t)$, which we can plug into the conditional loss and train our model.}

\subsection{Linear flow matching}\label{apd:linear_fm}
\rd{\textbf{Model.}} To compare the FM objective to the linear CNF model, we choose $\mathbf{u}_t^\theta(\mathbf{x})=W\mathbf{x}$. 
\rd{As before, because linear flows cannot split or join modes, we define our base and target distributions to be mixtures of Gaussians:
\[
    p_0(\mathbf{x})=\sum_{i=1}^m w_i \mathcal{N}(\mathbf{x}; \boldsymbol{\mu}_i, \sigma^2 I_d), \qquad p^*(\mathbf{x})=\sum_{j=1}^m \pi_j \mathcal{N}(\mathbf{x}; \boldsymbol{\mu}_j^*, \sigma^2 I_d) 
\]
We now choose all modes to be orthogonal:
\[
    \boldsymbol{\mu}_i\cdot\boldsymbol{\mu}_j^*=0,\qquad \boldsymbol{\mu}_i\cdot\boldsymbol{\mu}_j=\boldsymbol{\mu}_i^*\cdot\boldsymbol{\mu}_j^*=R^2 \delta_{ij}
\]
In App.~\ref{apd:optimal_FM_field}, we prove that for such a setup, the standard linear interpolation probability path yields an optimal solution which contains a non-linear weighting over modes, and thus our linear model cannot represent it. To account for this, we consider a teacher student setup where the optimal velocity field is $\mathbf{u}_t^*(\mathbf{x})=W^* \mathbf{x}$ and the probability path is $\bar{\mathbf{x}}(t)=e^{tW^*}\mathbf{x}(0)$ (i.e. the solution of the dynamics). We define $W^*$ as the generator matrix of a rotation which maps initial modes to their respective targets:
\[
    W^*=\frac{\pi}{2}(B-B^\top) ,\qquad B=\frac{1}{R^2}\sum_i \boldsymbol{\mu}_{j_0(i)}^*\boldsymbol{\mu}_i^\top
\]
where $j_0(i)$ is the correct assignment pattern, which we choose without loss of generality to be $j_0(i)=i$. Below, we briefly show that this defines a valid probability path.}

\rd{\textbf{Rotational probability path.} We start by defining \[
    p_0^{(i)}=\mathcal{N}(\boldsymbol{\mu}_i,\sigma^2 I_d), \qquad p^{*(j)}(\mathbf{x})=\mathcal{N}(\boldsymbol{\mu}_j^*,\sigma^2 I_d)
\]
as the densities of Gaussian mixtures around each mode. At sampling time, we choose an $i,j$ pair according to weights $w_i,\pi_j$ and draw $\mathbf{x}^{(i)}(0)=\boldsymbol{\mu}_i+\sigma \mathbf{z}_0$ and $\mathbf{x}^{(j)}(1)=\boldsymbol{\mu}_j^*+\sigma\mathbf{z}_1$ where $\mathbf{z}_0,\mathbf{z}_1\sim\mathcal{N}(\mathbf{0},I_d)$. Next, we construct the conditional path sample as:
\begin{align*}
    \bar{\mathbf{x}}(t)&=e^{tW^*}\mathbf{x}^{(i)}(0) \\
    &=e^{tW^*}(\boldsymbol{\mu}_i+\sigma\mathbf{z}_0) \\
    &=e^{tW^*}\boldsymbol{\mu}_i+ \sigma\tilde{\mathbf{z}}_0
\end{align*}
where in the last line we used the fact that since $W^*$ is skew-symmetric ($W^{*\top}=-W^*$), $e^{tW^*}$ is orthogonal and so $\tilde{\mathbf{z}}_0=e^{tW^*}\mathbf{z}_0\sim\mathcal{N}(\mathbf{0},I_d)$. From this expression we have that $\bar{\mathbf{x}}(0)=\boldsymbol{\mu}_i+\sigma\tilde{\mathbf{z}}_0\sim p_0^{(i)}$, and $\bar{\mathbf{x}}(1)=\boldsymbol{\mu}_{j_0(i)}^*+\sigma\tilde{\mathbf{z}}_0\sim p^{*(j)}$ from the definition of $W^*$. Thus, $\bar{\mathbf{x}}(t)$ defines a valid probability path.}

\rd{\textbf{Training.}} From the rotational probability path, we can compute the optimal conditional velocity as $\dot{\mathbf{x}}(t)=W^*e^{tW^*}\mathbf{x}(0)=W^*\bar{\mathbf{x}}(t)$, and so we can write the objective as a linear regression integrated over time:
\[
\mathcal{L}=\mathbb{E}_{\mathbf{x}(0)\sim p_0}\left[\int_0^1 ||W^*\bar{\mathbf{x}}(t) - W\bar{\mathbf{x}}(t)||^2 dt \right]
\]
To train the model, we use full-batch gradient descent with a fixed set of initial samples and time points. We initialise weights as $W_{ij}\sim\mathcal{N}(0, \sigma_{\text{init}}/\sqrt{n})$. The hyperparameters are:

\begin{center}
    \begin{tabular}{cccccc}
    \hline
    $n$ & $\sigma_{\text{init}}$ & lr & Iterations & Batch size & Optimizer \\ \hline
    3-10  & $1e^{-4}$   & $1e^{-3}$ & 3000 & 10k & SGD  \\ \hline
    \end{tabular}
\end{center}

In figure~\ref{fig:saddle_linear_fm} we parameterise the linear flow as $\mathbf{u}_\theta(\mathbf{x})=W_2W_1\mathbf{x}$. Classic work in learning theory\cite{saxe_exact_2014} has shown that this parameterisation leads to non-linear training dynamics, inducing saddle-to-saddle dynamics.

\subsection{\rd{Spectrum of the state operator}}\label{apd:rank_data_op}

\rd{\textbf{Data assumptions.} As described in the previous section, we consider orthogonal modes radius $R$ with mode pairing $\boldsymbol{\mu}_i\rightarrow \boldsymbol{\mu}_i^*$. We define the optimal weights as:
\[
    W^*=\frac{1}{R^2}\frac{\pi}{2}\sum_i \left( \boldsymbol{\mu}_i^* \boldsymbol{\mu}_i^\top - \boldsymbol{\mu}_i \boldsymbol{\mu}_i^{*\top}\right)
\]
with interpolant path $\bar{\mathbf{x}}(t)=e^{tW^*}\mathbf{x}(0)$ and target $\mathbf{u}_t^*=W^*\bar{\mathbf{x}}(t)$. Also assume $w_i=\pi_i$ so that the path ends at $p^*$}

\textbf{Covariance operator of the \rd{linear} flow matching objective.} To decompose the gradient of the FM objective into two linear operators, we note that the gradient (up to a scalar constant) is:
\[
    \nabla_W\mathcal{L} 
    = -\mathbb{E}_{\mathbf{x}(0)}\left[\int_0^1 (W^*-W)\bar{\mathbf{x}}(t)\bar{\mathbf{x}}(t)^\top  dt \right] 
\]
which we can rewrite as the composition of two linear operators:
\[
    \nabla_W \mathcal{L} = A \circ H  = \mathbb{E}\left[\int_0^1 \mathbf{r}(t) \otimes \mathbf{\bar x}(t) dt\right]
\]
where the state operator $H$ is given by the interpolant trajectory $\mathbf{\bar x}(t)$, and the adjoint operator $A$ by the residual, $\mathbf{r}(t) = W\mathbf{\bar x}(t)-\mathbf{u}_t^*(\bar{\mathbf{x}}(t))$. 
We can compute the rank of $H$ analytically. For this, we will compute the eigenvalues of the matrix:
\[
    C = \text{ad}(H)\circ H = \mathbb{E}_{\mathbf{x}(0)}\left[\int_0^1 \bar{\mathbf{x}}(t) \otimes \bar{\mathbf{x}}(t) dt\right] \in \mathbb{}R^{n \times n}
\]

\rd{\textbf{Spectrum of the state operator.} Substituting the form of $\bar{\mathbf{x}}_t$ yields:
\begin{align*}
    C&=\mathbb{E}_{\mathbf{x}(0)}\left[ \int_0^1 \left( e^{tW^*}\mathbf{x}(0) \right)\otimes \left( e^{tW^*}\mathbf{x}(0) \right)dt \right] \\
    &= \sum_i w_i \mathbb{E}_{\mathbf{z}\sim\mathcal{N}(0,I_d)}\left[ \int_0^1 \left( e^{tW^*}(\boldsymbol{\mu}_i+\sigma \mathbf{z}) \right)\otimes \left( e^{tW^*}(\boldsymbol{\mu}_i + \sigma \mathbf{z}) \right)dt\right] \\ 
    &= \sum_i w_i \int_0^1\left((e^{tW^*}\boldsymbol{\mu_i})\otimes (e^{tW^*}\boldsymbol{\mu}_i )+\sigma^2 e^{tW^*}\mathbb{E}\left[ \mathbf{z}\otimes \mathbf{z} \right]e^{tW^{*\top}} \right)dt \\
    &=\sigma^2 I_d +\sum_i w_i \int_0^1 (e^{tW^*}\boldsymbol{\mu}_i)\otimes (e^{tW^*}\boldsymbol{\mu}_i) dt
\end{align*}
where we split the expectation into a sum over samples from each initial mode: $\mathbf{x}^{(i)}(0)=\boldsymbol{\mu}_i+\sigma\mathbf{z}$ where $\mathbf{z}\sim \mathcal{N}(\mathbf{0},I_d)$, and we used the fact that $\mathbb{E}_\mathbf{z}[\mathbf{z}]=0$ to cancel the cross terms, $\mathbb{E}[\mathbf{z}\otimes\mathbf{z}]=I_d$ and $W^{*\top}=-W^*$ such that $e^{tW^*}e^{tW^{*\top}}=I_d$.}

\rd{Since $W^*$ is skew-symmetric, $e^{tW^*}$ is orthogonal. The optimal rotation acts on plane $i$ by rotating the initial mode to the target mode, so $e^{tW^*}\boldsymbol{\mu}_i=\cos(\frac{\pi t}{2})\boldsymbol{\mu}_i+\sin(\frac{\pi t}{2})\boldsymbol{\mu}_i^*$. Thus the $n\times n$ covariance matrix splits into $2\times2$ blocks, written in the basis of $(\frac{\boldsymbol{\mu}_i}{R},\frac{\boldsymbol{\mu}_i^*}{R})$:
\begin{align*}
    C_i&=\sigma^2 I_2+ w_i R^2 \int_0^1 \begin{pmatrix} \cos^2(\frac{\pi t}{2}) & \cos(\frac{\pi t}{2})\sin(\frac{\pi t}{2})\\ \cos(\frac{\pi t}{2})\sin(\frac{\pi t}{2}) &  \sin^2(\frac{\pi t}{2})  \end{pmatrix}dt\\
    &=\sigma^2I_2 +w_i R^2 \begin{pmatrix} \frac{1}{2} & \frac{1}{\pi}\\ \frac{1}{\pi} & \frac{1}{2} \end{pmatrix}
\end{align*}
Thus we can compute the eigenvalues and eigenvectors for each block as
\[
    \lambda_{i,\pm}=\sigma^2+w_i R^2\left( \frac{1}{2}\pm \frac{1}{\pi} \right), \qquad \mathbf{v}_{i, \pm}=\frac{1}{\sqrt2}\begin{pmatrix} 1\\\pm1 \end{pmatrix}
\]
and $C=\sigma^2$ in directions with no modes.}

\rd{Since we define $C=\text{ad}(H)\circ H$, the singular values of $H$ are the square roots of the eigenvalues of $C$: $s_k(H)=\sqrt{\lambda_k}$. We are interested in the stable rank of $H$, which is:
\[
    \text{SRank}(H)=\frac{\sum_k \sqrt{\lambda_k}}{\sqrt{\lambda_\text{max}}}=\frac{\sum_k \left( \sqrt{\lambda_{k,+}} +\sqrt{\lambda_{k,-}} \right) + (d-2m)\sigma}{\sqrt{\lambda_{\text{max},+}}}
\]
Hence, as $w_i$ decreases, its contribution to the sum decreases and the state operator rank decreases.}

\rd{\subsection{Sequential learning of modes}\label{apd:FM_sequential_learning}}
\rd{\textbf{Shallow model $\mathbf{u}=W\mathbf{x}$.} The gradient flow is 
\[
    \dot{W}=-(W-W^*)C
\]
which is linear and can be solved as $W(\tau)-W^*=(W(0)-W^*)e^{-C\tau}$. Thus, the residual in the plane $(\boldsymbol{\mu}_i, \boldsymbol{\mu}_i^*)$ decays according to the smallest eigenvalue for each block: $e^{-\lambda_{i,-}\tau}$. The learning time is:
\[
    \tau_i\propto \frac{1}{\lambda_{i,-}}=\frac{1}{\sigma^2+w_iR^2\left( \frac{1}{2}-\frac{1}{\pi} \right)}
\]}

\rd{\textbf{Deep model $\mathbf{u}=W_2W_1\mathbf{x}$.} We write the input-output correlation as $\Sigma_{yx}=W^*C$ and input correlation as $\Sigma_{xx}=C$. 
Computing the SVD of the input correlation using its eigenbasis: $\Sigma_{xx}=\sum_k \lambda_k \mathbf{v}_k \mathbf{v}_k^\top$, which leads to:
\begin{align*}
    \Sigma_{yx}&=\sum_k \lambda_k(W^* \mathbf{v}_k)\mathbf{v}_k^\top \\
    &=\sum_k\left( \frac{\pi}{2}\lambda_k \right)\mathbf{u}_k \mathbf{v}_k^\top \\
    &=\sum_k s_k \mathbf{u}_k \mathbf{v}_k^\top
\end{align*}
From the form of $W^*$ and the eigenvectors of $C$, we compute:
\[
    W^* \mathbf{v}_{i,+}=-\frac{\pi}{2}\mathbf{v}_{i,-}\equiv\frac{\pi}{2}\mathbf{u}_{i,+}, \quad W^* \mathbf{v}_{i,-}=\frac{\pi}{2}\mathbf{v}_{i,+}\equiv\frac{\pi}{2}\mathbf{u}_{i,-}
\]
and absorbing the factor into the singular values gives $s_k=\frac{\pi}{2}\lambda_k$. Thus, $\Sigma_{xx}$ and $\Sigma_{yx}$ share the right basis and the dynamics separate mode by mode\cite{saxe_exact_2014}. The learning time of the $i^\text{th}$ mode goes like $\mathcal{O}(s_i^{-1})$ or equivalently $\mathcal{O}(w_i^{-1})$ when $w_iR^2\gg \sigma^2$. Note that the model must learn 2 singular mode directions for each mode, $\lambda_{i,+}$ and $\lambda_{i,-}<\lambda_{i,+}$, and so each mode also has 2 distinct learning times. The learning time equations apply to these sub-modes directly, namely $\tau_{i,\pm}\propto 1/\lambda_{i,\pm}=\mathcal{O}(w_i^{-1})$, and we consider a data mode learned when the model learns the both singular mode directions.}

\rd{For each data mode, we can compute what is learned by the model at the plateau and end of learning. In the plane $\left(\frac{\boldsymbol{\mu}_i}{R},\frac{\boldsymbol{\mu}_i^*}{R}\right)$, the plateau corresponds to learning:
\[
    W_i^{\text{plateau}}=\frac{\pi}{2}\mathbf{u}_{i,+}\mathbf{v}_{i,+}^\top =\frac{\pi}{4}\begin{pmatrix} -1 \\ 1 \end{pmatrix} \begin{pmatrix}1&1  \end{pmatrix}=\frac{\pi}{4}\begin{pmatrix} -1 & -1 \\1&1 \end{pmatrix}
\]
This velocity defines a shear which sends $\boldsymbol{\mu}_i \rightarrow \left( 1-\frac{\pi}{4} \right)\boldsymbol{\mu}_i+\frac{\pi}{4}\boldsymbol{\mu}_i^*$, which yields overlap $s_{ii}=\frac{\pi}{4}\approx0.79$}. \rd{At the end of training, the model learns the other direction:
\[
    W_i^{\text{converge}}=W_i^\text{plateau} + \frac{\pi}{2}\mathbf{u}_{i,-}\mathbf{v}_{i,-}^\top=\frac{\pi}{4}\begin{pmatrix} -1&-1\\1&1 \end{pmatrix}+\frac{\pi}{4}\begin{pmatrix} 1&-1\\1&-1 \end{pmatrix}=\frac{\pi}{2}\begin{pmatrix} 0 &-1 \\ 1&0\end{pmatrix}
\]
which is a rotation generator mapping $\boldsymbol{\mu}_i\rightarrow \boldsymbol{\mu}_i^*$, and hence the mode is learned ($s_{ii}=1$).}

\begin{figure}
    \centering
    \includegraphics[width=0.95\linewidth]{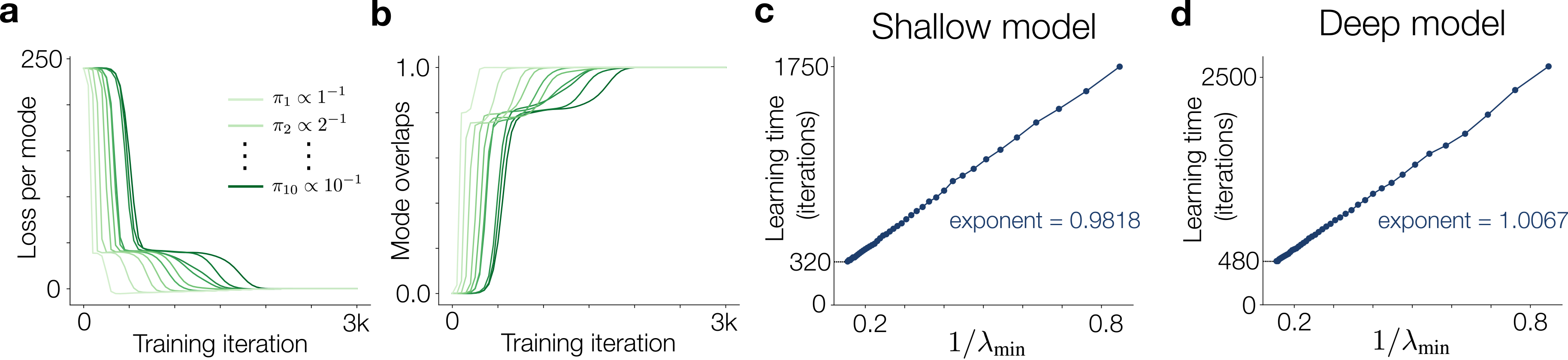}
    \caption{\textbf{Sequential learning for imbalanced mixture weights. }\textbf{a.} Loss per mode over training for a model with 10 modes and weightings $\pi_i\propto i^{-1}$. Progressively smaller modes are learned sequentially. Plateaus within a single mode curve correspond to learning the first and second singular direction corresponding to that mode. \textbf{b.} Mode overlaps $s_{ii}$ for each initial-target mode pair over training.  \textbf{c-d.} Plots of learning time of smallest mode against $1/\lambda_{\text{min}}$ for a shallow (\textbf{c}) and deep (\textbf{d}) linear model. Models trained with 3 modes with weightings $\left[\pi_\text{min}, \frac{1-\pi_\text{min}}{2}, \frac{1-\pi_\text{min}}{2}\right]$ with $\pi_\text{min}\in[0.01,0.3]$}
    \label{fig:apd_linear_learn_times}
\end{figure}

\rd{\textbf{Numerical validation.} To verify the scaling laws, we train different models whilst varying the weighting of the smallest mode, measuring how the time to learn each mode scales with weighting. To do this we track the per mode loss:
\[
    \mathcal{L}_i=\int_0^1 \lVert \mathbf{u}_t^*(\bar{\mathbf{x}}^{(i)}(t)) - \mathbf{u}_t^\theta(\bar{\mathbf{x}}^{(i)}(t))\rVert^2dt
\]
where $\bar{\mathbf{x}}^{(i)}(t)=e^{tW^*}\boldsymbol{\mu}_i$ is the transported $i^\text{th}$ initial mode, defining the $i^\text{th}$ mode learned when $\mathcal{L}_i$ drops below some threshold $\varepsilon$ (Fig.~\ref{fig:apd_linear_learn_times}a). Using the analytic form of $\lambda_\text{min}$, we also plot learning times against $1/\lambda_\text{min}$ for deep and shallow models, reporting the log-log exponent using np.polyfit which are both close to the expected value of 1 (Fig.~\ref{fig:apd_linear_learn_times}c-d). } 

\begin{figure}
    \centering
    \includegraphics[width=0.6\linewidth]{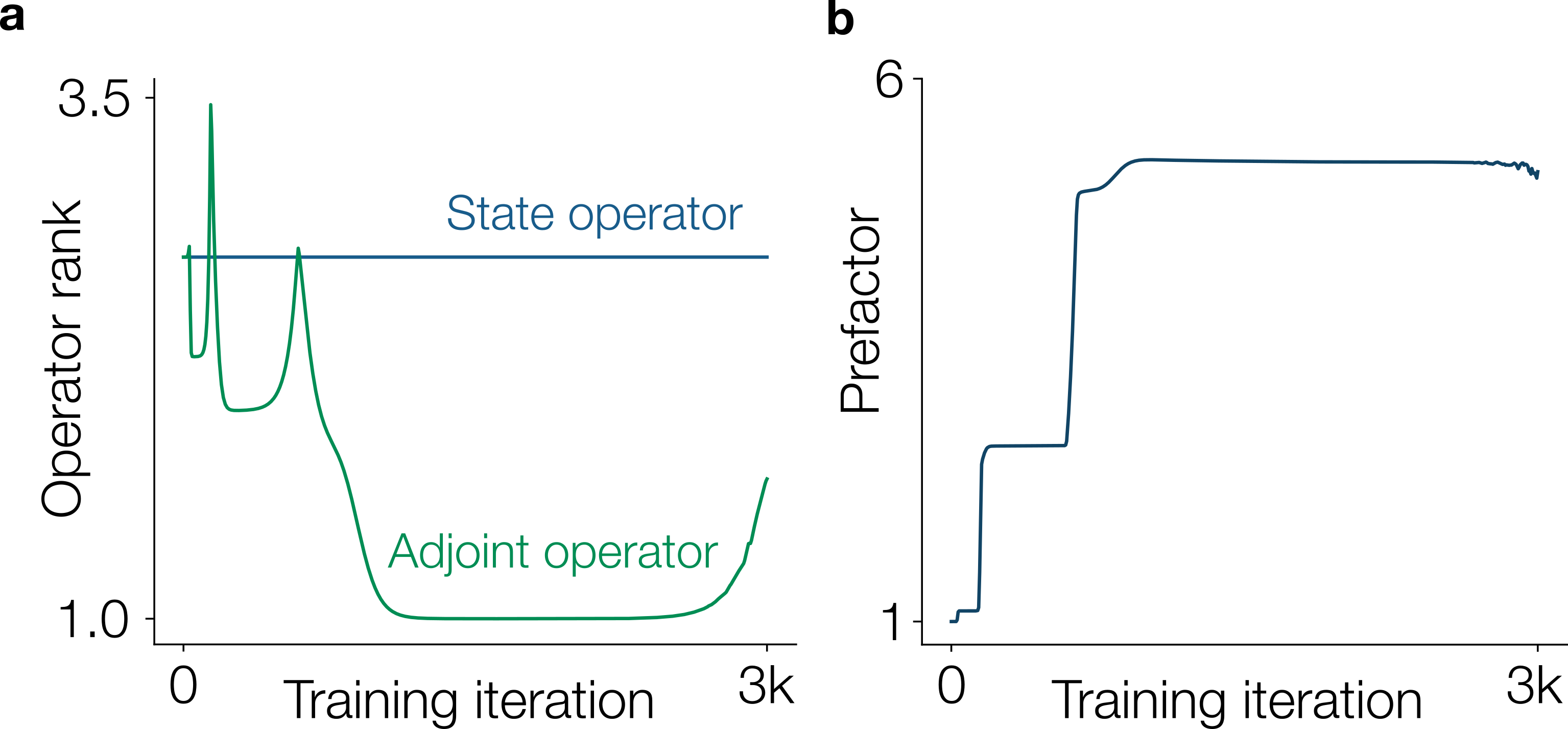}
    \caption{\textbf{Flow matching operators under imbalanced modes. }\textbf{a.} State and adjoint operator stable ranks over training. As expected the state operator depends on the data distribution and is constant. Adjoint rank decreases as modes are progressively learned \textbf{b.} Rank bound prefactor over training. Increases with each learned mode.}
    \label{fig:apd_linear_ops}
\end{figure}

\subsection{Non-linear flow matching}\label{apd:nonlinear_fm}

In section~\ref{sec:non-linear_FM}, we consider a task of learning a mixture of Gaussians target distribution with isotropic covariances:
\[
\mathbf{z}\sim p^*(\mathbf{z})=\sum_{i=1}^{m} \pi_i \mathcal{N}(\mathbf{z}|\boldsymbol{\mu}_i, \sigma_*^2 I_d)\in\mathbb{R}^d
\]
from a unimodal initial distribution
\[
\mathbf{x}(0)\sim p_0(\mathbf{x})=\mathcal{N}(\mathbf{x}|\mathbf{0}, \sigma^2 I_d)\in\mathbb{R}^d.
\]
with $\sigma=1.0$ and $\sigma_*=0.1$. The mixture weights are distributed according to a power law: $\pi_i=\frac{i^{-\alpha}}{\sum_i i^{-\alpha}}$, $\alpha=3$.
We choose the linear interpolation probability path, $\bar{\mathbf{x}}(t)=t\mathbf{z}+(1-t)\mathbf{x}(0)$. Motivated by the form of the analytic field derived in section~\ref{apd:optimal_FM_field}, we parameterise our model as:
\[
    \mathbf{u}_\theta(\mathbf{x},t)=c(t)\mathbf{x}+(1-tc(t))W_2\text{ReLU}(W_1\mathbf{x}+\mathbf{b})
\]
where $c(t)$ is given explicitly to the model and $W_2\in\mathbb{R}^{n\times m}$, $W_1\in\mathbb{R}^{m\times n}$, $\mathbf{b}\in\mathbb{R}^m$ are trained parameters.
We train using the conditional velocity objective:
\[
    \mathcal{L}=\mathbb{E}_{\mathbf{x}(0),\mathbf{z}}\left[\int_0^1 ||\mathbf{u}_\theta(\bar{\mathbf{x}}(t),t)-(\mathbf{z}-\mathbf{x}(0))||^2 dt\right]
\]
To train the model, we use full-batch gradient descent with a fixed set of initial samples and time points. We initialise weights as $W_{ij}\sim\mathcal{N}(0, \sigma_{\text{init}}/\sqrt{n})$ and $\mathbf{b}=\mathbf{0}$. The hyperparameters are:

\begin{center}
    \begin{tabular}{ccccccc}
    \hline
    $n$ & $m$ & $\sigma_{\text{init}}$ & lr & Iterations & Batch size & Optimizer \\ \hline
    5  & 64 & $1e^{-4}$   & $5e^{-4}$ & 10k & 1000 & Adam  \\ \hline
    \end{tabular}
\end{center}

\textbf{Learned density} To compute the log-density of the learned model, $\log p_{\theta,t}(\mathbf{x})$, we use the instantaneous change of variables for probability flow, derived from the continuity equation\cite{chen_neural_2018}:
\[
    \log p_{\theta,t}(\mathbf{x})=\log p_0(\mathbf{x}) - \int_0^t \nabla\cdot u_\theta(\mathbf{x},t) dt
\]
By solving the same augmented ODE as in appendix~\ref{apd:CNF_train_details}, we can compute the accumulated log-density at time $t$ at a grid of initial points.

\textbf{Time to learning vs mode imbalance} In figure~\ref{fig:nonlinear-fm}c, we plot the time taken to learn the smallest mode as a function of the imbalance between the weightings of the largest and smallest modes. We used a symmetric 3-mode setup in 2-d, with weightings $\boldsymbol{\pi}=[\alpha, \frac{1-\alpha}{2}, \frac{1-\alpha}{2}]$, and with $\alpha\in[0.04, 0.16]$. We define the time to learn as the first training iteration (except initialisation) where any generated sample is within a threshold distance of the smallest weighted mode.

\textbf{Validation on more realistic data.} To test whether the scaling of learning time with mode imbalance extends to more realistic data, we trained a one-hidden-layer ReLU MLP with 512 hidden units on a subset of MNIST containing two digit classes. To continuously control the relative weight of the two classes during training, we assigned separate weights to their losses:
\[
    \mathcal{L}(\theta) = w\mathcal{L}_{1}(\theta) + (1-w)\mathcal{L}_2(\theta),
\]
where $w\in(0,0.5]$ is the class weight. This weighting is equivalent to varying the relative proportions of the two classes in the data distribution, since the expected population loss under a mixture distribution decomposes into a weighted sum of the conditional expected losses for each class:
\[
    \mathcal{L} = \mathbb{E}_{x \sim p_w}[\ell(\theta;x)]
= w \mathbb{E}_{x \sim p_1}[\ell(\theta;x)] + (1-w)\mathbb{E}_{x \sim p_2}[\ell(\theta;x)]
\]
where $p_w(x) = wp_1(x) + (1-w)p_2(x)$. To approximate the gradient-flow regime, we used full-batch gradient descent with a small learning rate, replacing the conditional expectations with empirical averages over all samples from each class.

We considered the minority mode to be learned when its relative loss change $\rho_k = (\mathcal{L}_1(\theta_k)-\mathcal{L}_1(\theta_0))/\mathcal{L}_1(\theta_0)$ fell below a threshold $\delta$. However, fluctuations in the loss during training due to the approximation of gradient flow (e.g. step size) can cause repeated crossings of this threshold, making the first crossing an unreliable estimate of learning time. To obtain a more robust quantity, we instead defined a soft learning time by using a soft sum of the number of iterations this condition was not satisfied.
\[
    T = \sum_{k=0}^{n}\left(1 - \sigma\left(\beta(\delta-\rho_k)\right)\right),
    \qquad
    \sigma(u) = \frac{1}{1 + e^{-u}},
\]
where $\beta$ controls the sharpness of the threshold. In the gradient-flow limit, the learning time $T$ continuously depends on gradient flow $w$. We trained twelve models with $w \in \{0.1, \ldots, 0.5\}$. As for the linear models, we observed a power-law scaling of the learning time with the weight, but with lower exponent ($0.749$), akin to the nonlinear tractable model trained on the Gaussian mixture data. 

\begin{figure}[h]
    \centering
    \includegraphics[width=0.3\linewidth]{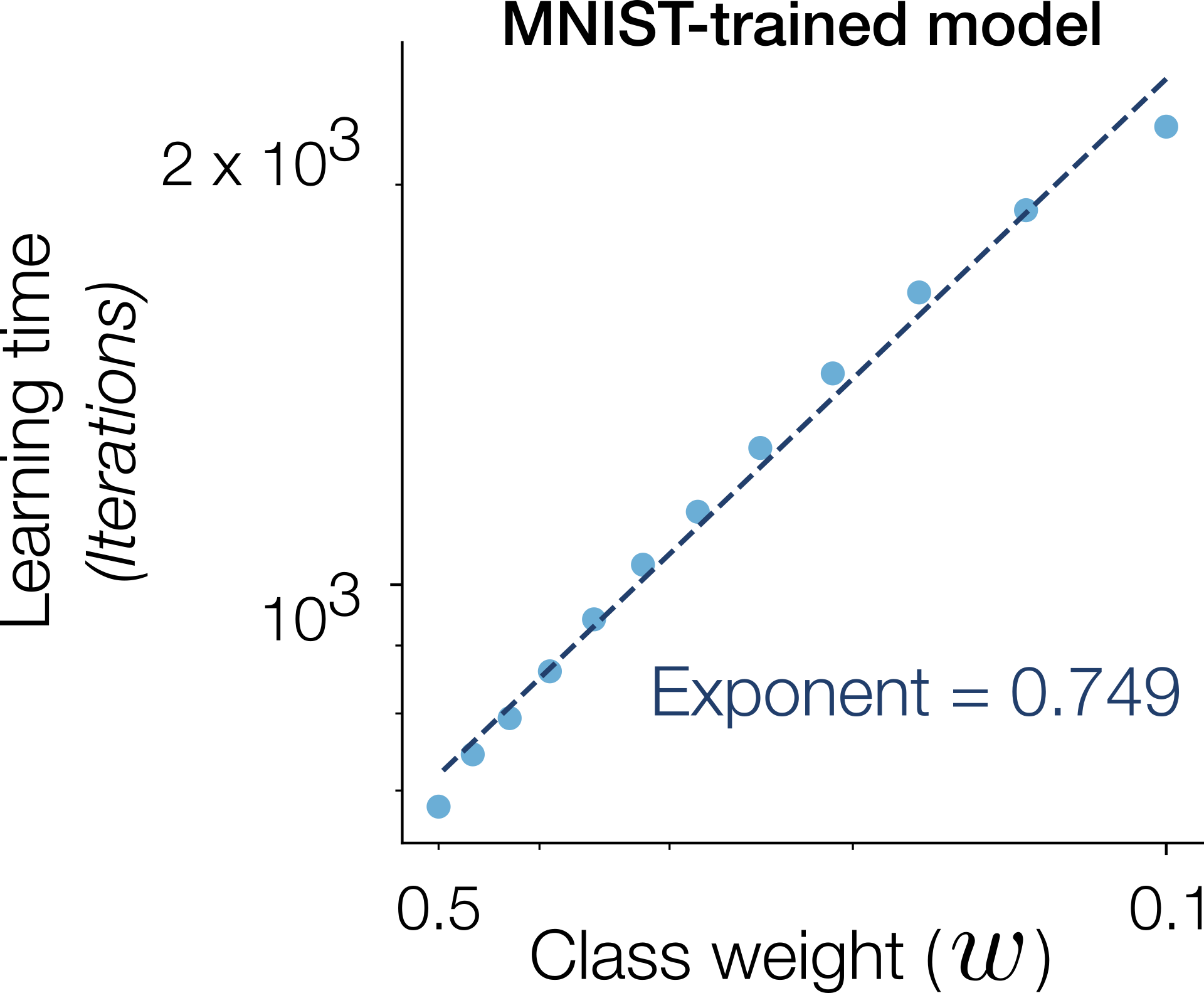}
    \caption{\textbf{Learning time of the smallest data mode as a function of the mode weight.}}
    \label{sup_fig:mnist}
\end{figure}

\subsection{Marginal distribution for MoG targets}\label{apd:FM_marginal_dist} At time $t$, the linear interpolation path yields the conditional (on a data sample $\mathbf{z}$) probability distribution
\[
p_t(\mathbf{x}|\mathbf{z})=\mathcal{N}(\mathbf{x}|t \mathbf{z}, (1-t)^2\sigma^2 I_d).
\]
Marginalising over the data distribution:
\begin{align*}
p_t(\mathbf{x})&=\int p_t(\mathbf{x}|\mathbf{z})p^*(\mathbf{z})d\mathbf{z} \\
&=\int \left(\mathcal{N}(\mathbf{x}|t \mathbf{z}, (1-t)^2\sigma^2 I_d)\right)\left(\sum_{i=1}\pi_i \mathcal{N}(\mathbf{z}|\mu_i, \sigma_*^2 I_d)\right)d\mathbf{z} \\
&= \sum_{i=1}^m \pi_i \int \mathcal{N}(\mathbf{x}|t \mathbf{z}, (1-t)^2\sigma^2 I_d)\mathcal{N}(\mathbf{z}|\mu_i, \sigma_*^2 I_d) d\mathbf{z} \\
&= \sum_{i=1}^m \pi_i p_t^{(i)}(\mathbf{x})
\end{align*}
where $p_t^{(i)}(\mathbf{x})=\mathcal{N}(\mathbf{x}|t\mu_i, \Sigma_t)$ is the probability density conditioned on target mode $i$ and $\Sigma_t=(t^2\sigma_*^2+(1-t)^2\sigma^2)I_d$.

\subsection{Marginal velocity field for MoG targets}\label{apd:optimal_FM_field}

We start by writing the conditional velocity field for the linear interpolation path:
\[
    \mathbf{u}_t(\mathbf{x}|\mathbf{z})=\dot{\mathbf{x}}(t)=\mathbf{z}-\mathbf{x}(0)
\]
Rearranging the probability path equation we get 
\[\mathbf{x}(0)=\frac{\bar{\mathbf{x}}(t)-t\mathbf{z}}{1-t}\]
which we can substitute into the conditional velocity to remove dependency on $\mathbf{x}_0$:
\[ 
    \mathbf{u}_t(\mathbf{x}|\mathbf{z})=\frac{1}{1-t}(\mathbf{z}-\mathbf{x})
\]
To obtain the marginal field (for the $i^{\text{th}}$ target mode) we marginalise over $\mathbf{z}$ by taking the expectation with respect to the posterior distribution $p_t(\mathbf{z}|\mathbf{x})$:
\begin{align*}
    u_t^{(i)}(\mathbf{x})&=\mathbb{E}_{\mathbf{z}\sim p_t(\mathbf{z}|\mathbf{x})}[u_t(\mathbf{x}|\mathbf{z})]\\
    &=\frac{1}{1-t}(\mathbb{E}_{\mathbf{z}\sim p_t(\mathbf{z}|\mathbf{x})}\left[\mathbf{z}\right]-\mathbf{x})
\end{align*}
To evaluate the expectation, we use the following result:
\begin{theorem}
Let $\mathbf{u} \sim \mathcal{N}(\mathbf{a}_{\mathbf{u}}, B_{\mathbf{u}})$ and $\mathbf{v} \sim \mathcal{N}(\mathbf{a}_{\mathbf{v}}, B_{\mathbf{v}})$.

$$ p(\mathbf{u}|\mathbf{v} + \mathbf{u}) = \mathcal{N} \left( \mathbf{a}_{\mathbf{u}} + B_{\mathbf{u}}(B_{\mathbf{u}} + B_{\mathbf{v}})^{-1} \left[ (\mathbf{u} + \mathbf{v}) - (\mathbf{a}_{\mathbf{u}} + \mathbf{a}_{\mathbf{v}}) \right], \quad B_{\mathbf{u}} - B_{\mathbf{u}}(B_{\mathbf{u}} + B_{\mathbf{v}})^{-1}B_{\mathbf{u}} \right) $$

I.e. the conditional distribution of the sum of two Gaussians is itself Gaussian.
\end{theorem}
Applying this, we obtain $\mathbb{E}[\mathbf{z}|\mathbf{x}]=\boldsymbol{\mu}_i+t\sigma_*^2I_d\Sigma_t^{-1}(\mathbf{x}-t\boldsymbol{\mu}_i)$ where $\Sigma_t =t^2\sigma_*^2+(1-t)^2\sigma^2$. Thus the field becomes:
\begin{align*}
    \mathbf{u}^{(i)}_t(\mathbf{x})&=\frac{1}{1-t}\left(\boldsymbol{\mu}_i+t\sigma_*^2I_d\Sigma_t^{-1}(\mathbf{x}-t\boldsymbol{\mu}_i)\right) -\frac{1}{1-t}\mathbf{x} \\
    &\stackrel{(1)}{=} \frac{1}{1-t}\boldsymbol{\mu}_i - \frac{1}{1-t}(\mathbf{x} - t\boldsymbol{\mu}_i) - \frac{t}{1-t}\boldsymbol{\mu}_i + \frac{t}{1-t}\sigma_*^2 I \Sigma_t^{-1}(\mathbf{x} - t\boldsymbol{\mu}_i) \\
    &\stackrel{(2)}{=} \boldsymbol{\mu}_i +\left[\frac{t}{1-t}\sigma_*^2I - \frac{1}{1-t}\Sigma_t \right]\Sigma_t^{-1}(\mathbf{x}-t\boldsymbol{\mu}_i) \\
    &\stackrel{(3)}{=}\boldsymbol{\mu}_i +\frac{t\sigma_*^2-(1-t)\sigma^2}{t^2 \sigma_*^2 +(1-t)^2 \sigma^2}(\mathbf{x}-t\boldsymbol{\mu}_i)
\end{align*}
where in $(1)$ we have added and subtracted by $\frac{t}{1-t}\boldsymbol{\mu}_i$, in $(2)$ we took out a factor of $\Sigma_t^{-1}(\mathbf{x}-t\boldsymbol{\mu}_i)$ on the right, and in $(3)$ we have substituted in the expression for $\Sigma_t$ and simplified.
Defining $c(t)=\frac{t\sigma_*^2-(1-t)\sigma^2}{t^2\sigma_*^2+(1-t)^2 \sigma^2}$ we can write the field as
\[
    \mathbf{u}^{(i)}_t(\mathbf{x})=c(t)\mathbf{x}+(1-tc(t))\boldsymbol{\mu}_i
\]
Note that this is still conditional on a specific target mode $\boldsymbol{\mu}_i$. To get the full unconditional field, we use the continuity equation:
\begin{theorem}[Continuity Equation]\label{thrm:continuity}
For a flow model with vector field $u_t$ and $\mathbf{x}_0\sim p_\text{init}$, then $\bar{\mathbf{x}}_t\sim p_t$ for all $0\leq t \leq1$ if and only if
\[
\partial_t p_t(\mathbf{x})=-\nabla\cdot(p_t u_t)(\mathbf{x}) 
\]
for all $\mathbf{x}\in\mathbb{R}^d$, $0\leq t\leq1$.
\end{theorem}

Then, using the form of the marginal probability path in section~\ref{apd:FM_marginal_dist}, we can write the LHS as
\begin{align*}
    \partial_tp_t(\mathbf{x})
    &=\partial_t\left(\sum_{i=1}^m \pi_i p_t^{(i)}(\mathbf{x})\right) \\
    &\stackrel{(1)}{=} \sum_{i=1}^m \pi_i \partial_t p_t^{(i)}(\mathbf{x}) \\
    &\stackrel{(2)}{=} -\sum_{i=1}^m \pi_i \nabla\cdot(p_t^{(i)}u_t^{(i)})(\mathbf{x}) \\
    &\stackrel{(3)}{=}-\nabla\cdot\left(\sum_{i=1}^m \pi_i p_t^{(i)}(\mathbf{x})u_t^{(i)}(\mathbf{x})\right)
\end{align*}
where in $(1)$ we used the linearity of derivatives, in $(2)$ we used the continuity equation for the $i^{\text{th}}$ target mode and in $(3)$ we use the linearity of the divergence operator. Comparing this to the RHS of the continuity equation, we get that
\[
p_t(\mathbf{x})u_t(\mathbf{x})=\sum_{i=1}^m \pi_i  p_t^{(i)}(\mathbf{x})u_t^{(i)}(\mathbf{x})
\] 
which we can rearrange to get
\[
    u_t(\mathbf{x})
    = \sum_{i=1}^m\gamma_t^{(i)}(\mathbf{x})u_t^{(i)}(\mathbf{x})
\]
where $\gamma_t^{(i)}(\mathbf{x})=\frac{\pi_i p_t^{(i)}(\mathbf{x})}{p_t(\mathbf{x})}$ acts as a weighting between different modes. Substituting the expression for the field conditioned on target mode $i$, we obtain the final form of the marginal field:
\[
    \mathbf{u}_t(\mathbf{x})=c(t)\mathbf{x}+(1-tc(t))\sum_{i=1}^m \gamma_t^{(i)}(\mathbf{x})\boldsymbol{\mu}_i
\]
where we used the fact that $\sum_i\gamma_t^{(i)}(\mathbf{x})=1$.

\subsection{Operators of the non-linear model}\label{apd:fm_non_linear_operators}
Consider the loss functional:
\[
    \mathcal{L}=\mathbb{E}_{\mathbf{x}_0,\mathbf{x}_1}\left[\int_0^1 \ell(\bar{\mathbf{x}}(t))dt\right]
\]
where $\ell(\bar{\mathbf{x}}(t))=||\Phi(\bar{\mathbf{x}}(t))-\mathbf{u}(\bar{\mathbf{x}}(t))||^2$ is the MSE objective for model $\Phi$ and target velocity $\mathbf{u}$, and $\bar{\mathbf{x}}(t)$ is the interpolated trajectory. Also consider a deep neural network architecture:
\[
    \Phi=W_k \circ \phi \circ W_{k-1} \circ \phi \circ\dots\circ W_1
\]
which has $i^{\text{th}}$ layer pre-activations $\mathbf{z}_i=W_i\circ\phi\dots$ and post-activations $\mathbf{h}_i=\phi(\mathbf{z}_i)$ for some non-linearity $\phi$. The gradients with respect to the $i^{\text{th}}$ layer weights are:
\begin{align*}
    \nabla_{W_i}\mathcal{L}&=\mathbb{E}_{\mathbf{x}_0, \mathbf{x}_1}\left[\int_0^1 \partial_{\mathbf{z}_i} \ell(\bar{\mathbf{x}}(t)) \otimes \mathbf{h}_{i-1}(\bar{\mathbf{x}}(t)) dt\right] \\
    &= \mathbb{E}_{\mathbf{x}_0, \mathbf{x}_1}\left[\int_0^1 R(\bar{\mathbf{x}}(t)) \frac{\partial \Phi}{\partial\mathbf{z}i} \otimes \mathbf{h}_{i-1}(\bar{\mathbf{x}}(t)) dt\right]
\end{align*}
where $R(\bar{\mathbf{x}}(t))=\Phi(\bar{\mathbf{x}}(t))-\mathbf{u}(\bar{\mathbf{x}}(t))$ is the residual per sample.

Now we define our network as in section~\ref{sec:non-linear_FM} (omitting the skip connection which contributes zero to the gradient, and the time-dependent pre-factor which just scales the gradient):
\[
    \Phi(\mathbf{x}) = W_2 \text{ReLU}(W_1\mathbf{x}+\mathbf{b})
\]
We can then compute the gradients as:
\[
    \nabla_{W_1}\mathcal{L}=\mathbb{E}_{\mathbf{x}_0, \mathbf{x}_1}\left[\int_0^1 \underbrace{\text{diag}\left(\text{ReLU}'(\mathbf{z}_1)\right)W_2^\top R(\bar{\mathbf{x}}(t))}_{A_1} \otimes \underbrace{\bar{\mathbf{x}}(t)}_{\bar{H}_1} dt\right]
\]
\[
    \nabla_{W_2}\mathcal{L}=\mathbb{E}_{\mathbf{x}_0, \mathbf{x}_1}\left[\int_0^1 \underbrace{R(\bar{\mathbf{x}}(t))}_{A_2} \otimes \underbrace{\mathbf{h}_1(\bar{\mathbf{x}}(t))}_{\bar{H}_2} dt\right]
\]
where we define $A_1$, $A_2$ as the adjoint (backward) operators and $\bar{H}_1$, $\bar{H}_2$ as the state (forward) operators for layer 1 and layer 2 respectively.

\section{Code and compute requirements}
The appendix contains all architectures and training parameters necessary to reproduce the analyses. A Git repository containing all training functions and analyses notebooks will be released upon publication. 

\textbf{Packages} The code makes use of standard python packages for computations and model training: NumPy, JAX, Diffrax, Equinox. 

\textbf{Compute} The linear CNF and non-linear FM models were trained in under an hour using an A100 GPU. The linear FM models were trained in a few minutes on a 16 core CPU.

\AtNextBibliography{\small}
\printbibliography[heading=subbibliography,title={\large Supplementary References}]
\end{refsection}
%\clearpage

%\input{checklist}

\end{document}